\documentclass[conference]{IEEEtran}
\IEEEoverridecommandlockouts
\usepackage{cite}
\usepackage{amsmath,amssymb,amsfonts}
\usepackage{algorithmic}
\usepackage{graphicx}
\usepackage{textcomp}
\usepackage{xcolor}
\usepackage{amssymb}
\usepackage[hidelinks]{hyperref}
\usepackage{booktabs}
\usepackage{xcolor}
\usepackage{bbold}
\usepackage{multirow}
\usepackage{booktabs}
\usepackage{subcaption} 
\usepackage{adjustbox}

\definecolor{MyDarkGreen}{rgb}{0.0, 0.5, 0.0} 
\def\BibTeX{{\rm B\kern-.05em{\sc i\kern-.025em b}\kern-.08em
    T\kern-.1667em\lower.7ex\hbox{E}\kern-.125emX}}
\usepackage[hidelinks]{hyperref}
\begin{document}

\title{Link Prediction with Physics-Inspired\\ Graph Neural Networks

\thanks{This work was partially supported by projects FAIR (PE0000013) and SERICS (PE00000014) under the MUR National Recovery and Resilience Plan funded by the European Union - NextGenerationEU.  This work has been supported by the project NEREO project funded by the Italian Ministry of Education and Research (PRIN) Grant no. 2022AEFHAZ}
}
\author{%
  \IEEEauthorblockN{%
    Andrea Giuseppe Di Francesco\textsuperscript{\textsection}\IEEEauthorrefmark{1}\IEEEauthorrefmark{2},
    Francesco Caso\IEEEauthorrefmark{1},
    Maria Sofia Bucarelli\IEEEauthorrefmark{1},
    Fabrizio Silvestri\IEEEauthorrefmark{1}\IEEEauthorrefmark{2}%
  }%
}

\maketitle

\begingroup
  \renewcommand\thefootnote{\fnsymbol{footnote}}
  \footnotetext[1]{Department of Computer Science, Control and Management Engineering, Sapienza University of Rome, Rome, Italy.}
  \makeatletter\def\Hy@Warning#1{}\makeatother\footnotetext[2]{Institute of Information Science and Technologies "Alessandro Faedo" - ISTI-CNR, Pisa, Italy.}
  \renewcommand\thefootnote{\textsection}
  \footnotetext{Corresponding author, \texttt{difrancesco@diag.uniroma1.it}.}
  \renewcommand\thefootnote{\arabic{footnote}} 
\endgroup

\begin{abstract}
The message-passing mechanism underlying Graph Neural Networks (GNNs) is not naturally suited for heterophilic datasets, where adjacent nodes often have different labels. 
Most solutions to this problem remain confined to the task of node classification. In this article, we focus on the valuable task of link prediction under heterophily, an interesting problem for recommendation systems, social network analysis, and other applications.  GNNs like GRAFF have improved node classification under heterophily by incorporating physics biases in the architecture. Similarly, we propose GRAFF-LP, an extension of GRAFF for link prediction. We show that GRAFF-LP effectively discriminates existing from non-existing edges by learning implicitly to separate the edge gradients.  
Based on this information, we propose a new readout function inspired by physics. Remarkably, this new function not only enhances the performance of GRAFF-LP but also improves that of other baseline models, leading us to reconsider how every link prediction experiment has been conducted so far.
Finally, we provide evidence that even simple GNNs did not experience greater difficulty in predicting heterophilic links compared to homophilic ones. This leads us to believe in the necessity for heterophily measures specifically tailored for link prediction, distinct from those used in node classification. The code and appendix are available at \url{https://github.com/difra100/Link_Prediction_with_PIGNN_IJCNN}.
\end{abstract}
\section{Introduction}
Graph Neural Networks (GNNs) work as feature extractors that can be trained to perform some typical tasks on graphs, such as \textit{link prediction}, and \textit{node} or \textit{graph classification}.\\
Among these, link prediction consists of computing the probability of a link between two nodes.\\
Most GNNs rely on the \textit{message-passing} formalism \cite{messagepassing}.
In heterophilic graphs, where connected nodes tend to have different labels, Message-Passing Neural Networks (MPNNs) may struggle in the classification task as they tend to generate similar representations for adjacent nodes \cite{geomgcn}, an issue commonly known as over-smoothing \cite{oversmoothing}.
There have been efforts to improve the performance of GNNs on heterophilic graphs \cite{neuralsheaf_heterophily}, however, these works are limited to the context of node classification. An example of this is the \textit{gradient flow framework} (GRAFF) \cite{GRAFF2}, which deals with heterophily through physics-inspired biases.  Although GRAFF shows competitive performance in both heterophilic and homophilic graphs, existing research has focused solely on node classification tasks, not investigating its potential in link prediction, which has significant applications in several domains. This task becomes particularly complex for heterophilic graphs, where we may have a link between two nodes with dissimilar characteristics. Unlike homophilic graphs, where connected nodes are similar, the reason behind the connection of two dissimilar entities can be latent.
To the best of our knowledge, link prediction under heterophily has only been discussed and brought to the community's attention by \cite{DisenLink}. \\
We propose GRAFF-LP (GRAFF for Link Prediction), a link prediction framework built upon GRAFF for node classification \cite{GRAFF2}. 
We tested our model on newly introduced heterophilic datasets \cite{criticallookatgnn} to overcome the limitations of standard benchmarks. The contributions of our work are the following. 
\begin{enumerate}
   \item We are the first to propose a Physics-Inspired GNN for link prediction.
   \item We demonstrate that GRAFF-LP can achieve competitive performance w.r.t. the other examined methods, showing consistent performance across graphs from different contexts and increasing size.
   \item We propose a novel Physics-Inspired readout function that leads to consistent performance improvements for GRAFF-LP as well as other baseline models. Additionally, the new readout gives GRAFF-LP more transparency in its behavior at inference time.
   \item We set a new link prediction baseline on a recently created collection of heterophilic graphs \cite{criticallookatgnn}, originally designed for node classification. This baseline serves as a foundation for future work in this area. Enhancing the current, yet not well-explored, literature on link prediction under heterophily.
   \item Unexpectedly, we show that most of the time, classic models do not struggle in predicting the connections between nodes of different classes, conversely with what happens with heterophilic node classification. 
\end{enumerate}

\begin{figure*}[t]
    \centering
    \includegraphics[width =0.85\linewidth]{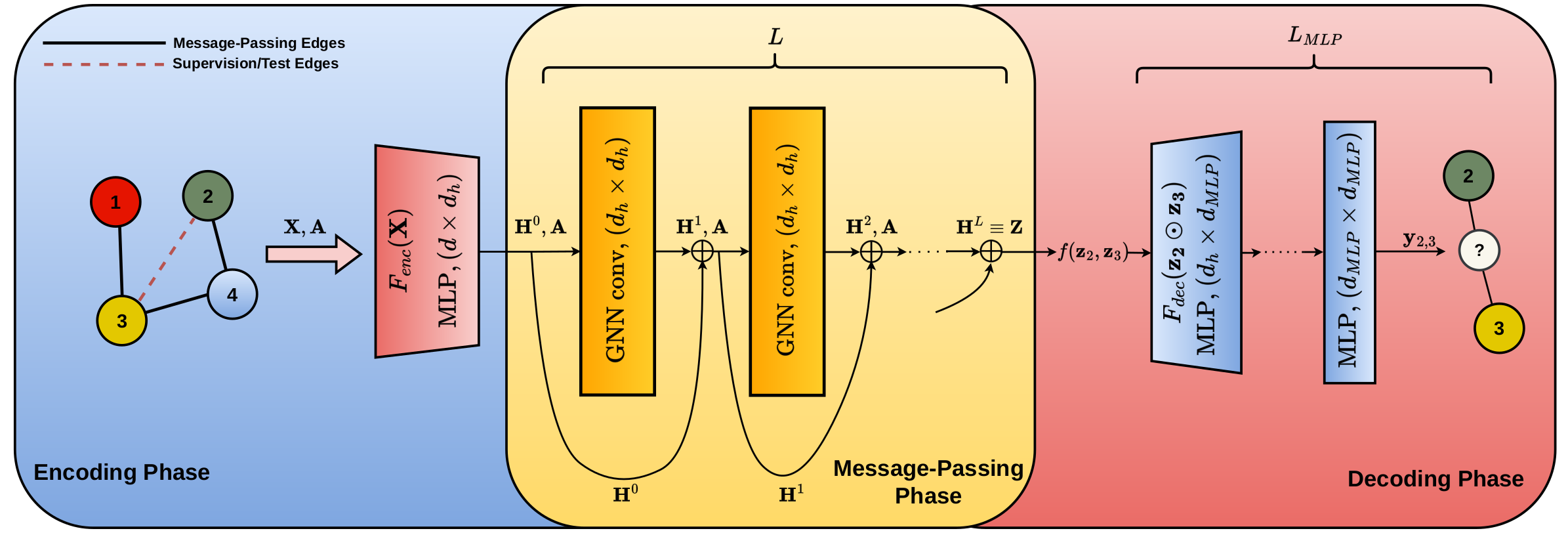}
    \caption{General overview of the link prediction pipeline.}
    \label{fig:architecture}
\end{figure*}
\section{Related Works} 
\textbf{Graph Neural Networks for Link Prediction.} Link prediction methods can be divided into \textit{non-neural-based} approaches, such as heuristics \cite{GNNBook}, and \textit{neural-based} methods, such as GNNs. Although heuristics work well in specific cases, GNNs offer a general framework by learning both graph structure and content features simultaneously. Neural-based methods include node-based models like Graph Auto-Encoders \cite{VGAE} and \textit{subgraph-based} paradigm, led by SEAL \cite{SEAL}. However, while more expressive, subgraph-based methods are inefficient when scaling the graph size. Additionally, \cite{heart} showed that when datasets are carefully split, the performance gap between node-based and subgraph-based models is not as significant. Our work builds upon the node-based paradigm.\\
\textbf{Physics-Inspired vs. Physics-Informed.} Physics-Informed (PI) neural networks incorporate physical priors to improve performance and generalizability \cite{PIRL}. Physics-Inspired (PIrd) networks are a subset of PI methods, where physical constraints are embedded in the architecture itself, acting as inductive biases \cite{graphcoupledoscillatornetworks}.\\
\textbf{Physics-Inspired Graph Neural Networks.} PIrd GNNs incorporate physics principles directly into the model's structure. Examples include GNNs based on gradient flows \cite{GRAFF2}, reaction-diffusion equations \cite{GREAD}, based on nonlinear controlled and damped oscillators \cite{graphcoupledoscillatornetworks}. While these approaches have been applied to node classification, they remain unexplored for link prediction. Our work provides the first perspective on PIrd biases in this setting.
The extended discussion on related works is available in Appendix \ref{sec:relatedworks}.
\section{Preliminaries}
\textbf{Notation}. Let $\mathcal{G} = (\mathcal{V}, \mathcal{E})$ be an undirected graph, with $\mathcal{V}$ the set of nodes and $\mathcal{E} \subseteq \mathcal{V} \times \mathcal{V}$ be the set of edges. $|\mathcal{V}| = N$ is the number of nodes.
We denote by $\Gamma(i)$ the neighborhood of the node $i$. 
$\mathbf{D}$ is the diagonal matrix in  $  \mathbb{R}^{N \times N}$, such that
$D_{ii} = |\Gamma(i)|$.
$\mathbf{x}_i \in \mathbb{R}^d$ represents the features  of node $i$ and $y_i$ its label.
The node representations can be ordered in a unique matrix, which we refer to as the \textit{instance matrix} $\mathbf{X} \in \mathbb{R}^{N \times d_0}$. 
$\mathbf{A} \in \{0, 1\}^{N \times N}$ is the adjacency matrix, with $A_{ij} = 1$ if nodes $i$ and $j$ are connected, and $A_{ij} = 0$ otherwise. We distinguish $\mathbf{H}^T$ from  $\mathbf{H}^{\top}$, as the hidden representation of the nodes $\mathbf{H}$ at the time-step $T$, and the transpose operation $\cdot^{\top}$.\\
\textbf{Homophily measures}. The homophily assumption in graphs refers to the tendency of similar nodes to be connected. An unambiguous similarity measure is missing in the literature. Those that have been used the most with GNNs are \textit{edge homophily} $\xi_{edge}$ \cite{edgehomophily} and \textit{node homophily} $\xi_{node}$ \cite{geomgcn}. The former was more considered within the node classification benchmarks \cite{neuralsheaf_heterophily, GRAFF2}, and can be computed as 
\begin{equation}
\label{eq:edge_hom}
    \xi_{edge} = \frac{|(i, j) \in \mathcal{E} : y_i = y_j|}{|\mathcal{E}|}
\end{equation}
Both $\xi_{edge}$ and $\xi_{node}$ rely on the labels associated with the nodes. As an example, Equation \eqref{eq:edge_hom} measures the fraction of edges that connect nodes from the same class. Generally, if we record a low homophily, we consider the graph as heterophilic. 
Traditional homophily measures are unsuitable for cross-datasets comparison because they are sensitive to class numbers and sample balance \cite{adjustedhomophily,criticallookatgnn}. To address this, the adjusted homophily metric $\xi_{adj}$ was introduced in \cite{adjustedhomophily}:
\begin{equation}
    \xi_{adj} = \frac{\xi_{edge} - \sum_{k \in \mathcal{S}} \mathcal{D}_k^{2}/(2|\mathcal{E}|)^2}{ 1 - \sum_{k \in \mathcal{S}} \mathcal{D}_k^{2}/(2|\mathcal{E}|)^2},
\end{equation}
$\mathcal{Y} = \{1, ...., C\}$ is the set of possible labels associated with each node, and $\mathcal{D}_k = \sum_{i: y_i = k} D_{ii}$. This measure is comparable across graphs and upper-bounded by 1, but it lacks a lower bound and does not consider node features. \\
\textbf{Graph Neural Networks as gradient flow}. Let us consider an $N$-dimensional dynamic system evolving as $\dot{\mathbf{H}}(t) = F(\mathbf{H}(t))$, with $\mathbf{H}(t) \in \mathbb{R}^{N \times d}$
If there exists a function $E: \mathbb{R}^{N \times d} \rightarrow \mathbb{R}$, s.t. $F(\mathbf{H}(t)) = - \nabla E(\mathbf{H}(t))$, the evolution equation $\dot{\mathbf{H}}(t)$ is the gradient flow of the energy $E$. 
Gradient flows are useful for studying the underlying dynamics of the system, provided the knowledge of $E$. 
In our case, $N$ represents the graph nodes whose representations 
$\mathbf{H}(t)$ evolve through a GNN over time, and $E(\mathbf{H}(t))$, is an energy functional associated with the node representations.
Let $GNN: \mathbb{R}^d \rightarrow \mathbb{R}^d$ be an intermediate layer of a generic GNN.
By treating the GNN layers as continuous time $t$ and defining  $GNN(\mathbf{H}(t)) = - \nabla E(\mathbf{H}(t))$, the evolution of the features through the GNN is described as the gradient flow of $E(\mathbf{H}(t))$. $E$ can be selected as the Dirichlet Energy $E^{dir}$:
\begin{align}
\label{eq: dirichlet}
   E^{dir}(\mathbf{H}(t)) &:= \sum_{(i,j) \in \mathcal{E}} \| (\nabla \mathbf{H}(t))_{ij} \|^2 
\end{align}
where \( (\nabla \mathbf{H}(t))_{ij} = \frac{\mathbf{h}^t_j}{\sqrt{D_{jj} + 1}} -  \frac{\mathbf{h}^t_i}{\sqrt{D_{ii} + 1}} \) and 
 $\mathbf{h}^t_i$ denotes the feature of node $i$ at time $t$.
Poor performance on heterophilic graphs and over-smoothing are often linked to the Dirichlet energy of features decaying to zero as layers increase \cite{oversmoothingdirichlet, oversmoothingdirichlet2}. Equation \eqref{eq: dirichlet} shows that decreasing $E^{dir}$ brings adjacent nodes closer in feature space. Conversely, from what was commonly thought, \cite{GRAFF2} proved that linear graph convolutions with symmetric weights shared among layers can induce edge-wise attraction (repulsion) through their positive (negative) eigenvalues. This control mechanism effectively influences whether the features are smoothed or sharpened, making the model successfully handle node classification within heterophilic graphs. This was possible by defining layers as a gradient flow of a parametrized Dirichlet energy:
\begin{align}
    \label{eq:param_dirich2}
    E_{\theta}^{dir}(\mathbf{H}(t)) &= \sum_i \langle \mathbf{h}^t_i, \mathbf{\Omega} \mathbf{h}^t_i \rangle - \sum_{i,j} a_{ij} \langle \mathbf{h}_i, \mathbf{W} \mathbf{h}^t_j \rangle \\
    \nonumber
    &= \hspace{-3pt} \sum_i \langle\mathbf{h}^t_i, (\mathbf{\Omega}\hspace{-1pt} - \hspace{-1pt}\mathbf{W}) \mathbf{h}^t_i \rangle  \hspace{-3pt}  + \frac{1}{2} \sum_{i,j} \hspace{-1pt} \| \Theta_{+}( \nabla \mathbf{H})_{ij} \|^2 \\
    \nonumber
    & -  \frac{1}{2} \sum_{i,j} \| \Theta_{-}( \nabla \mathbf{H}(t))_{ij} \|^2,
\end{align}
 $\Theta_{+}$ and $\Theta_{-}$ depend on the positive and negative eigenvalues of the weight matrices, respectively.
By applying Euler discretization to the gradient flow and replacing $\mathbf{H}(t)$ by $\mathbf{H}^t$:
\begin{equation}
\label{eq: discrete_gradient_flow3}
  \mathbf{H}^{t+\tau} =  \mathbf{H}^{t} + \tau (- \mathbf{H}^{t} \mathbf{\Omega} + \pmb{\mathbb{A}} \mathbf{H}^{t}\mathbf{W} - \mathbf{H}^0 \tilde{\mathbf{W}})  
\end{equation}
 $\tau$ is the integration step, $T = \tau L$ is the total integration time, $\mathbf{\Omega}$, $\mathbf{W}$ and $\tilde{\mathbf{W}}$ are the trainable matrices, that are symmetric and shared across the layers,  and $\pmb{\mathbb{A}} = \mathbf{\tilde{D}}^{-\frac{1}{2}}\mathbf{\tilde{A}}\mathbf{\tilde{D}}^{-\frac{1}{2}}$, since self-loops are included in $\mathbf{A}$ and $\mathbf{D}$ (i.e. $\tilde{\mathbf{A}} = \mathbf{I} + \mathbf{A}$). 
Equation \eqref{eq: discrete_gradient_flow3} takes the form of a residual network \cite{resnet}, where the interpretation of GNNs as gradient flows is referred to in the literature as PIrd GNN.
For details on these derivations, see \cite{GRAFF2}.
In this work, we take advantage of these results and build upon this architecture for link prediction.
\section{Proposed Framework: GRAFF-LP}
\label{sec:method}
GRAFF-LP operates in a transductive setting, where the graph retains all the nodes both in training and inference. Figure \ref{fig:architecture} presents a general overview of our approach.  
The whole scheme is designed as a node-based method for link prediction \cite{GNNBook} and consists of three different phases.\\ 
\textbf{Encoding phase}. This is the transition from $\mathbf{X}$ to $\mathbf{H}^{0}=F_{\text{enc}}(\mathbf{X})$ . In the experiments, we used one linear layer followed by dropout for all the models taken into consideration. \\
\textbf{Message-Passing phase}. This phase is formalized by Equation \eqref{eq: discrete_gradient_flow3}. $L$, $\tau$ and $d_h$, are hyperparameters of our architecture. The latter is the dimensionality of the intermediate representation that is fixed across the layers to maintain the dynamic system interpretation. Following the non-linear gradient flow approach proposed by \cite{GRAFF2}, we interleave layers with non-linear functions, specifically the Rectified Linear Unit (ReLU). Although this means the network is no longer a discretized gradient flow, it still preserves the physical interpretation of the weight matrices. We adopt this strategy due to its improved performance over the linear counterpart in node classification experiments \cite{GRAFF2}. The message-passing of GRAFF-LP  is:
\begin{equation}
\label{eq: discrete_gradient_flow4}
  \mathbf{H}^{t+\tau} =  \mathbf{H}^{t} + \tau \sigma(- \mathbf{H}^{t} \mathbf{\Omega} + \pmb{\mathbb{A}} \mathbf{H}^{t}\mathbf{W} - \mathbf{H}^0 \tilde{\mathbf{W}}),  
\end{equation}
where $\sigma(\cdot)$ is the ReLU operation. To enforce the symmetry in the trainable parameters, we follow the \textit{diagonally-dominant} approach used by \cite{GRAFF2}.\\
\textbf{Decoding phase}. Let $\mathbf{z}_i = \mathbf{h}^L_i$ be the output of the message passing phase for node $i$. We describe the probability that nodes $i$ and $j$ share a link as $\hat{y}$:
\begin{align}
\label{eq: classic_decoder}
    \hat{y}(\mathbf{z}_i, \mathbf{z}_j) = F_{dec}(\mathbf{z}_{i,j}), \; \; \text{ with }   \mathbf{z}_{i,j} = f(\mathbf{z}_i, \mathbf{z}_j)
\end{align}
$f(\cdot)$ is the readout function associated with the link, which aggregates the features coming from two nodes. $F_{dec}$ is an MLP of $L_{MLP}$ layers with width $d_{MLP}$ and nonlinear activations (see Figure \ref{fig:architecture}). 
We use two types of readout in our experiments: 
\begin{align}
\label{eq: readouts}
    f_{h}(\mathbf{z}_i, \mathbf{z}_j) &= \mathbf{z}_i \odot \mathbf{z}_j & \textit{(Hadamard)}, \\
    f_{g}(\mathbf{z}_i, \mathbf{z}_j) &= (\nabla\mathbf{H}^T)_{i,j} \odot (\nabla\mathbf{H}^T)_{i,j} & \textit{(Gradient)}.
\end{align}
The Hadamard readout is commonly used with node-based link predictors \cite{GNNBook}.
In this paper, we are the first to propose an alternative function $f_g$, which we believe is more compatible with the GRAFF backbone. As seen in Equation \eqref{eq:param_dirich2}, gradient flow operates by minimizing or maximizing the squared norm of edge gradients, and $f_g$ can be directly related to them through the following relation: 
\begin{equation}
\label{eq:gradient_connection}
    ||(\nabla\mathbf{H}^T)_{i,j}||^2 = \sum_d (f_{g}(\mathbf{z}_i, \mathbf{z}_j))_d.
\end{equation}
More precisely, the gradient flow operates by minimizing or maximizing the squared norm of the gradients multiplied by the terms $\Theta_+$ and $\Theta_-$; we preferred to define the readout function without including those terms to speed up the calculation of $f_g$. We use the Hadamard product of the gradients to avoid any bias on the edge directionality since we deal with undirected graphs. However, for directed graphs, we could  simply have $f_{g}(\mathbf{z}_i, \mathbf{z}_j) = (\nabla\mathbf{H}^T)_{i,j}$. We do not consider other readouts such as concatenation of $z_i$ and $z_j$, since this would provide a bias on directionality as well, we would lose the physics-inspired bias, and we would double $d_{MLP}$, negatively affecting the space and time complexity.

All the node-based methods that we implement in our experiments follow the scheme in Figure \ref{fig:architecture}, more details on each implementation can be found in the code.

\subsection{Complexity Analysis of GRAFF-LP}
We now consider the space complexity in terms of the number of parameters. GRAFF-LP, due to its weight-sharing mechanism, maintains constant complexity w.r.t. $L$ and exhibits quadratic complexity w.r.t. the number of hidden dimensions, as the elements in $\mathbf{\Omega}$ and $\mathbf{W}$ are of size $d_h^2$, where $d_h$ is the hidden dimension. Since these matrices are symmetric, there is redundancy in the parameters, reducing the total number to approximately $\frac{1}{2}d_h^2$.\\ To summarize, the number of parameters for GCN scales as $\mathcal{O}(Ld_h^2)$. Since GCN shares a similar message-passing as Equation \eqref{eq: discrete_gradient_flow3}, but it does not use weight sharing, and we set $\mathbf{\Omega} \equiv 0$. In the case of GRAFF-LP, both $\mathbf{W}$ and $\mathbf{\Omega}$ use weight sharing, and the overall complexity scales as $\mathcal{O}(d_h^2)$.
In terms of time complexity, GRAFF-LP does not have any specific advantage over other models.
In Table \ref{tab:runtime}, we show the runtime analysis for all the models, reporting also the number of parameters.

\section{Experiments}
In this Section, we outline our experimental setup and results to address the following research questions:
\begin{description}
    \item[\textbf{RQ1:}] How the $f_{g}$ readout contribute to the GNNs performance? 
    \item[\textbf{RQ2:}] Can GRAFF-LP induce attraction and repulsion among existing and non-existing edges, resembling what is observed at node-level in node classification?
    \item[\textbf{RQ3:}] How much class heterophily impacts models' performances?
\end{description}
\begin{table}[t]
    \centering
    \caption{Dataset Information}
    \label{tab:datasets}
    \scalebox{1}{
    \begin{tabular}{lcccccc}
        \hline
        \textbf{Datasets} & $N$ & $|\mathcal{E}|$ & $d$ & $|C|$ & $\xi_{edge}$ & $\xi_{adj}$ \\
        \hline
        \textbf{Amazon Ratings}    & 24492 &186100 & 300 & 5& 0.38& 0.14\\
        \textbf{Roman Empire}      & 22662& 65854 & 300 & 18 & 0.05 & -0.05\\
        \textbf{Minesweepers}       & 10000 &78804  &7 & 2& 0.68 & 0.01\\
        \textbf{Questions}         & 48921 & 307080 & 301 & 2& 0.84& 0.02\\
        \textbf{Tolokers}         & 11758 & 519000 & 10 & 2& 0.59& 0.09\\
        
        \hline
    \end{tabular}}
\end{table}

\subsection{Experimental Set-up}
\subsubsection{Datasets}
Table \ref{tab:datasets} presents graph statistics after conversion to undirected graphs, a standard GNN procedure. 
The datasets selected belong to a recent collection \cite{criticallookatgnn}, which was proposed to enrich the current dataset availability for the GNN experimental setting under heterophily.  They have not yet been applied to link prediction. 
We used four datasets, \texttt{Amazon Ratings}, \texttt{Roman Empire}, \texttt{Minesweeper}, \texttt{Questions}, and \texttt{Tolokers}. Additional details and dataset descriptions can be found in Appendix \ref{sec:datasets}. These datasets differ in context, size, and structural properties \cite{criticallookatgnn}.\\
We did not consider datasets from the WebKB \cite{webkb2, geomgcn} collection, since they lead to unstable and statistically insignificant results as already noted in \cite{criticallookatgnn}, also in our experiments all the performances were not comparable, making it impossible to understand what model was more effective w.r.t. the others.
\subsubsection{Baselines} We compare against several baselines, including a Multi-layer Perceptron (MLP) using only node features, and both node-based and subgraph-based methods. \\ \textbf{GCN and GraphSAGE:} Convolutional MPNNs \cite{everythingisconnected}; for GraphSAGE, we evaluate both \textit{mean} and \textit{max} variants. \\ \textbf{GAT:} An attentional MPNN \cite{everythingisconnected} that uses an attention mechanism for neighbor aggregation, particularly effective for heterophilic graphs. \\ \textbf{ELPH: \cite{subgraphsketching}} A subgraph-based method for link prediction that avoids explicit subgraph computation, using a GCN-based feature extractor, though compatible with other MPNNs. \\ 
\textbf{NCNC: \cite{ncnc}} A link prediction method, based on structural features as ELPH, but these are related to higher-order common neighbor information. This method is currently state-of-the-art in a recent link prediction benchmark \cite{heart}.

We also intended to test \textbf{Disenlink} \cite{DisenLink}, a model designed for link prediction under heterophily, but the official implementation prevented its use on our datasets, because of their size. This was due to an inefficient implementation and use of the adjacency matrix.

Further experimental details are available in Appendix \ref{sec:experimentalsetup}.
\subsection{Results}
\subsubsection{Effectiveness}
\label{subsec:hyperparamters}
To evaluate model performance, we use the \textit{Area Under the Receiver Operating Characteristic} (AUROC). Table \ref{tab:combined_results} summarizes the results across all datasets for both readout functions, $f_{h}$ and $f_{g}$. The results with $f_h$ are related to the best configuration of hyperparameters with $f_{h}$. Concerning $f_g$, we just took the same configuration found with $f_h$ and trained the models again via $f_g$. Performance metrics are averaged over 10 random seeds. We adopted a single split since the graphs are sufficiently large and less sensitive to high variance in the data. The statistical significance of the results is computed through the Wilcoxon test \cite{wilcoxontest}, via the accuracy of positive and negative edges in the test set; details about the procedure can be found in the code. 

With $f_{h}$, GRAFF-LP leads in 3 out of 4 datasets and in 2 out of 4 using $f_{g}$, ranking consistently in the top 2 across all datasets, indicating adaptability not seen with the other models. These aspects also apply to \texttt{Tolokers}, whose results and comments are reported in Appendix \ref{sec:results}.

On the \texttt{Roman Empire} dataset, which resembles a chain-like graph, GRAFF-LP achieves the highest AUROC in both setups, with a neat advantage w.r.t. other baselines.
ELPH, and NCNC also present a wide gap w.r.t node-based methods, but they have access to the subgraph features that are a clear advantage to perform link prediction in such a structure. This advantage and superiority of ELPH and NCNC are found with \texttt{Minesweeper}, where the graph is a regular grid, and link prediction can be easily solved by understanding the grid structure. 

In both cases, the physics-inspired bias seems to let the model use the structure properly and achieve state-of-the-art performance. According to our hyperparameter optimization, ELPH fails to obtain competitive performance in \texttt{Amazon Ratings} and \texttt{Questions}. We relate this gap to the limited depth of ELPH (which is up to 3 layers). NCNC instead can reach competitive performance in \texttt{Amazon Ratings}, as well as \texttt{Questions}. Despite this, it never surpasses GRAFF-LP. We conclude that subgraph-based methods do not perform better than GRAFF-LP on class heterophilic datasets. This implies that structural features are not critical or determining to perform well in these settings.

Another observation is that GCN, both in \texttt{Amazon Ratings} and \texttt{Questions}, is among the top-scoring models, and in particular gets the closest to GRAFF-LP. This is not unexpected, since they share a similar message-passing and expressivity \cite{GCN, GRAFF2}. They mainly differ in the weight design. We associate these similarities to the same phenomenon observed by \cite{GRAFF2}, who observed that in homophilic node classification, the two models perform the same. We conjecture in this case that for \texttt{Amazon Ratings} and \texttt{Questions} the edge-wise repulsion is not required, thus GCN can afford similar results. We also report the results of the MLP. 
The MLP results consistently lag behind the GNN models, emphasizing the need for graph representation learning for effective link prediction in these new datasets. For example, in \texttt{Minesweeper}, MLP struggles to infer missing links due to its lack of graph context and the dataset’s grid structure.

\looseness=-1
\begin{table*}[t]
\centering
\caption{Performance of models across datasets with $f_h$ and $f_g$. Asterisks indicate statistical significance (p-value = \{* $\rightarrow$ 0.01, ** $\rightarrow$ 0.05, *** $\rightarrow$ 0.1\}). Text color refers to the \textcolor{red}{first}, \textcolor{blue}{second}, and \textcolor{MyDarkGreen}{third} model according to the mean.}
\resizebox{\textwidth}{!}{  
\begin{tabular}{lcccccccccc}
\toprule
\textbf{Datasets} & \multicolumn{2}{c}{\textbf{Amazon Ratings}} & \multicolumn{2}{c}{\textbf{Roman Empire}} & \multicolumn{2}{c}{\textbf{Minesweeper}} & \multicolumn{2}{c}{\textbf{Questions}} \\ \hline
\textbf{Models} & $f_h$ & $f_g$ & $f_h$ & $f_g$ & $f_h$ & $f_g$ & $f_h$ & $f_g$ \\ 
\midrule
MLP    & 66.80 $\pm$ 9.5*  & 70.18 $\pm$ 0.1*  & 64.74 $\pm$ 1.7*  & 64.60 $\pm$ 1.1* & 59.19 $\pm$ 1.8* & 47.26 $\pm$ 3.8* & 77.44 $\pm$ 0.7*  & 73.92 $\pm$ 3.9*  \\
GCN    & \textcolor{MyDarkGreen}{93.97 $\pm$ 0.8}*  & \textcolor{blue}{98.90 $\pm$ 0.2}*  & 51.32 $\pm$ 1.6* & 56.28 $\pm$ 4.1* & 94.44 $\pm$ 1.0*  & \textcolor{MyDarkGreen}{98.56 $\pm$ 0.3}* & \textcolor{blue}{97.63 $\pm$ 0.1}***  & \textcolor{red}{97.56 $\pm$ 0.1}  \\
SAGE   & 65.58 $\pm$ 22.5* & 69.24 $\pm$ 5.5*  & 65.17 $\pm$ 1.4*  & 66.57 $\pm$ 3.6* & \textcolor{MyDarkGreen}{97.99 $\pm$ 0.7}* & 96.40 $\pm$ 2.1*  & 91.48 $\pm$ 1.4*  & 93.20 $\pm$ 0.9*  \\
GAT    & 60.31 $\pm$ 6.0*  & 72.22 $\pm$ 6.1* & 71.33 $\pm$ 1.4*  & 73.85 $\pm$ 3.4* & 95.05 $\pm$ 2.8*  & 98.23 $\pm$ 0.7*  & 78.66 $\pm$ 9.7*  & 69.81 $\pm$ 3.6*  \\ \hline
ELPH   & 55.89 $\pm$ 12.5* & 55.89 $\pm$ 12.5* & \textcolor{blue}{87.01 $\pm$ 1.1}*  & \textcolor{blue}{87.01 $\pm$ 1.1}* & \textcolor{red}{99.88 $\pm$ 0.2}  & \textcolor{red}{99.88 $\pm$ 0.2}  & 82.13 $\pm$ 10.2* & 82.13 $\pm$ 10.2*  \\
NCNC & \textcolor{blue}{98.20 $\pm$ 1.3} & \textcolor{MyDarkGreen}{98.20 $\pm$ 1.3}* & \textcolor{MyDarkGreen}{86.73 $\pm$ 6.5}* & \textcolor{MyDarkGreen}{86.73 $\pm$ 6.5}* & 95.42 $\pm$ 11.4* & 95.42 $\pm$ 11.4* & \textcolor{MyDarkGreen}{94.61 $\pm$ 0.5}* & \textcolor{MyDarkGreen}{94.61 $\pm$ 0.5}* 
\\ \hline
GRAFF-LP & \textcolor{red}{98.69 $\pm$ 0.4}  & \textcolor{red}{99.47 $\pm$ 0.1} & \textcolor{red}{98.23 $\pm$ 0.8}  & \textcolor{red}{99.34 $\pm$ 0.4} & \textcolor{blue}{99.01 $\pm$ 0.4}  & \textcolor{blue}{99.41 $\pm$ 0.2}  & \textcolor{red}{97.64 $\pm$ 0.0}  & \textcolor{blue}{97.54 $\pm$ 0.1}  \\
\bottomrule
\end{tabular}  
}
\label{tab:combined_results}
\end{table*}

In Appendix \ref{sec:homophily_experiments}, we provide examples of GRAFF-LP performing on link prediction under homophily, to show that our approach can reach competitive as well as state-of-the-art performance, also under homophily.

In Table \ref{tab:improvements}, we show the relative percentage improvement we obtained using $f_g$ as the readout method. We did not include ELPH and NCNC, since their official derivation is based on Hadamard or other readouts.
\begin{table}[t]
    \centering
    
    \caption{Percentage Increase of Models Across Datasets, when going from $f_h$ to $f_g$.}
    \resizebox{\columnwidth}{!}{  
    \begin{tabular}{lcccc}
        \toprule
        \textbf{Model} & \textbf{A. Ratings} & \textbf{R. Empire} & \textbf{Minesweeper} & \textbf{Questions} \\
        \midrule
        \textbf{MLP}    & \textcolor{MyDarkGreen}{+4.48\%}    &   0\%     &  \textcolor{red}{-20.34\%}      &  \textcolor{red}{-3.90\%}      \\
        \textbf{GCN}    & \textcolor{MyDarkGreen}{+5.32\%}    &   \textcolor{MyDarkGreen}{+9.80\%}    &  \textcolor{MyDarkGreen}{+5.32\%}     &    0\%    \\
        \textbf{SAGE}   & \textcolor{MyDarkGreen}{+4.55\%}    &   \textcolor{MyDarkGreen}{+3.08\%}   &  \textcolor{red}{-2.04\%}      &   \textcolor{MyDarkGreen}{+2.20\%}    \\
        \textbf{GAT}    & \textcolor{MyDarkGreen}{+20\%}      &   \textcolor{MyDarkGreen}{+4.23\%}   &  \textcolor{MyDarkGreen}{+3.16\%}    & \textcolor{red}{-11.39\%}      \\ \hline
        \textbf{GRAFF-LP} &  \textcolor{MyDarkGreen}{+0.70\%}                       &   \textcolor{MyDarkGreen}{+1.02\%}   &     0\%   &    0\%    \\
        \bottomrule
    \end{tabular}}
    \label{tab:improvements}
\end{table}


\subsubsection{Physics-Inspired Link Prediction}
\begin{figure*}[t]
    \centering
    \begin{minipage}{0.22\linewidth}
        \centering
        \includegraphics[width=\linewidth]{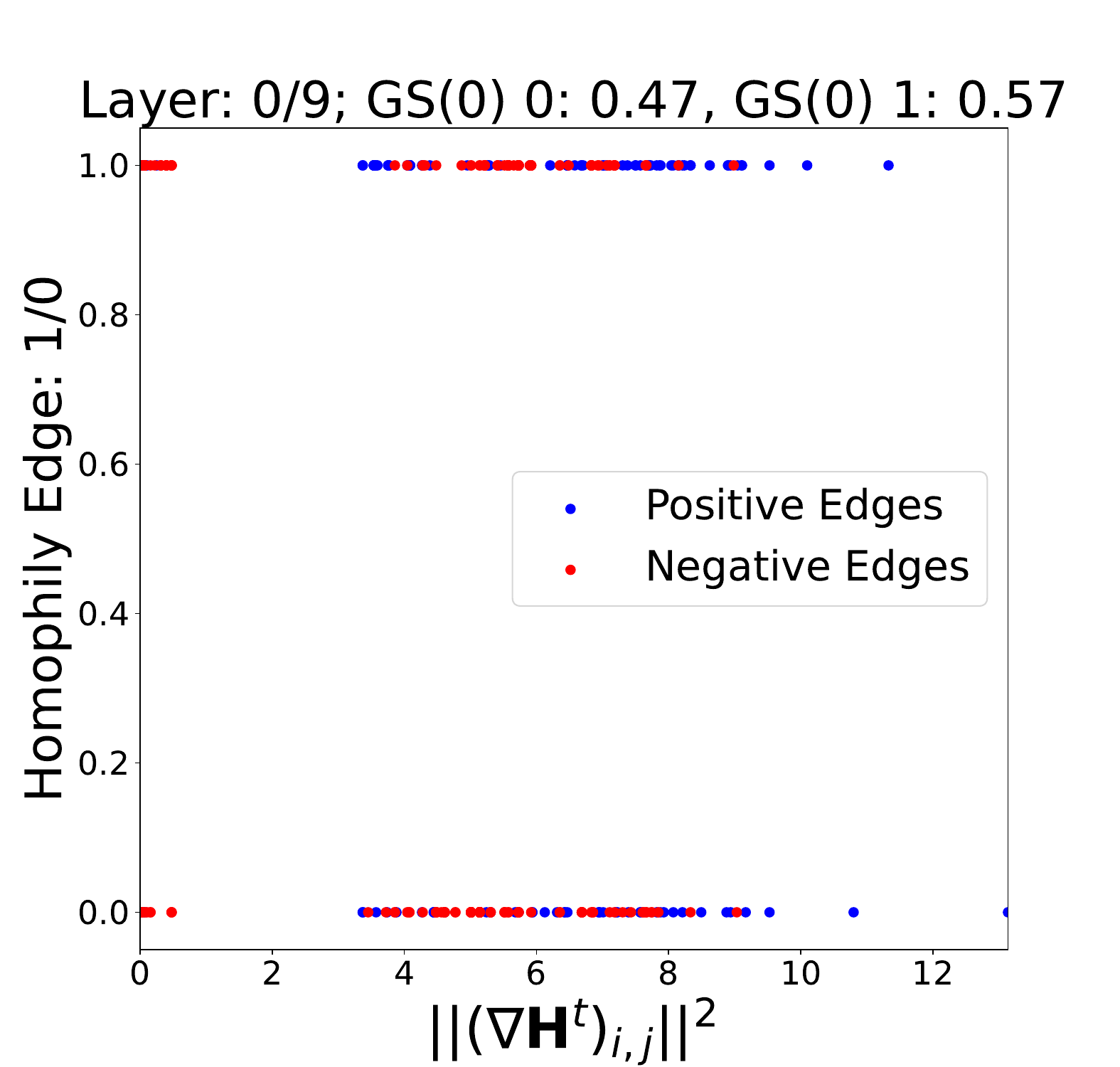}
        \small (a) $GS^0_{hm, hm}$, $GS^0_{ht, ht}$
    \end{minipage}
    \begin{minipage}{0.22\linewidth}
        \centering
        \includegraphics[width=\linewidth]{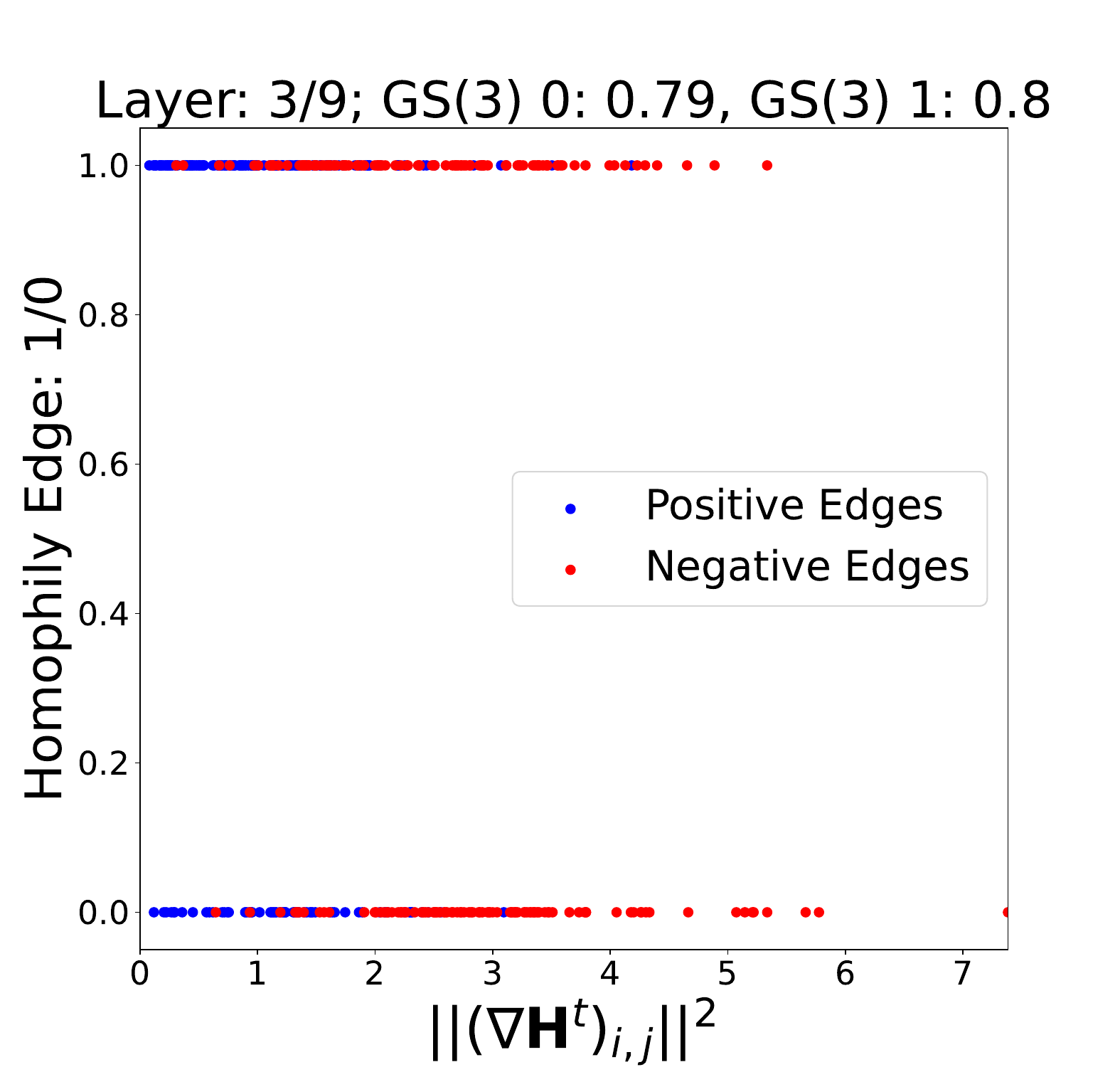}
        \small (b) $GS^3_{hm, hm}$, $GS^3_{ht, ht}$
    \end{minipage}
    \begin{minipage}{0.22\linewidth}
        \centering
        \includegraphics[width=\linewidth]{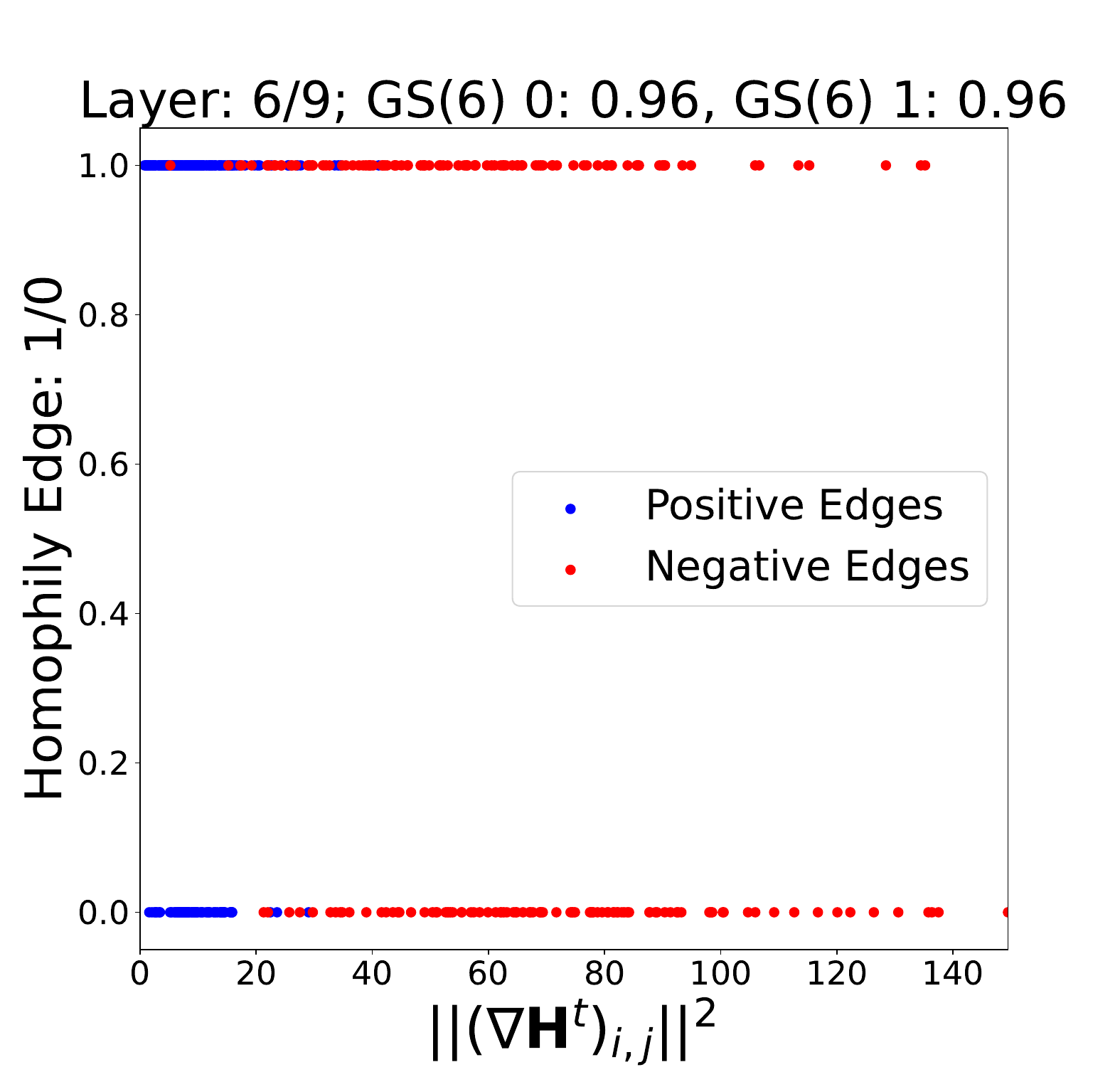}
        \small (c) $GS^6_{hm, hm}$, $GS^6_{ht, ht}$
    \end{minipage}
    \begin{minipage}{0.22\linewidth}
        \centering
        \includegraphics[width=\linewidth]{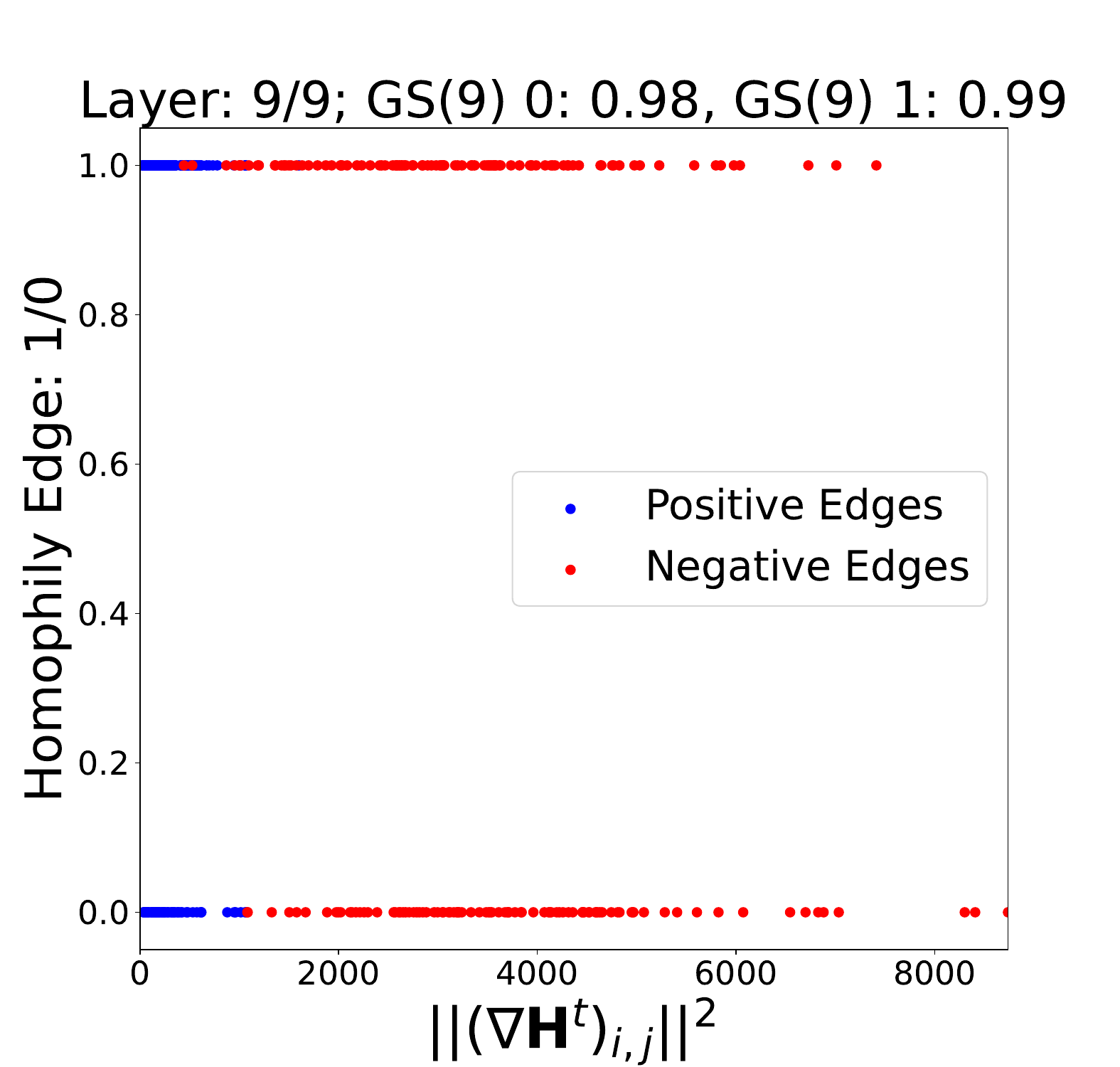}
        \small (d) $GS^9_{hm, hm}$, $GS^9_{ht, ht}$
    \end{minipage}
    \caption{$||(\nabla \mathbf{H}^t)_{i,j}||^2$ evolution with a fully-trained 9-layer GRAFF-LP via $f_g$ on \texttt{Minesweeper}.}
    \label{fig:attraction_repulsion_mines_g}
\end{figure*}

\begin{figure}[t]
    \centering
    \includegraphics[width=\linewidth]{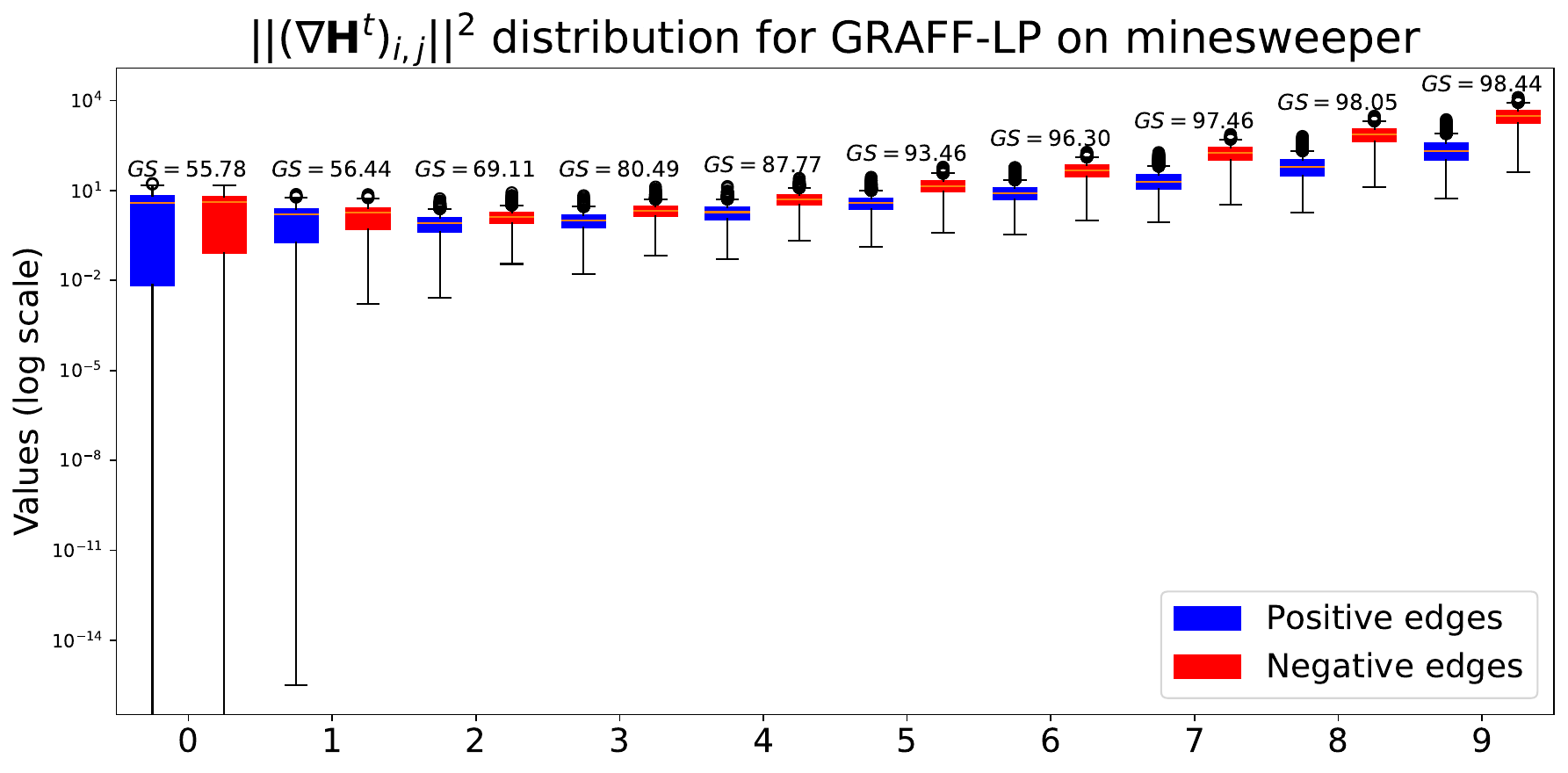}
        \caption{\texttt{Minesweeper}: Edge Gradients Distribution after each message-passing phase,  for a 9-layer GRAFF-LP.
 }
    \label{fig:boxplots_mines_grad}
 
\end{figure}

\cite{GRAFF2} proposed the GRAFF architecture and showed the attraction and repulsion behavior among adjacent nodes with different labels. This is expressed in Equation \eqref{eq:param_dirich2}. In the link prediction scenario, we believe that the same behavior can be induced among two nodes that, even though they seem different, should attract themselves, and vice versa, where two nodes are similar but do not present any interaction. 
Since we build upon the same framework of \cite{GRAFF2}, we expect the edge gradients associated with $\Theta_{+}$ to become smaller, because of attraction in the feature space, and the edge gradients associated with $\Theta_{-}$ to become larger in the feature space because of repulsion. In this way, we can deduce that GRAFF-LP can induce attraction and repulsion in the same fashion as in GRAFF for node classification, and we would answer positively to \textbf{RQ2}. To understand whether this phenomenon happens, we introduce a novel metric, namely the \textbf{gradient separability} ($GS$). Let us assume we have a set of edge gradients $\nabla = \{(\nabla\mathbf{H}^t)_{i,j} | 0 \leq t \leq T \wedge (i,j) \in \mathcal{E}_t\}$, where $\mathcal{E}_t = \{\mathcal{E}_{pos} \cup \mathcal{E}_{neg} \}$, is the set of evaluation edges, for example the test edges, which comprises both positive edges $\mathcal{E}_{pos}$, and negative edges $\mathcal{E}_{neg}$. Roughly speaking, $\nabla$ contains the edge gradients of positive and negative test edges computed at each layer of the model. This is done only for message-passing layers, excluding those in the Encoding and Decoding phases, to isolate the effect of the message-passing. We specify the squared norm of the positive (or negative) edge gradients at time t as follows: 
\begin{align}
    \nabla^t_{pos} &= \{ ||(\nabla \mathbf{H}^t)_{i,j}||^2 | (i,j) \in \mathcal{E}_{pos}\} \\
    \nabla^t_{neg} &= \{ ||(\nabla \mathbf{H}^t)_{i,j}||^2 | (i,j) \in \mathcal{E}_{neg}\} 
\end{align}
More formally, by minimizing \eqref{eq:param_dirich2}, we expect the squared norm of the positive (negative) gradients to decrease (increase). To measure this, we take advantage of the AUROC metric, considering the positives as class \texttt{0}, while the negatives as class \texttt{1}. Through the AUROC, we evaluate how much these two classes are separated based on their scores, regardless of any threshold. In particular if we consider as ground truth a stack of 1's and 0's for negatives and positives $\mathcal{S}_{\mathcal{E}_t} = \{\mathbb{1}_{\{(i,j) \in \mathcal{E}_{neg}\}} |\forall (i,j) \in \mathcal{E}_t\}$, $GS$ at time $t$ is computed as 
\begin{equation}
    GS^t = AUC(\mathcal{S}_{\mathcal{E}_t}, \{\nabla^t_{pos}\cup \nabla^t_{neg}\}).
\end{equation}
This way, we can monitor $GS^t$ over time to understand the trend of the edge gradients. We can also distinguish the edges connecting nodes of the same class, and those connecting edges from different ones. \\Let us define $\nabla^t_{pos} = \{\nabla^t_{pos, hm} \cup \nabla^t_{pos, ht}\}$, $\nabla^t_{neg} = \{\nabla^t_{neg, hm} \cup \nabla^t_{neg, ht}\}$, as the composition of the edge gradients coming from homophilic edges $\mathcal{E}_{hm} = \{(i,j) \in \mathcal{E} \wedge y_i = y_j\}$, and heterophilic ones $\mathcal{E}_{ht} = \{(i,j) \in \mathcal{E} \wedge y_i \neq y_j\}$, according to this distinction we can evaluate the subset of the edges based on their class labels. The gradient separability can be written accordingly: 
\begin{equation}
    GS^t_{\mathcal{U},\mathcal{V}} = AUC(\mathcal{S}_{\mathcal{U} \cup \mathcal{V}}, \{\nabla^t_{pos, \mathcal{U}}\cup \nabla^t_{neg, \mathcal{V}}\})
\end{equation}
$GS^t_{ht, hm}$ measures the ability to classify as \texttt{0} the positive heterophilic edges, and as \texttt{1} the negative homophilic edges. In other words, it measures how much we can separate the positive and the negative edges based on the squared norm of their edge gradient. 
We provide a visual example in Figure \ref{fig:attraction_repulsion_mines_g}, where we consider \texttt{1} on the y-axis as the homophilic edges, while the \texttt{0} y-axis refers to the heterophilic edges. We visualize a total of 50 samples, but the scores refer to those computed within the whole test set. These are produced from a fully-trained GRAFF-LP, monitoring $GS^t_{hm, hm}$ and $GS^t_{ht, ht}$. Figure \ref{fig:boxplots_mines_grad} also includes the distributions of the squared norm gradients to better understand how they separate. 
These visualizations, along with those in Appendix \ref{sec:additional_examples}, provide qualitative and quantitative evidence that GRAFF-LP induces edge attraction and repulsion via $f_g$ as the nodes go through the network's layers. We answer positively to \textbf{RQ2}.

In Table \ref{tab:gradient_separability_results}, instead we show $GS^T$, which is the total gradient separability computed after the last message-passing layer, for each model and averaged across 10 random seeds. We report the results and comments about \texttt{Tolokers} in Appendix \ref{sec:additional_results_tolokers}. In this analysis, we do not use any statistical test, since we do not want to assess which model has the highest $GS^T$; we want to understand what models have a sufficiently high value of $GS^T$ to acknowledge their ability to separate edge gradients. We consider $GS^T>90\%$ as evidence for the model to be able to distinguish the nature of the edges, based uniquely on the squared norm of the gradients. 
The most interesting aspect of $GS^T$ is the consistent increase that we record for GRAFF-LP, which we associate with the capability of this model to understand the negative and positive edges based on the evolution of their gradients. Indeed, obtaining $GS^T = 100\%$ implies that the edges can be classified as positive or not, based solely on $\{\nabla^T_{pos}, \nabla^T_{neg}\}$, which are available even before the Decoding phase. According to our results, GRAFF-LP allows us to do so even though it is not trained explicitly to do it. Equation \eqref{eq:param_dirich2} lets the model minimize the squared norm of the gradients, but it refers to the message-passing edges, not to the unseen edges that we want to predict. In the \texttt{Roman Empire} experiments, when we go from $f_h$ to $f_g$ the AUROC improvement is limited to $+1.02\%$ (see Table \ref{tab:improvements}), but when it comes to gauging the gradient separability difference, we have a $+65.52\%$. Since the only difference in the two models is the readout, we affirm that $f_g$ induces the model to learn to separate the edges following a strategy driven by the gradients. The only dataset where GRAFF-LP seems not to catch this behavior is \texttt{Questions}, where $GS^T$ is close to 0. Anyway, this is not an issue, since the AUROC is faulty when it approaches 50\%, in our case, it means that the negative edges decrease, rather than the positive ones. For this reason, we conjecture that GRAFF-LP is learning the opposite behavior. The same happens to GCN, which is even closer to 0 than GRAFF-LP since it reaches down to $GS^T = 14.32\%$. Surprisingly, not only is our model able to learn to separate the edge gradients, but also the other node-based methods, except for the MLP, ELPH, NCNC. Nonetheless, NCNC scores $GS^T>90$ in \texttt{Minesweeper}. Despite this, we recognize GRAFF-LP as the approach that consistently shows this behavior; we can conclude that this is due to the PIrd message-passing and readout schemes. ELPH reaches state-of-the-art performance in \texttt{Minesweeper}, leaving GRAFF-LP as the second best performing model, but our model offers more transparency in the model's behavior, as confirmed by $GS^T$. These results describe the full contribution of our newly introduced readout function $f_g$ answering \textbf{RQ1}. Specifically, $f_g$ can enhance the prediction performance (see Table \ref{tab:improvements}), and also the model's transparency at the inference phase, as illustrated in this Section.
\begin{table*}[t]
    \centering
    \caption{Comparison of $GS^T$, when the model is trained with $f_h$ or $f_g$. The score that is closest to 100 or 0 is highlighted in bold. We report the percentage variation between the results of $f_h$ and $f_g$ as $\Delta$.}
    \label{tab:gradient_separability_results}
    \resizebox{0.9\textwidth}{!}{
    \begin{tabular}{lcccccccccc}
        \toprule
        \textbf{Model} & \textbf{Readout Type} & \multicolumn{2}{c}{\textbf{Amazon Ratings}} & \multicolumn{2}{c}{\textbf{Roman Empire}} & \multicolumn{2}{c}{\textbf{Minesweeper}} & \multicolumn{2}{c}{\textbf{Questions}} \\
        \cmidrule(lr){3-4} \cmidrule(lr){5-6} \cmidrule(lr){7-8} \cmidrule(lr){9-10}
        & & $GS^T$ & $\Delta$ & $GS^T$ & $\Delta$ & $GS^T$ & $\Delta$ & $GS^T$ & $\Delta$ \\
        \midrule
        \multirow{2}{*}{\textbf{MLP}}    
        & \( f_h \) & 48.49 ± 14 & - & 35.97 ± 0.42 & - & 63.56 ± 1.6 & - & 47.65 ± 9.93 & - \\
        & \( f_g \) & 30.04 ± 0.06 & \textcolor{red}{-37.5\%} & 36.92 ± 1.7 & \textcolor{MyDarkGreen}{+2.78\%} & 62.04 ± 2.4 & \textcolor{red}{-3.13\%} & 43.18 ± 4.6 & \textcolor{red}{-10.42\%} \\
        \midrule
        \multirow{2}{*}{\textbf{GCN}}    
        & \( f_h \) & 71.9 ± 2 & - & 35.78 ± 0.2 & - & 76.7 ± 2.1 & - & 15.57 ± 2.5 & - \\
        & \( f_g \) & 86.32 ± 3.9 & \textcolor{MyDarkGreen}{+19.44\%} & 35.79 ± 0.2 & 0\% & 94.44 ± 1.1 & \textcolor{MyDarkGreen}{+22.08\%} & \textbf{14.32 ± 1.9} & \textcolor{red}{-12.5\%} \\
        \midrule
        \multirow{2}{*}{\textbf{SAGE}}    
        & \( f_h \) & 30.12 ± 0.05 & - & 37.26 ± 1.5 & - & 77.69 ± 1.6 & - & 63.31 ± 1.7 & - \\
        & \( f_g \) & 30.11 ± 0.06 & 0\% & 40.56 ± 3.6 & \textcolor{MyDarkGreen}{+10.81\%} & 68.81 ± 1.4 & \textcolor{red}{-11.54\%} & 63.41 ± 0.9 & 0\% \\
        \midrule
        \multirow{2}{*}{\textbf{GAT}}     
        & \( f_h \) & 33.45 ± 4.3 & - & 42.66 ± 4.8 & - & 93.22 ± 1.7 & - & 39.14 ± 3.9 & - \\
        & \( f_g \) & 31.13 ± 2.5 & \textcolor{red}{-6.06\%} & 46.24 ± 2.9 & \textcolor{MyDarkGreen}{+6.98\%} & 97.45 ± 1.01 & \textcolor{MyDarkGreen}{+4.3\%} & 37.42 ± 1.07 & \textcolor{red}{-5.13\%} \\
        \midrule
        \textbf{ELPH}
        & \( f_h \) & 46.76 ± 7.3 & - & 36.16 ± 0.11 & - & 73.44 ± 5.7 & - & 31.92 ± 8.08 & - \\
        \midrule
        \textbf{NCNC}
        & \( f_h \) & 30.49 ± 0.04 & - & 36.55 ± 0.23 & - & 91.31 ± 0.99 & - & 37.44 ± 1.08 & - \\
        \midrule
        \multirow{2}{*}{\textbf{GRAFF-LP}}   
        & \( f_h \) & 94.52 ± 2.04 & - & 58.46 ± 2.7 & - & 93.59 ± 2.13 & - & 26.74 ± 3.71 & - \\
        & \( f_g \) & \textbf{96.43 ± 0.9} & \textcolor{MyDarkGreen}{+1.05\%} & \textbf{95.96 ± 2.4} & \textcolor{MyDarkGreen}{+65.52\%} & \textbf{98.10 ± 0.32} & \textcolor{MyDarkGreen}{+4.26\%} & 18.20 ± 3.5 & \textcolor{red}{-33.33\%} \\
        \bottomrule
    \end{tabular}}
\end{table*}
\subsubsection{Runtime Analysis}
We briefly show some results about the time and space complexities that these baselines have. We already discussed space complexity in terms of parameters, but in this Section, we report the real-world values associated with the experiments. In Table \ref{tab:runtime}, we show the number of parameters of each model and the inference time, computed on the test edges for 10 epochs. All the models have $L = 3$, $L_{MLP} = 1$, and $d_h = d_{MLP} = 64$, to make a fair comparison across models. What we notice is that according to our discussion on complexity, GRAFF-LP is the lightest model, and also comparable with the other node-based methods such as GCN, GAT and GraphSAGE. ELPH, which is a subgraph-based method has the highest inference time, as expected. While NCNC, even though it is known to be more efficient than ELPH, and more similar to the node-based methods, it still presents a significant gap in terms of inference, due to the computation of structural features. Furthermore, it leads to out-of-memory issues in \texttt{Tolokers}. In Appendix \ref{sec:extended_runtime}, we report additional results on runtime analysis.
\begin{table}[t]
    \centering
    \caption{Comparison of model performance across datasets, showing the number of parameters and runtime (in seconds) for each model. The inference time is averaged across 10 trials.}
    \label{tab:runtime}
    \resizebox{0.48\textwidth}{!}{
    \begin{tabular}{lcccccccc}
        \toprule
        \multirow{2}{*}{\textbf{Model}} & \multicolumn{2}{c}{\textbf{Amazon Ratings}} & \multicolumn{2}{c}{\textbf{Minesweeper}} \\
        \cmidrule(lr){2-3} \cmidrule(lr){4-5}
        & \textbf{Parameters} & \textbf{Runtime (s)} & \textbf{Parameters} & \textbf{Runtime (s)} \\
        \midrule
        \textbf{MLP}                & 32256 & $0.0909 \pm 0.01$ & 13504 & $0.0512 \pm 0.01$ \\
        \textbf{GCN}                & 31680 & $0.118 \pm 0.01$ & 12928 & $0.0832 \pm 0.01$ \\
        \textbf{GAT}                & 32064 & $0.1079 \pm 0.01$ & 13312 & $0.0727 \pm 0.01$ \\
        \textbf{SAGE}               & 43968 & $0.0919 \pm 0.02$ & 25216 & $0.0818 \pm 0.01$ \\
        \textbf{ELPH}               & 40542 & $0.8751 \pm 0.05$ & 21790 & $0.4563 \pm 0.03$ \\
        \textbf{NCNC}               & 27584 & $0.1935 \pm 0.02$ & 8832  & $0.1143 \pm 0.01$ \\
        \textbf{GRAFF-LP} (\( f_h \)) & 23617 & $0.1072 \pm 0.01$ & 4865  & $0.0756 \pm 0.01$ \\ 
        \textbf{GRAFF-LP} (\( f_g \)) & 23617 & $0.1021 \pm 0.01$ & 4865  & $0.0796 \pm 0.01$ \\
        \bottomrule
    \end{tabular}}
\end{table}
\subsubsection{Does Class Heterophily Impact GNN performance?}
\begin{figure}[t]
    \centering
    \includegraphics[width=0.8\linewidth]{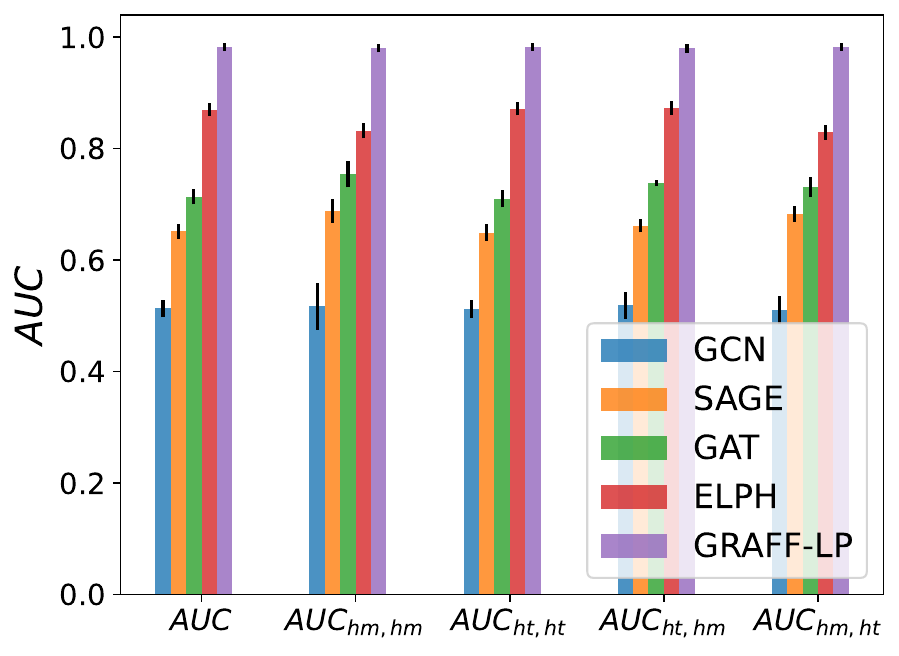}
    \caption{\texttt{Roman Empire}: Ability of the model to predict homophilic edges or heterophilic both as negatives or positives.}
    \label{fig:roman_empire_aurocs}
\end{figure}

In this paper's premises, we questioned the absence of explicit methods that tackle heterophily in link prediction. This is a gap in the literature that is worth studying since we have seen in the previous Sections, that in some datasets, even subgraph-based approaches fail to achieve competitive performance. Other than the mere performance, in the literature is pointed out the complicated nature of the task \cite{DisenLink}, namely, how can the latent factors that connect two entities be understood? In this paper, we focused on datasets that are known to be heterophilic for node classification. Are they representative enough for link prediction under heterophily, even if $\xi_{adj}$ is determined only from class labels? Do simple GNN baselines struggle to learn from these graphs, in the same fashion as node classification? In other words, can they learn to predict existing(non-existing) links when two entities have different(same) classes?
These concerns are related to $\textbf{RQ3}$. To answer it, we measure the AUROC that we obtain when we try to classify all the possible class mixes, namely when the positive or negative edges are homophilic or heterophilic. In this way, it is possible to measure 4 different AUROCs that are representative of the model's ability to separate edges of different natures. \\
Let us define $AUC_{\mathcal{U},\mathcal{V}}$ as the ability of a binary classifier to correctly distinguish the sets $\mathcal{U}$ and $\mathcal{V}$ as existing or non-existing edges. Here, the ground truth is that $\mathcal{U}$ contains positives and $\mathcal{V}$ the negatives. The nature of these can be homophilic or heterophilic, namely $\mathcal{U},\mathcal{V} \in \{hm, ht\}$. In Figure \ref{fig:roman_empire_aurocs}, we take into consideration \texttt{Roman Empire} as an example of a dataset where the models fail to achieve high performance, and we show that is not due to heterophilic edges (i.e. $AUC_{ht, hm}, AUC_{ht, ht}$), since we have poor performance also on the homophilic ones $AUC_{hm, ht}, AUC_{hm, hm}$. In Appendix \ref{sec:additional_robustness_to_heterophily_edge}, we provide additional evidence of this behavior, which also manifests when models reach high performance. Generally, we understand that the performance does not vary significantly, and the models' rankings remain consistent. We conclude that models that achieve competitive performances in homophilic link prediction do not perform as well in our benchmark, but this is not due to class heterophily. This result answers negatively to \textbf{RQ3} and highlights the need for new homophilic measures that can better identify datasets representative of link prediction under heterophily, because of its relevance in several applications.
\section{Conclusions}
This paper presented GRAFF-LP, a link prediction model built upon GNNs and driven by a PIrd bias. Our method is designed to handle heterophilic graphs, enriching the current literature available on the topic, which is often confined to node classification. We show that GRAFF-LP outperforms the other baselines in 3 out of 4 datasets. We also propose $f_g$, a PIrd readout function based on the edge gradients, that, without any tuning, is able to improve the GRAFF-LP performance. Surprisingly, we observe that other models also benefit from $f_g$, but GRAFF-LP consistently ranks among the top 2. To better understand the effect of $f_g$, we study the evolution of the squared norm of the edge gradients via a novel metric named $GS$, and we surprisingly found that, under the lens of this norm, the GRAFF-LP's message-passing separates the positive and negative edges. GRAFF-LP showcases this behavior more consistently than the other baselines, favored by the attraction and repulsion bias of the gradient flow interpretation. We also brought some evidence that the baselines do not struggle to perform link prediction because of class heterophily. This result exposes the need for new measures that help identify challenging scenarios to advance our knowledge of link prediction under heterophily. 
Future works should address the aspects related to new heterophily metrics based on the input features since they are not task-dependent. Moreover, we plan to further understand how GRAFF-LP can learn the edge gradient separation, even though it was not trained explicitly to do so. Then, it would be interesting to adapt other PIrd GNNs or develop new ones specifically suited for link prediction.

\bibliographystyle{IEEEtran} 
\bibliography{main}
\onecolumn

\appendix

\section*{Overview of the Appendix}  

To guide the reader through the appendix, we outline the content and structure below.
\begin{itemize}
    \item \textbf{\hyperref[sec:datasets]{Dataset Descriptions.}}
    We describe the motivations behind the datasets choice, highlighting their characteristics, and the edge distributions both for the negatives and positives.
    
    \item \textbf{\hyperref[sec:experimentalsetup]{Experimental Set-up: Additional Details.}}
    Here, we describe additional details on our experiments, such as the hyperparameter space, and the details related to the hardware that we used and also on the optimization strategy.
    
    \item \textbf{\hyperref[sec:results]{Additional Results.}}
    This section presents extended results from our experiments, including several examples of attraction and repulsion among gradients, and also bar plots to evaluate how the models predict heterophilic edges w.r.t. homophilic ones.
    
    \item \textbf{\hyperref[sec:extended_runtime]{Extended Runtime Analysis.}}
    This section presents extended results from the runtime analysis. We include those datasets missing from the main manuscript, showing a consistent behavior w.r.t. that already discussed.
    
    \item \textbf{\hyperref[sec:relatedworks]{Extended Related Works.}}
    Here we include a thorough discussion of link prediction methods, describing how the state-of-the-art practices have evolved over the years.
\end{itemize}

\section*{Datasets Description}
\label{sec:datasets}

In this Section of the appendix, we provide information on the datasets that we used, and how they have been split for training and evaluation.\\ 
We first start with a description of the datasets, which also helps in understanding why they are heterophilic graphs. \\
\textbf{\texttt{Amazon Ratings}:} Nodes represent products, with edges linking items frequently bought together. Node classes denote product ratings. The goal in link prediction is to anticipate likely co-purchases. \\ 
\textbf{\texttt{Roman Empire}:} This graph is built from the Roman Empire's Wikipedia article \cite{wikipedia_dump}, with nodes as non-unique words. Links exist if words are connected in dependency trees or appear sequentially in text, forming a nearly chain-like structure with shortcuts. Node classes indicate syntactic roles, and link prediction involves reconstructing the chain structure and syntactic dependencies. \\ 
\textbf{\texttt{Minesweeper}:} This synthetic 100x100 grid has cells as nodes, each linked to at most 8 neighbors. Cells are classified as mines or traversable, and link prediction aims to capture the grid structure amid varying cell types. \\ \textbf{\texttt{Questions}:} Based on a Q\&A website, nodes are users connected by interaction over time. Users are classified as active or inactive. Link prediction consists of predicting what users are likely to interact with.\\
\textbf{\texttt{Tolokers}:} This dataset is based on data from a crowdsourcing platform. Nodes represent workers who have participated in at least one of 13 selected
projects. Labels for each node are binary, identifying what workers have been banned from a project. An edge connects two tolokers if they have worked on the same task. Thus, link prediction has the objective to predict which workers will likely collaborate.

The modality of the experiments follows the transductive link prediction paradigm. Graphs have been split into training, validation, and test positive edges ($N_{pos} $) with 80\%, 10\%, and 10\% percentages. Negative edges ($N_{neg}$) for each split were also sampled, and in the experiments, we adopted several ratios $N_{neg}/N_{pos}$ since there is not a specific policy to follow when selecting $N_{neg}$ \cite{negativeseffect}. We select negative edges through the random negative sampling routine implemented by PyTorch Geometric \cite{pytorchgeometric}.
As concerns, the number of evaluation edges, and those contained in the training, validation, and test sets, Table \ref{tab:edge_statistics} showcases the proportion of the edges in our experimental set-up.
\begin{table}[htbp]
    \centering
    \begin{adjustbox}{width = \textwidth}  
    \begin{tabular}{l l c c c c}
        \toprule
        \textbf{Dataset}               & \textbf{Splits} & \textbf{Message Passing Edges} & \textbf{Positive Edges} & \textbf{Negative Edges} \\ 
        \midrule
        \textbf{Minesweeper}        & \textbf{train}  & 50,436               & 6,304          & 1,260,800      \\
                                     & \textbf{val}    & 63,044               & 3,940          & 777,933        \\
                                     & \textbf{test}   & 70,924               & 3,940          & 771,958        \\
        \midrule
        \textbf{Amazon Ratings}    & \textbf{train}  & 119,104              & 14,888         & 2,977,600      \\
                                    & \textbf{val}    & 148,880              & 9,305          & 1,851,634      \\
                                    & \textbf{test}   & 167,490              & 9,305          & 1,845,817      \\
        \midrule
        \textbf{Questions}          & \textbf{train}  & 196,532              & 24,566         & 4,913,200      \\
                                     & \textbf{val}    & 245,664              & 15,354         & 3,064,559      \\
                                     & \textbf{test}   & 276,372              & 15,354         & 3,060,429      \\
        \midrule
        \textbf{Roman Empire}      & \textbf{train}  & 42,150               & 5,268          & 1,053,600      \\
                                    & \textbf{val}    & 52,686               & 3,292          & 657,070        \\
                                    & \textbf{test}   & 59,270               & 3,292          & 656,211        \\
        \midrule
        \textbf{Tolokers}      & \textbf{train}  & 664,320             &  83,040        & 16,608,000      \\
                                    & \textbf{val}    & 830,400               & 51,900          & 10,380,000        \\
                                    & \textbf{test}   & 934,200              & 51,900        & 10,380,000        \\
        \bottomrule
    \end{tabular}
    \end{adjustbox}
    \caption{Dataset Statistics for Message Passing Edges, Positive Edges, and Negative Edges.}
    \label{tab:edge_statistics}
\end{table}
We defined the split to avoid data leakage among the edges in the set, in particular, we followed the specifics established by \cite{link_prediction_pitfalls}. The negatives were sampled more, as we see in Table \ref{tab:edge_statistics}, since we use the number of negatives as a hyperparameter. Regarding the evaluation edges, we ensured an equal balance of negative and positive samples. The total number of edges and how they differ in terms of homophilic and heterophilic is displayed in Table \ref{tab:eval_edges}.
\begin{table}[htbp]
    \centering
    \begin{adjustbox}{width = \textwidth}  
    \begin{tabular}{l c c c c c} 
        \toprule
        \textbf{Dataset}     & \textbf{Positives $\mathcal{E}_{hm}$} & \textbf{Positives $\mathcal{E}_{ht}$} & \textbf{Negatives $\mathcal{E}_{hm}$} & \textbf{Negatives $\mathcal{E}_{ht}$} & \textbf{$|\mathcal{E}_{pos}| = |\mathcal{E}_{neg}| $}\\ 
        \midrule
        \textbf{Amazon Ratings} & 3,507  & 5,798  & 2,483  & 6,822 & 9,305 \\
        \textbf{Roman Empire}   & 122    & 3,170  & 301    & 2,991 & 3,292 \\
        \textbf{Minesweeper}    & 2,672  & 1,268  & 2,534  & 1,406 & 3,940 \\
        \textbf{Questions}      & 12,906 & 2,448  & 14,456 & 898  & 15,354  \\
        \textbf{Tolokers}      & 30,731 & 21,169  & 33,354 & 18,546 & 51,900  \\
        
        \bottomrule
    \end{tabular}
    \end{adjustbox}
    \caption{Statistics for Positives and Negatives in $\mathcal{E}_{hm}$ and $\mathcal{E}_{ht}$ categories. This refers to the test edges for which we report the main results in the paper.}
    \label{tab:eval_edges}
\end{table}

\section*{Experimental Set-up: Additional Details}
\label{sec:experimentalsetup}
In this Section, we report additional details on the experimental setup, such as the metrics, optimization algorithm, and the hyperparameters.
\subsection{Implementation Details}
We chose AUROC since it is commonly used in link prediction tasks with GNNs \cite{SEAL, NESS}. We trained them using negative log-likelihood. The loss function is optimized using Adam \cite{adam} with early stopping \cite{earlystopping}, with patience of 300 epochs, monitoring the AUROC. The experiments were run on a single Nvidia GeForce RTX 3090 Ti 24 GB.
\subsection{Hyperparameters}
We considered several hyperparameters in our model, including the learning rate $\alpha$, weight decay $\gamma$, and hidden dimension $d_h$, which was kept constant across all layers during the message-passing phase. We also adjusted the hidden dimension of the decoder $d_{MLP}$ for the decoding phase. The dropout rates for the encoding and decoding phases, denoted as $\rho$ and $\rho_{MLP}$ respectively, were optimized. Additionally, we tuned the number of layers for message passing $L$ and decoding $L_{MLP}$, examined the use of batch normalization in the decoder, and considered the ratio of negative to positive samples $\frac{N_{neg}}{N_{pos}}$. In the GRAFF-LP experiment, we also considered the value of the step size $\tau$. Our hyperparameter space is reported in Table \ref{tab:hyperparameters}.
\begin{table}[htbp]
    \centering
    \begin{tabular}{l r}
        \toprule
        \textbf{Hyperparameter}        & \textbf{Value} \\ 
        \midrule
         $\alpha$         & \{0.01, 0.001\}          \\
        $\gamma$          & \{0, 0.01, 0.001\}         \\
        $d_h$         & \{128, 256\}            \\
        $d_{MLP}$     &  \{32, 64\}             \\
        $\rho$   &  \{0.1, 0.3, 0.5\}             \\
        $\rho_{MLP}$ & \{0.1, 0.3, 0.5\}        \\
        $L$           & \{1, 3, 5, 7, 9, 12\}              \\
        $L_{MLP}$     & \{0, 1, 2\}             \\
        Batch Norm.  & \{Yes, No\}            \\
        $\frac{N_{neg}}{N_{pos}}$ & \{0.25, 0.5, 1, 2, 4, 8\}             \\
        $\tau$               & \{0.1, 0.25, 0.5\}           \\
        \bottomrule
    \end{tabular}
    \caption{Hyperparameter Space for Experiments.}
    \label{tab:hyperparameters}
\end{table}

\section*{Additional Results}
\label{sec:results}

\subsection{Additional results on \texttt{Tolokers}}
\label{sec:additional_results_tolokers}
\begin{figure}
    \centering
    \caption{\texttt{Questions}: Edge Gradients Distribution, at the end of each message-passing phase of a 3-layer fully-trained GRAFF-LP via $f_g$.}
    \includegraphics[width=\linewidth]{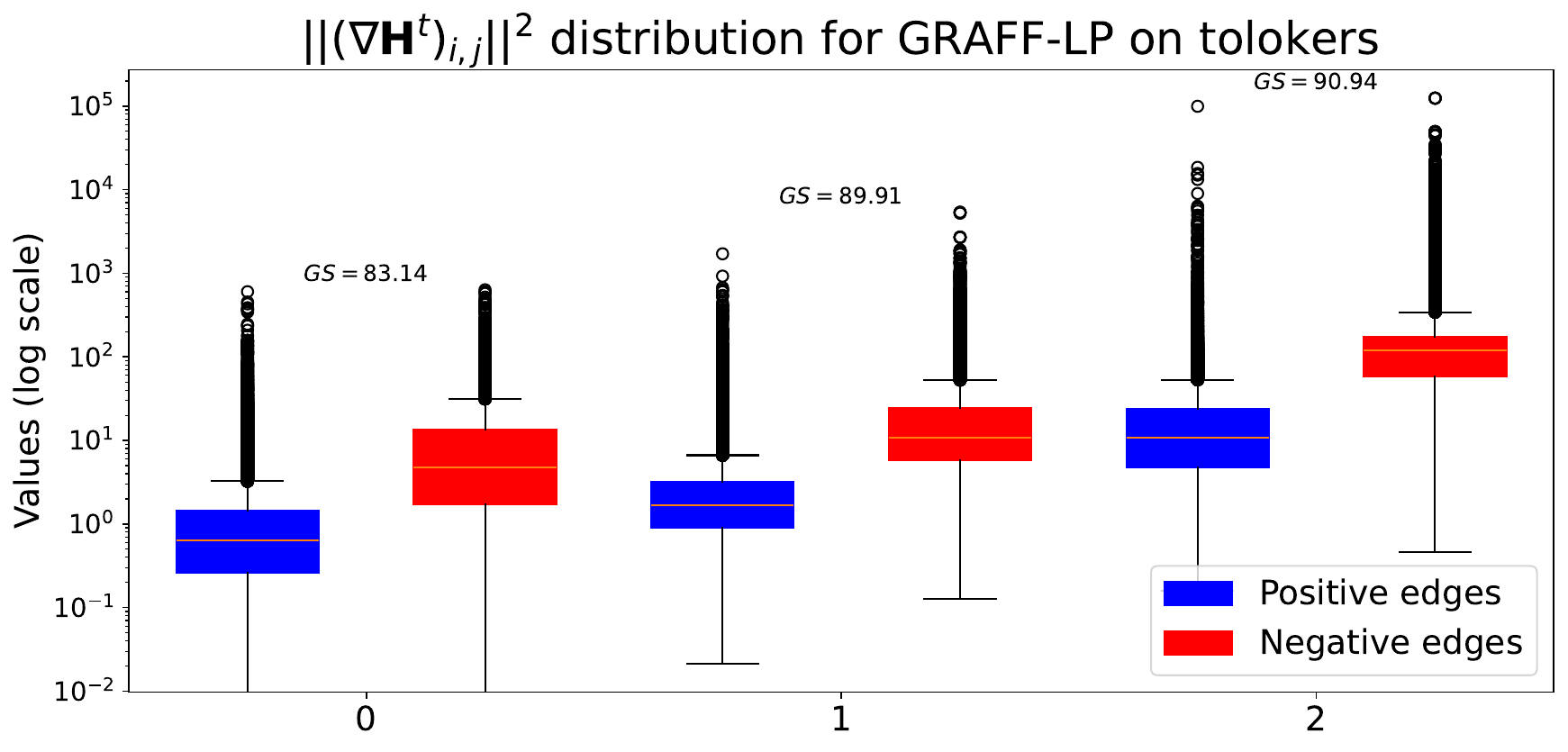}
    \label{fig:boxplots_tolokers_grad}
\end{figure}

For space constraints, we report here the AUROC performance for \texttt{Tolokers} in Table \ref{tab:tolokers_results}. We can see that GNN models surpass the MLP performance, underlining the informative nature of edges in \texttt{Tolokers} for the link prediction task. GRAFF-LP ranks among the top-3 models coherently with the other datasets. However, here GraphSAGE can outperform the other baselines.\\
In Table \ref{tab:tolokers_gradient_separability}, we also report the gradient separability performance. In this case, all the models have high gradient separability, and using our readout improves such separability measure, even though in \texttt{Tolokers}, the performance with the gradient readout is slightly worse than the Hadamard product. However, we use this metric to understand whether GNNs are learning to separate edge gradients, which is confirmed by the consistently high value of $GS^T$, and also the gradient separability trend of the test edges shown in Figure \ref{fig:boxplots_tolokers_grad}. We can see that even with 3 layers, GRAFF-LP improves the separability in terms of gradients. 
\begin{table*}[h!]
\centering
\caption{Performance of models on the \texttt{Tolokers} dataset with $f_h$ and $f_g$. Asterisks indicate statistical significance (p-value = \{* $\rightarrow$ 0.01, ** $\rightarrow$ 0.05, *** $\rightarrow$ 0.1\}). Text color refers to the \textcolor{red}{first}, \textcolor{blue}{second}, and \textcolor{MyDarkGreen}{third} model according to the mean. OOM means out of memory.}
\resizebox{0.6\textwidth}{!}{
\begin{tabular}{lcc}
\toprule
\textbf{Models} & $f_h$ & $f_g$ \\ 
\midrule
MLP      & 92.97 $\pm$ 1.14$^*$ & 92.37 $\pm$ 0.73$^*$ \\
GCN      & \textcolor{MyDarkGreen}{98.13 $\pm$ 0.28}$^*$ & \textcolor{blue}{97.95 $\pm$ 0.17} \\
SAGE     & \textcolor{red}{98.60 $\pm$ 0.26} & \textcolor{red}{98.73 $\pm$ 0.19} \\
GAT      & 96.97 $\pm$ 0.32$^*$ & 96.50 $\pm$ 0.14$^*$ \\ \hline
ELPH     & 90.09 $\pm$ 0.81$^*$ & - \\
NCNC     & OOM & - \\ \hline
GRAFF-LP & \textcolor{blue}{98.24 $\pm$ 0.19} & \textcolor{MyDarkGreen}{97.76 $\pm$ 0.03} \\
\bottomrule
\end{tabular}}  
\label{tab:tolokers_results}
\end{table*}
\begin{table*}[ht]
    \centering
    \caption{Comparison of $GS^T$ on the Tolokers dataset, when the model is trained with $f_h$ or $f_g$. We report the percentage variation between the results of $f_h$ and $f_g$ as $\Delta$. OOM means out of memory.}
    \label{tab:tolokers_gradient_separability}
    \resizebox{0.6\textwidth}{!}{
    \begin{tabular}{lccc}
        \toprule
        \textbf{Model} & \textbf{\( f_h \)} & \textbf{\( f_g \)} & \textbf{\(\Delta\)} \\
        \midrule
        \textbf{MLP}      & 87.50 ± 0.54   & 91.74 ± 0.76   & \textcolor{MyDarkGreen}{+4.55\%} \\
        \textbf{GCN}      & 90.52 ± 0.34  & 92.01 ± 0.28   & \textcolor{MyDarkGreen}{+1.64\%} \\
        \textbf{SAGE}     & 90.52 ± 0.20  & 92.22 ± 0.26 & \textcolor{MyDarkGreen}{+1.87\%} \\
        \textbf{GAT}      & 89.12 ± 0.56  & 88.61 ± 0.0038   & \textcolor{red}{-0.57\%} \\
        \textbf{ELPH}     & 83.73 ± 0.65  & -         & -\\
        \textbf{NCNC}     & OOM        & -         & - \\
        \textbf{GRAFF-LP} & 90.27 ± 0.33  & 90.69 ± 0.14   & \textcolor{MyDarkGreen}{+0.46\%} \\
        \bottomrule
    \end{tabular}}
\end{table*}

\subsection{Additional results on Homophilic Datasets}
\label{sec:homophily_experiments}
For the sake of completeness, we decided to assess how GRAFF-LP ranks in the recently proposed HeaRT benchmark \cite{heart}. This is a benchmark for link prediction under homophily, where the positive and negative edges used in the evaluation are chosen in a way that simple heuristics may fail to predict them, by sampling hard positives and negatives. This benchmark sheds light on how node-based methods are not significantly worse than subgraph-based methods or those approaches based on structural features. This gap reduction underlines how the datasets that are typically used in benchmarks do not require advanced methods like NCNC or ELPH for high performance, which is also what we observed in our experiments.
We used the same hyperparameter set proposed in \cite{heart}, and the best configuration result on the test set is presented in Table \ref{tab:results_homophilic}. We report the original Table from \cite{heart}, including also GRAFF-LP modalities (i.e., $f_h$ and $f_g$). In this benchmark, we have additional baselines, and those that we implemented in the heterophilic experiments, namely GCN, GAT and GraphSAGE, are implemented differently from ours. Details on implementation can be found in our code, as well as in the HeaRT repository, to reproduce their experiments as well.
In Table \ref{tab:results_homophilic}, we can also see that within homophilic link prediction, GRAFF-LP achieves competitive performance, specifically in PubMed, where it sets the new state-of-the-art. While in the other datasets, GRAFF-LP ranks in the top-3 only in Citeseer, leaving a significant gap in Cora. Further research is required to better understand the meaning of homophily in link prediction and how edge gradients separate in these settings.
\begin{table}[h!]
\centering
\caption{Results on Cora, Citeseer, and Pubmed (\%) under HeaRT. Highlighted are the results ranked \textbf{\textcolor{red}{first}}, \textbf{\textcolor{blue}{second}}, and \textbf{\textcolor{MyDarkGreen}{third}}.}
\label{tab:results_homophilic}
\scalebox{1}{
\begin{tabular}{lcccccc}
\toprule
\multirow{2}{*}{Models} & \multicolumn{2}{c}{Cora} & \multicolumn{2}{c}{Citeseer} & \multicolumn{2}{c}{Pubmed} \\
\cmidrule{2-7}
& MRR & Hits@10 & MRR & Hits@10 & MRR & Hits@10 \\
\midrule
\textbf{Heuristic} & & & & & & \\
CN & 9.78 & 20.11 & 8.42 & 18.68 & 2.28 & 4.78 \\
AA & 11.91 & 24.10 & 10.82 & 22.20 & 2.63 & 5.51 \\
RA & 11.81 & 24.48 & 10.84 & 22.86 & 2.47 & 4.90 \\
Shortest Path & 5.04 & 15.37 & 5.83 & 16.26 & 0.86 & 0.38 \\
Katz & 11.41 & 22.77 & 11.19 & 24.84 & 3.01 & 5.98 \\
\midrule
\textbf{Embedding} & & & & & & \\
Node2Vec & 14.47 $\pm$ 0.60 & 32.77 $\pm$ 1.29 & 21.17 $\pm$ 1.01 & 45.82 $\pm$ 2.01 & 3.94 $\pm$ 0.24 & 8.51 $\pm$ 0.77 \\
MF & 6.20 $\pm$ 1.42 & 15.26 $\pm$ 3.39 & 7.80 $\pm$ 0.79 & 16.72 $\pm$ 1.99 & 4.46 $\pm$ 0.32 & 9.42 $\pm$ 0.87 \\
MLP & 13.52 $\pm$ 0.65 & 31.01 $\pm$ 1.71 & 22.62 $\pm$ 0.55 & 48.02 $\pm$ 1.79 & 6.41 $\pm$ 0.25 & 15.04 $\pm$ 0.67 \\
\midrule
\textbf{GNN} & & & & & & \\
GCN & \textbf{\textcolor{blue}{16.61 $\pm$ 0.30}} & \textbf{\textcolor{blue}{36.26 $\pm$ 1.14}} & 21.09 $\pm$ 0.88 & 47.23 $\pm$ 1.88 & 7.13 $\pm$ 0.27 & 15.22 $\pm$ 0.57 \\
GAT & 13.84 $\pm$ 0.68 & 32.89 $\pm$ 1.27 & 19.58 $\pm$ 0.84 & 45.30 $\pm$ 1.30 & 4.95 $\pm$ 0.14 & 9.99 $\pm$ 0.64 \\
SAGE & 14.74 $\pm$ 0.69 & 34.65 $\pm$ 1.47 & 21.09 $\pm$ 1.15 & 48.75 $\pm$ 1.85 & \textbf{\textcolor{MyDarkGreen}{9.40 $\pm$ 0.70}} & \textbf{\textcolor{MyDarkGreen}{20.54 $\pm$ 1.40}} \\
GAE & \textbf{\textcolor{red}{18.32 $\pm$ 0.41}} & \textbf{\textcolor{red}{37.95 $\pm$ 1.24}} & 25.25 $\pm$ 0.82 & 49.65 $\pm$ 1.48 & 5.27 $\pm$ 0.25 & 10.50 $\pm$ 0.46 \\
\midrule
\textbf{GNN+Pairwise Info} & & & & & & \\
SEAL & 10.67 $\pm$ 3.46 & 24.27 $\pm$ 6.74 & 13.16 $\pm$ 1.66 & 27.37 $\pm$ 3.20 & 5.88 $\pm$ 0.53 & 12.47 $\pm$ 1.23 \\
BUDDY & 13.71 $\pm$ 0.59 & 30.40 $\pm$ 1.18 & 22.84 $\pm$ 0.36 & 48.35 $\pm$ 1.18 & 7.56 $\pm$ 0.18 & 16.78 $\pm$ 0.53 \\
Neo-GNN & 13.95 $\pm$ 0.39 & 31.27 $\pm$ 0.72 & 17.34 $\pm$ 0.84 & 41.74 $\pm$ 1.18 & 7.74 $\pm$ 0.30 & 17.88 $\pm$ 0.71 \\
NCN & 14.66 $\pm$ 0.95 & 35.14 $\pm$ 1.04 & \textbf{\textcolor{red}{28.65 $\pm$ 1.21}} & \textbf{\textcolor{blue}{53.41 $\pm$ 1.46}} & 5.84 $\pm$ 0.22 & 13.22 $\pm$ 0.56 \\
NCNC & 14.98 $\pm$ 1.00 & \textbf{\textcolor{MyDarkGreen}{36.70 $\pm$ 1.57}} &  24.10 $\pm$ 0.65 & \textbf{\textcolor{red}{53.72 $\pm$ 0.97}} & 8.58 $\pm$ 0.59 & 18.81 $\pm$ 1.16 \\
NBFNet & 13.56 $\pm$ 0.58 & 31.12 $\pm$ 0.75 & 14.29 $\pm$ 0.80 & 31.39 $\pm$ 1.34 & \multicolumn{2}{c}{>24h} \\
PEG & \textbf{\textcolor{MyDarkGreen}{15.73 $\pm$ 0.39}} & 36.03 $\pm$ 0.75 & 21.01 $\pm$ 0.77 & 45.56 $\pm$ 1.38 & 4.40 $\pm$ 0.41 & 8.70 $\pm$ 1.26 \\
\midrule
\textbf{Physics-Inspired GNN} & & & & & & \\ 
GRAFF.LP (Hadamard) &  15.73 $\pm$ 0.77 & 34.76 $\pm$ 1.13 &  \textbf{\textcolor{blue}{26.77 $\pm$ 1.1}} & \textbf{\textcolor{MyDarkGreen}{51.76 $\pm$ 1.68}} &\textbf{\textcolor{red}{13.49 $\pm$ 0.8}}  & \textbf{\textcolor{red}{27.20 $\pm$ 0.84}}\\  
GRAFF-LP (Gradient) &13.75 $\pm$ 0.66 &31.56 $\pm$ 1.57 & \textbf{\textcolor{MyDarkGreen}{25.7 $\pm$ 1.32}} &  49.98 $\pm$ 1.1 & \textbf{\textcolor{blue}{12.29 $\pm$ 0.79}}& \textbf{\textcolor{blue}{26.46 $\pm$ 2.69}}\\
\bottomrule

\end{tabular}}

\end{table}

\subsection{Additional Examples of Attraction and Repulsion through $GS^T$}
\label{sec:additional_examples}
Here we show some examples where GRAFF-LP can learn to separate the gradients. We have examples both with $f_h$ as well as with $f_g$. These examples help to answer more profoundly to \textbf{RQ2}.
\begin{figure*}[htbp]
    \centering
    \caption{$||(\nabla \mathbf{H}^t)_{i,j}||^2$ evolution with a fully-trained 9-layers GRAFF-LP via $f_g$ on \texttt{Amazon Ratings}.}
    \begin{subfigure}{0.24\textwidth}
        \centering
        \includegraphics[width=\linewidth]{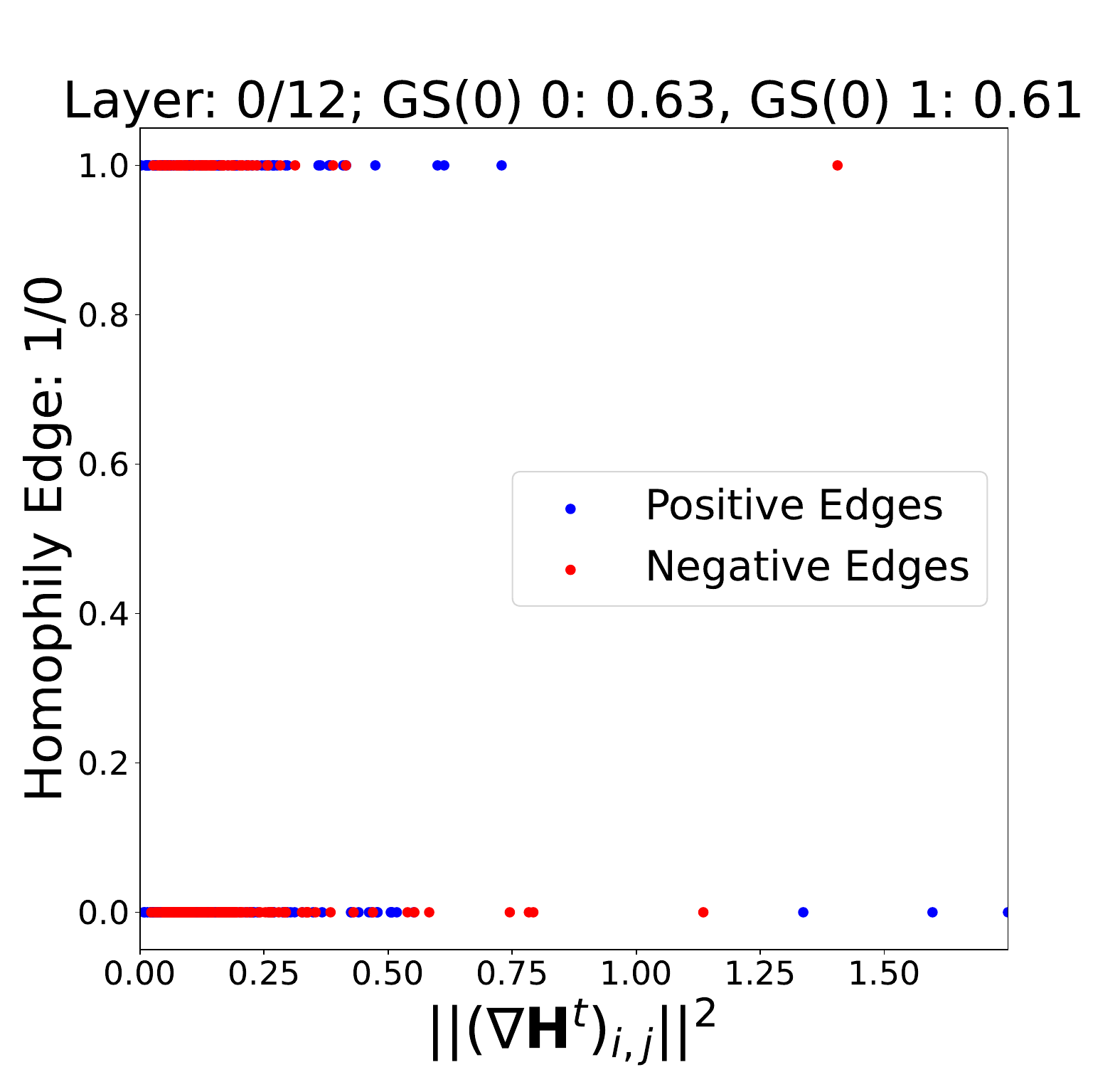}  
        \caption{$GS^0_{hm, hm}$, $GS^0_{ht, ht}$}
        
    \end{subfigure}
    \begin{subfigure}{0.24\textwidth}
        \centering
        \includegraphics[width=\linewidth]{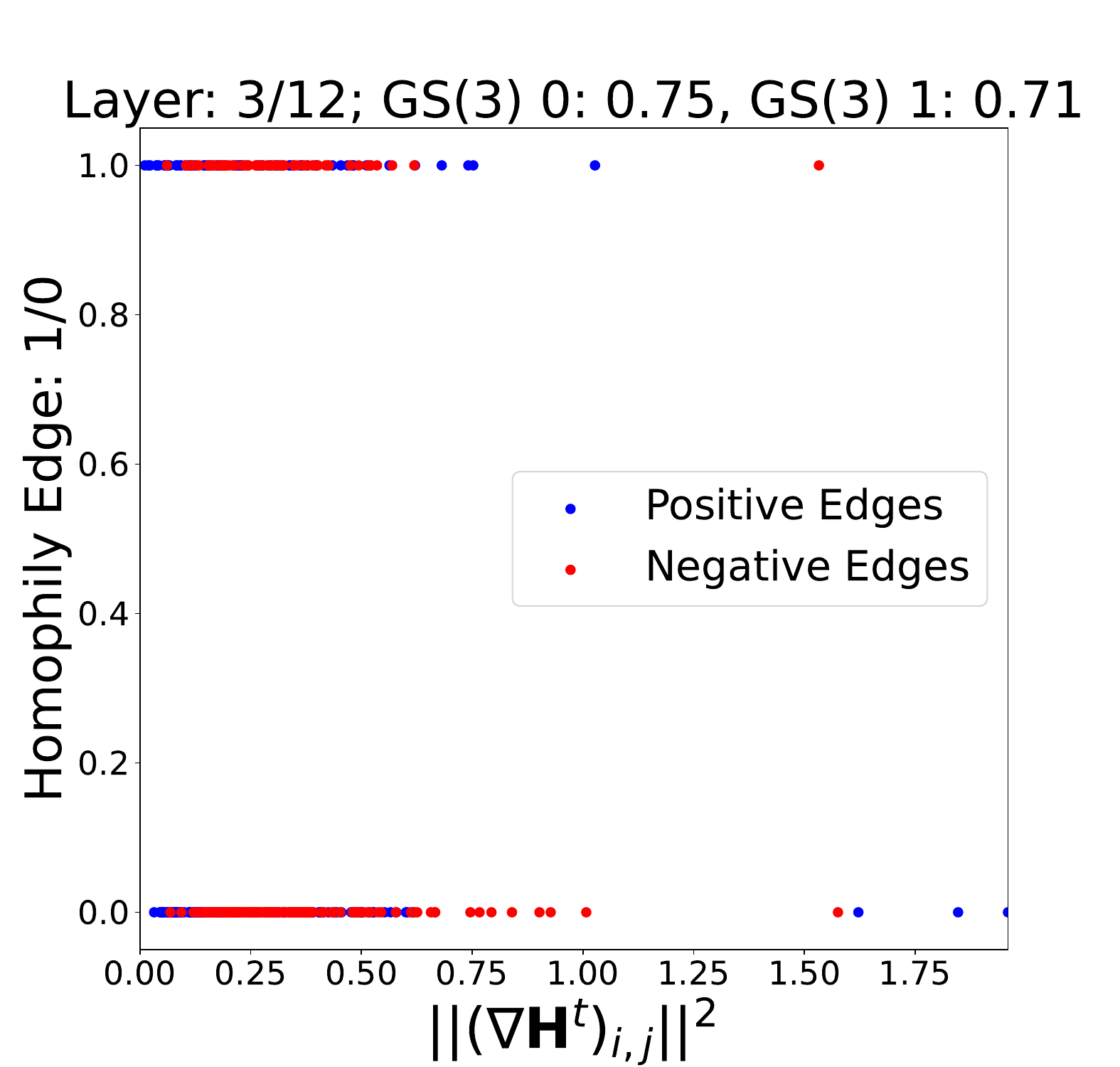}  
        \caption{$GS^3_{hm, hm}$, $GS^3_{ht, ht}$.}
        
    \end{subfigure}
    \begin{subfigure}{0.24\textwidth}
        \centering
        \includegraphics[width=\linewidth]{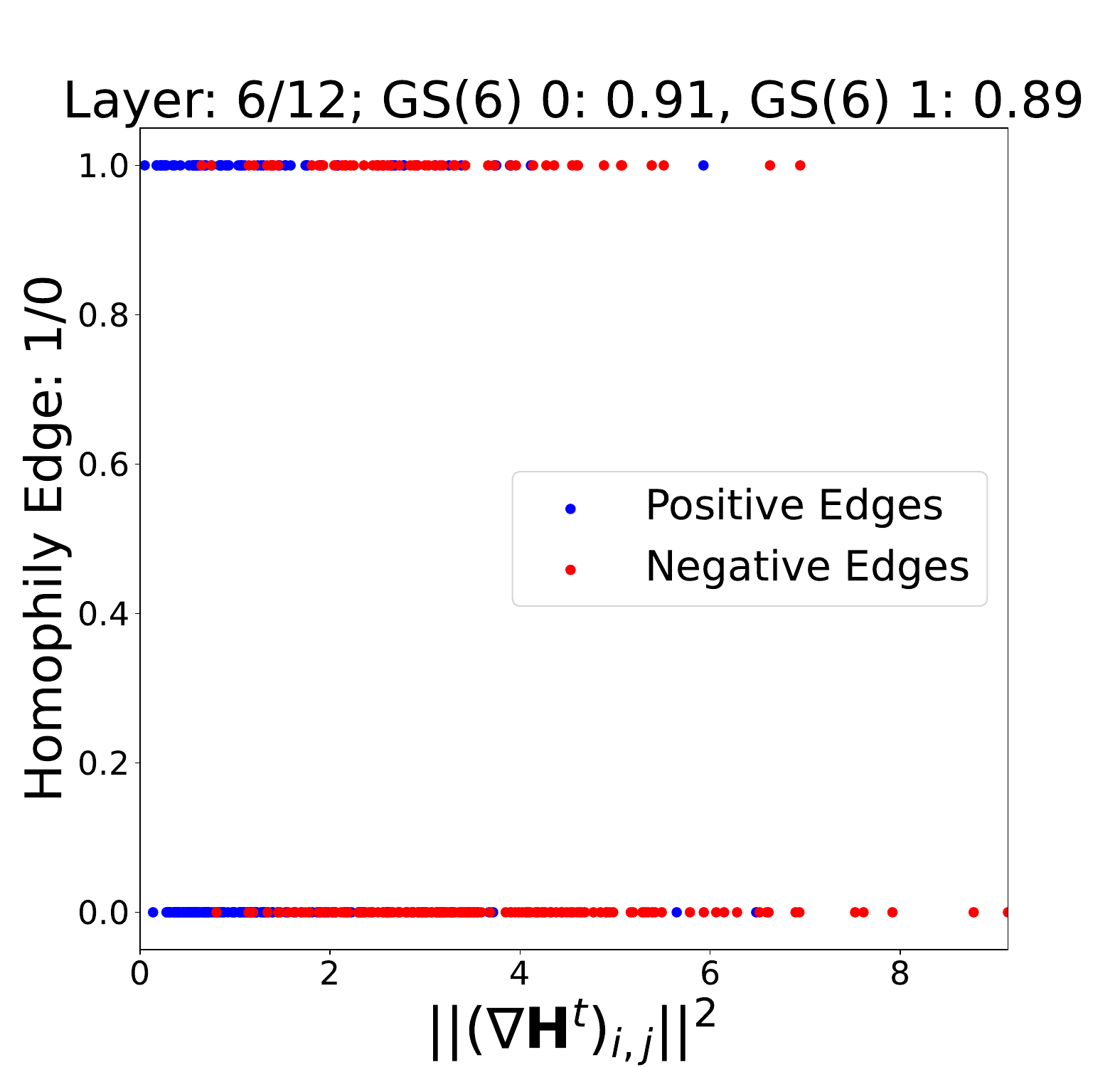}  
        \caption{$GS^6_{hm, hm}$, $GS^6_{ht, ht}$}
        
    \end{subfigure}
    \begin{subfigure}{0.24\textwidth}
        \centering
        \includegraphics[width=\linewidth]{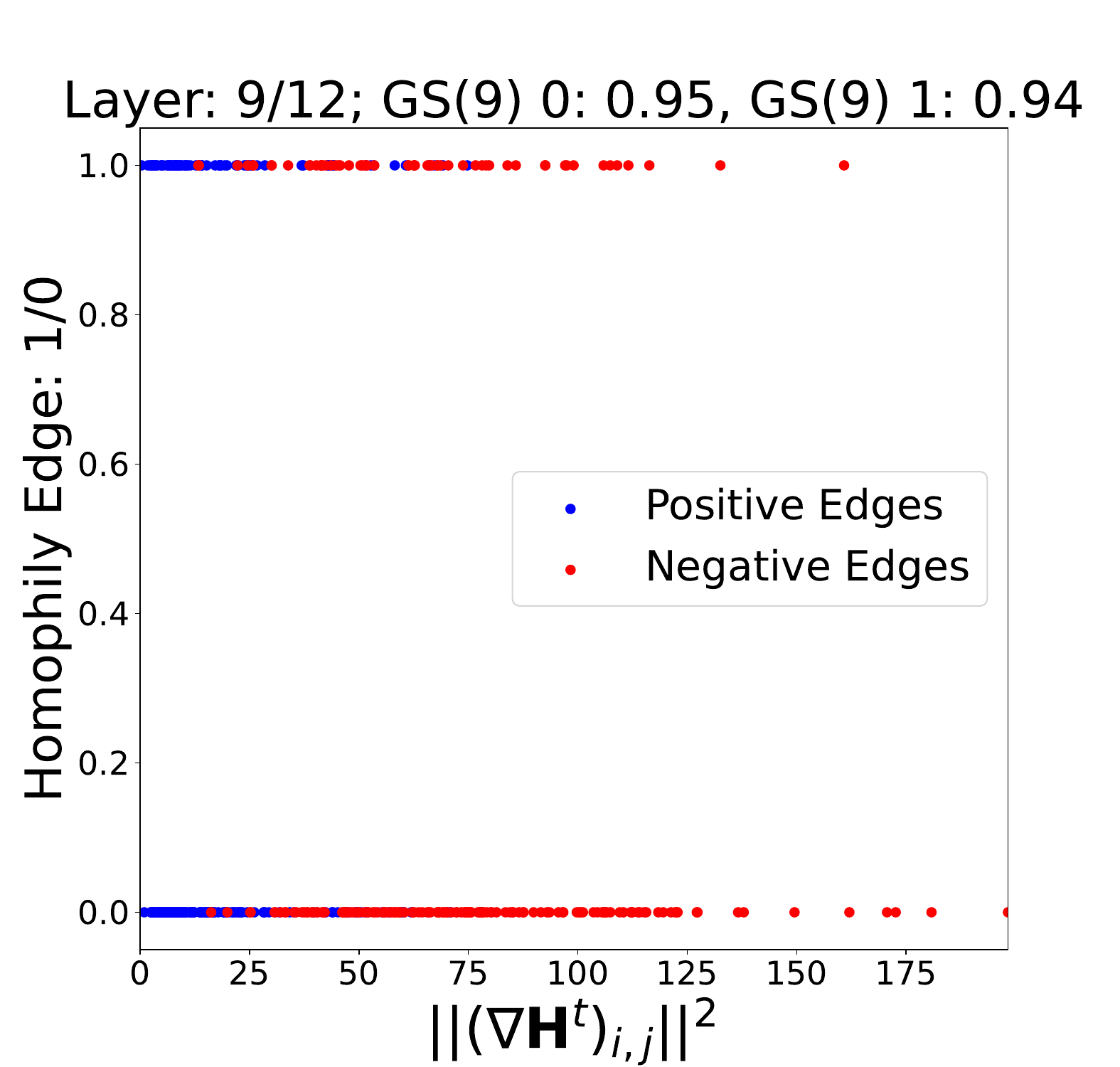}  
        \caption{$GS^9_{hm, hm}$, $GS^9_{ht, ht}$}
        
    \end{subfigure}
    \label{fig:attraction_repulsion_amazon_grad}
\end{figure*}

\begin{figure}[htbp]
    \centering
    \caption{$||(\nabla \mathbf{H}^t)_{i,j}||^2$ evolution with a fully-trained 9-layers GRAFF-LP via $f_h$ on \texttt{Amazon Ratings}.}
    \begin{subfigure}{0.24\textwidth}
        \centering
        \includegraphics[width=\linewidth]{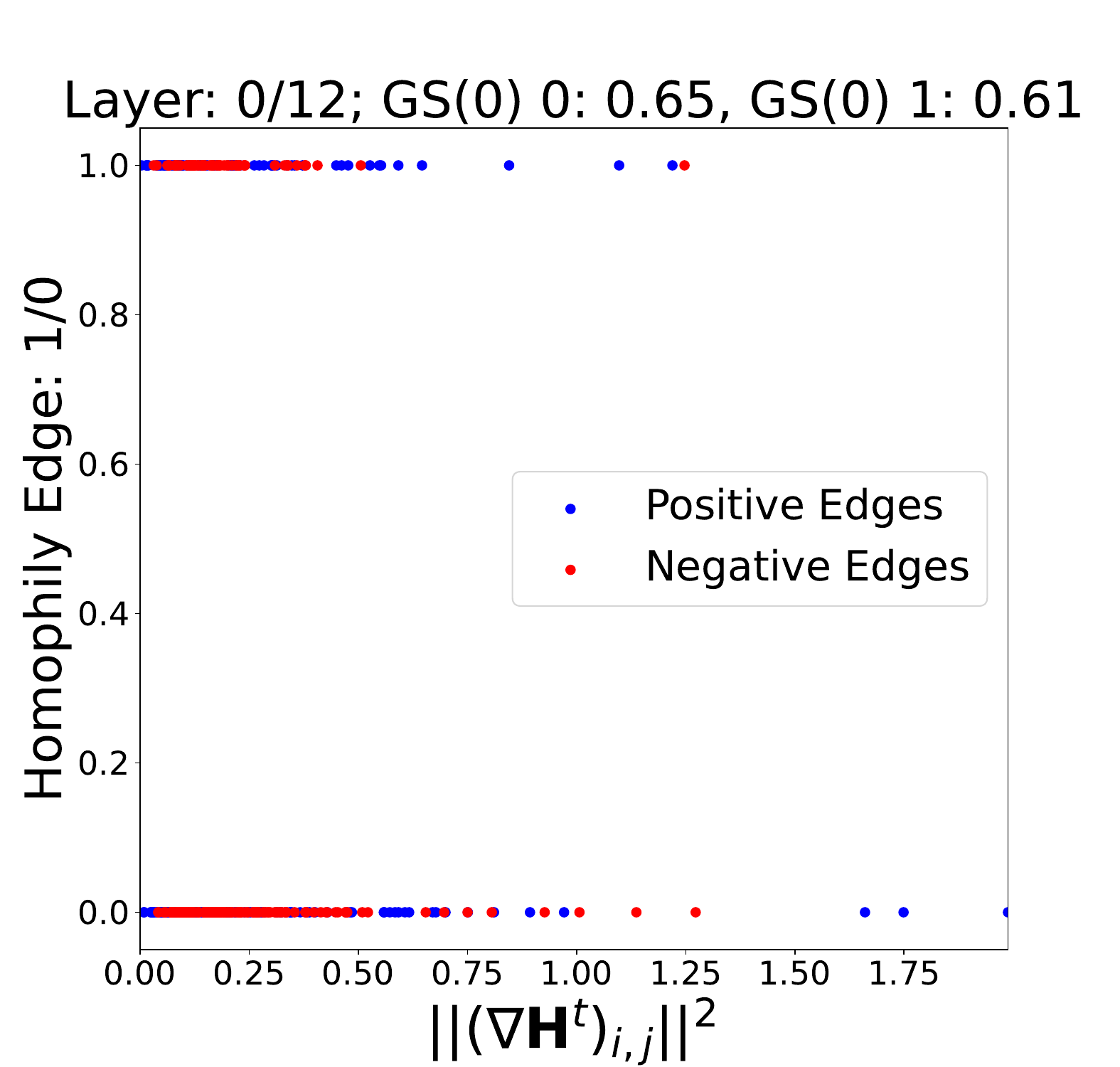}  
        \caption{$GS^0_{hm, hm}$, $GS^0_{ht, ht}$}
        
    \end{subfigure}
    \begin{subfigure}{0.24\textwidth}
        \centering
        \includegraphics[width=\linewidth]{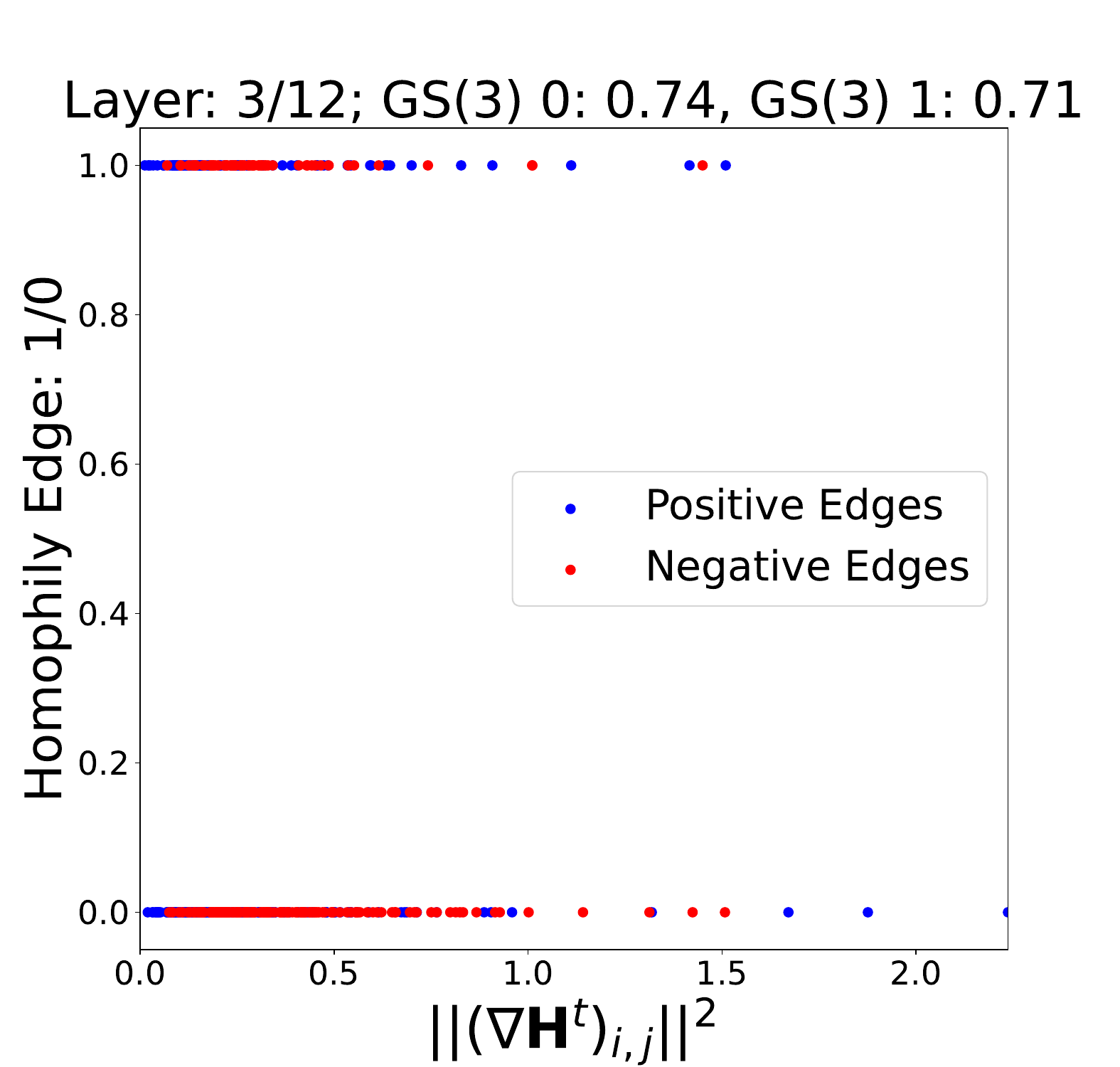}  
        \caption{$GS^3_{hm, hm}$, $GS^3_{ht, ht}$.}
        
    \end{subfigure}
    \begin{subfigure}{0.24\textwidth}
        \centering
        \includegraphics[width=\linewidth]{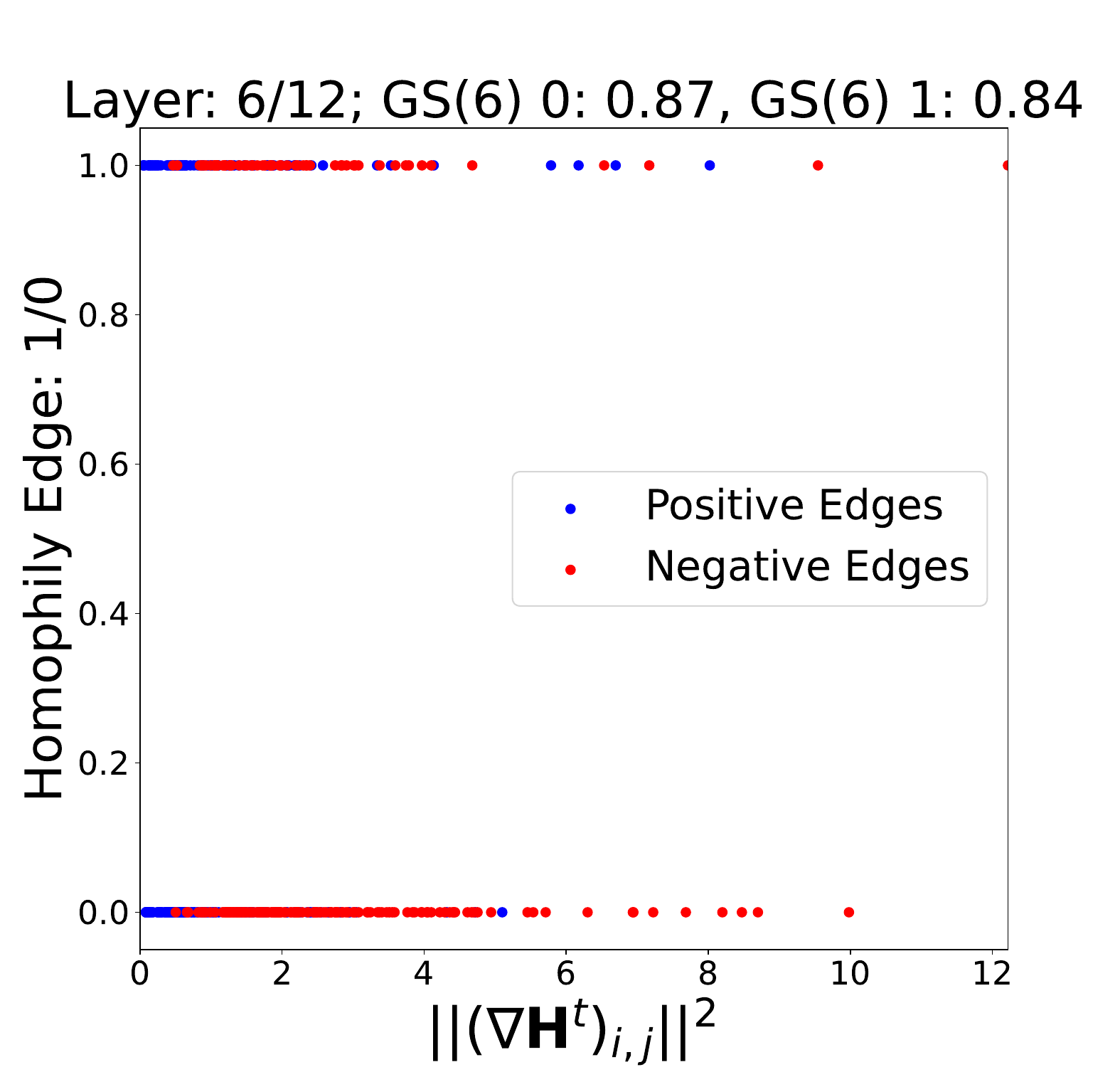}  
        \caption{$GS^6_{hm, hm}$, $GS^6_{ht, ht}$}
        
    \end{subfigure}
    \begin{subfigure}{0.24\textwidth}
        \centering
        \includegraphics[width=\linewidth]{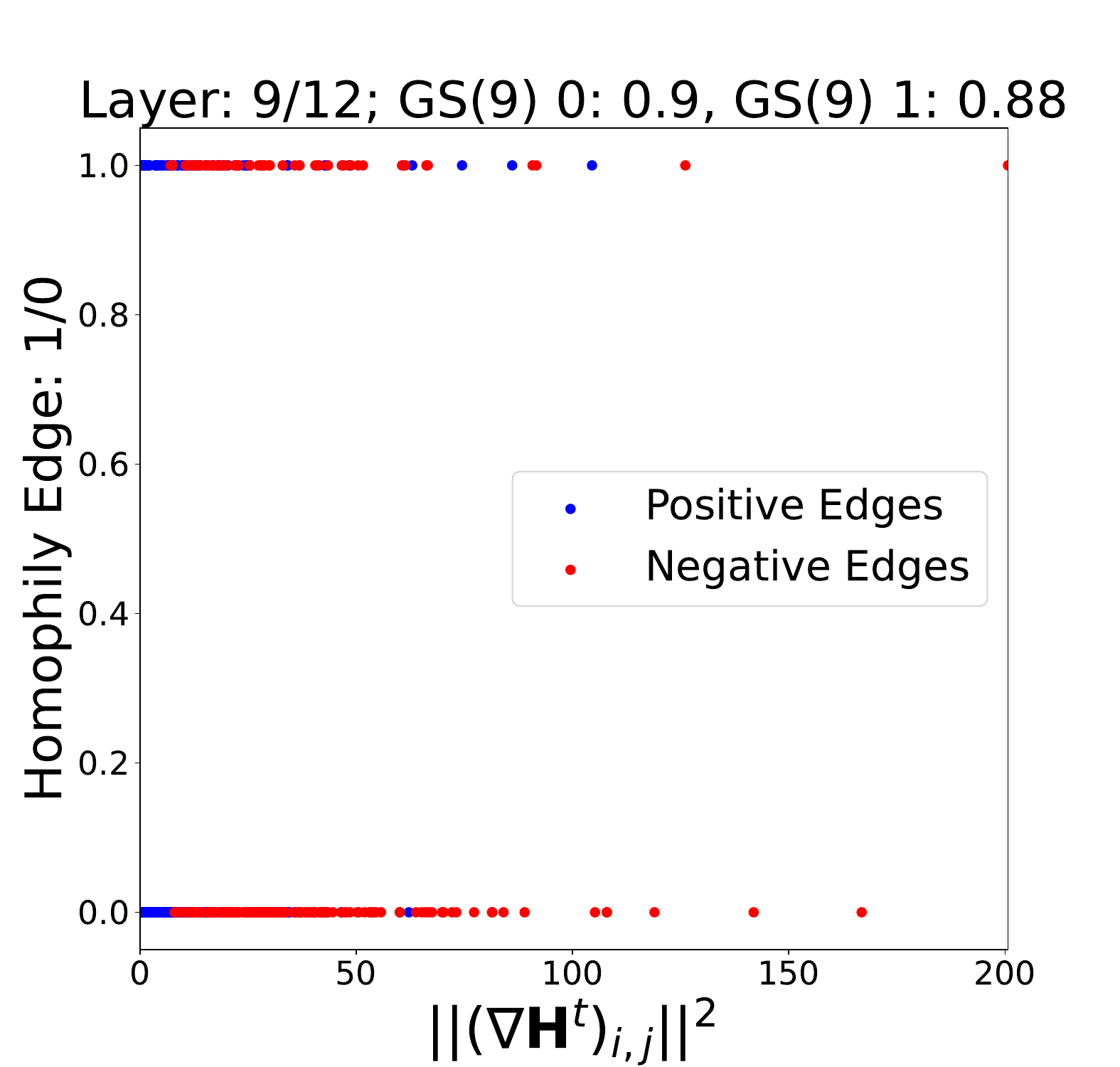}  
        \caption{$GS^9_{hm, hm}$, $GS^9_{ht, ht}$}
        
    \end{subfigure}
    \label{fig:attraction_repulsion_amazon}
\end{figure}
\begin{figure}[htbp]
    \centering
    \caption{$||(\nabla \mathbf{H}^t)_{i,j}||^2$ evolution with a fully-trained 9-layers GRAFF-LP via $f_h$ on \texttt{minesweeper}.}
    \begin{subfigure}{0.24\textwidth}
        \centering
        \includegraphics[width=\linewidth]{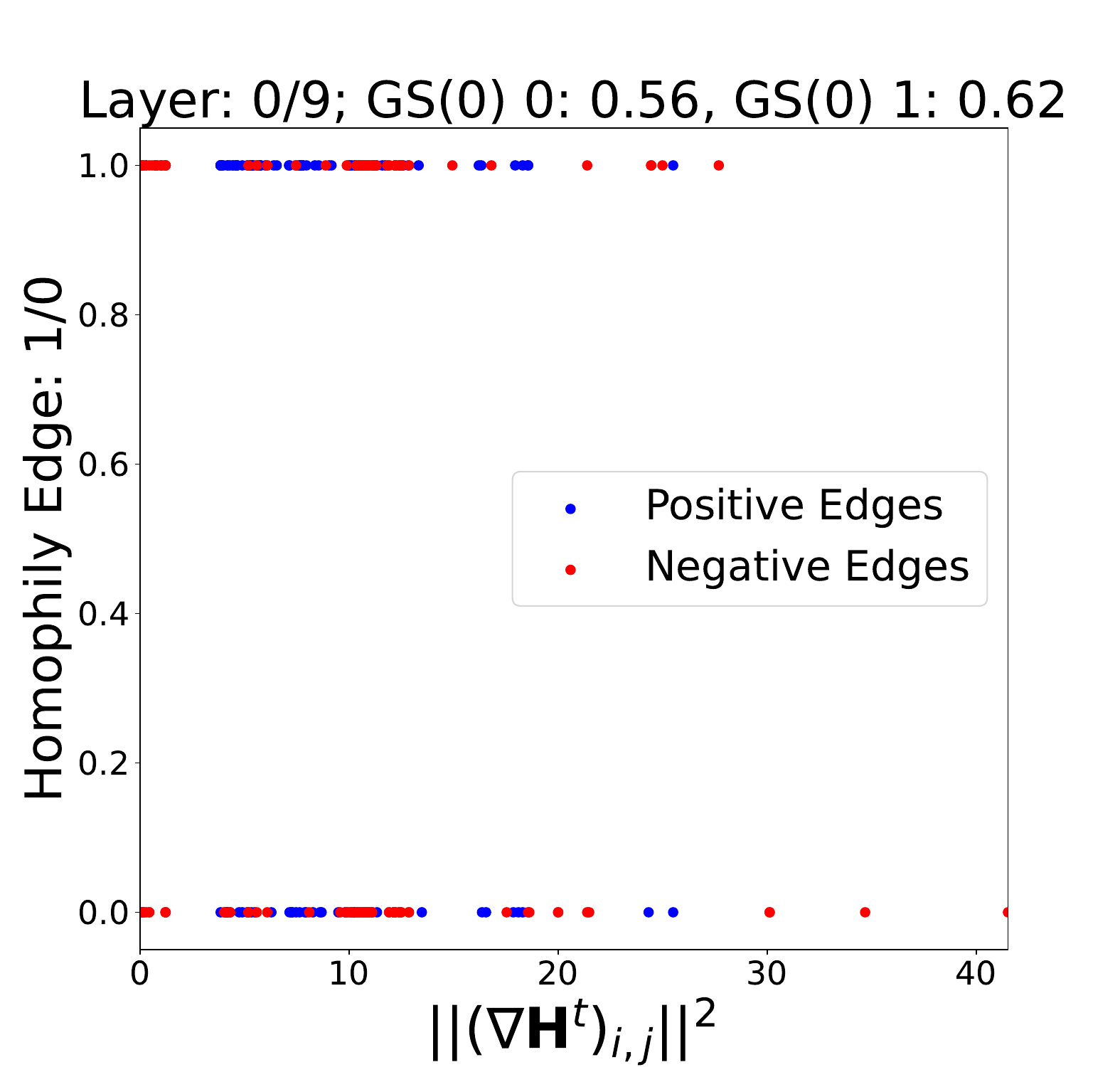}  
        \caption{$GS^0_{hm, hm}$, $GS^0_{ht, ht}$}
        
    \end{subfigure}
    \begin{subfigure}{0.24\textwidth}
        \centering
        \includegraphics[width=\linewidth]{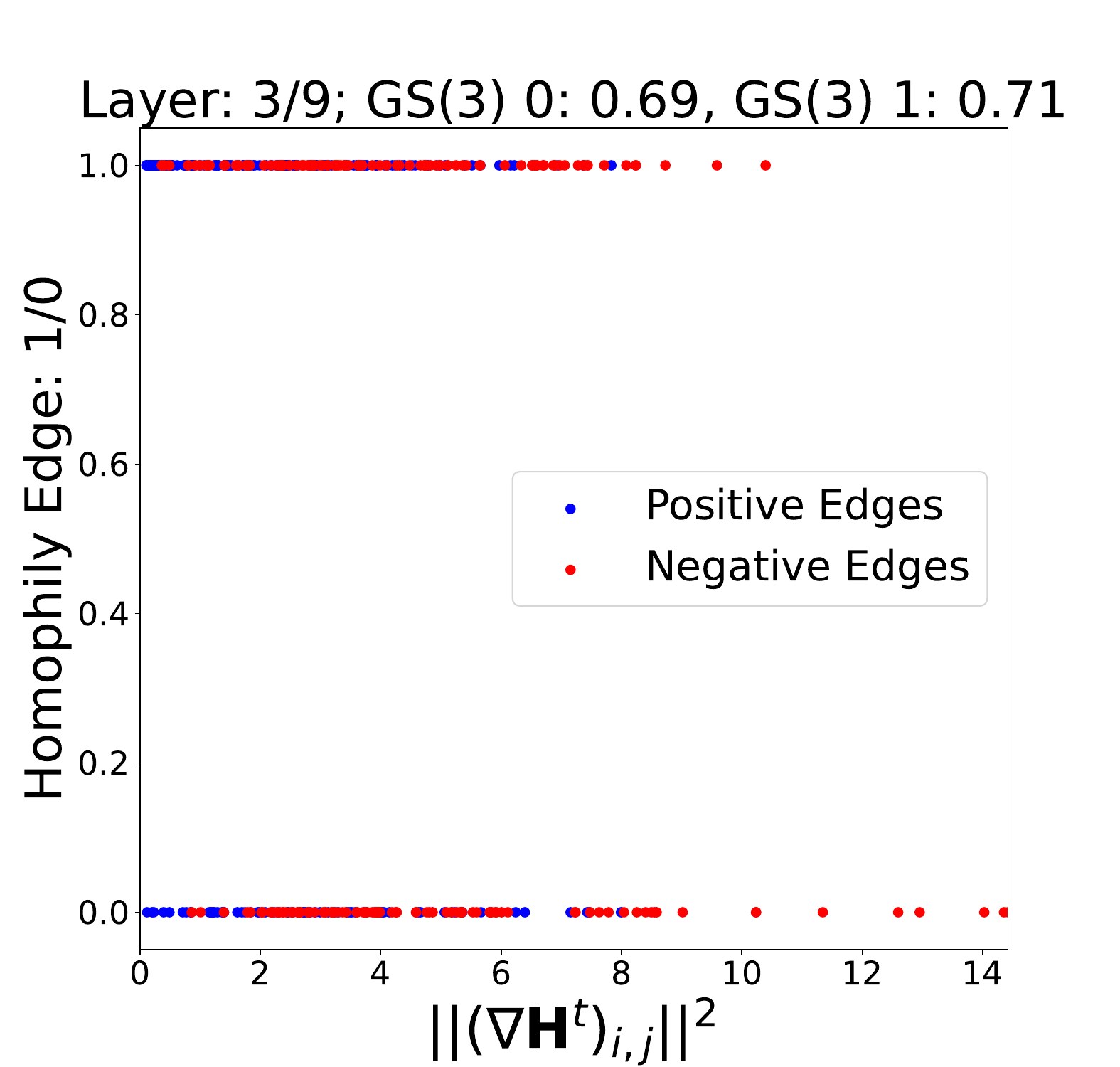}  
        \caption{$GS^3_{hm, hm}$, $GS^3_{ht, ht}$.}
        
    \end{subfigure}
    \begin{subfigure}{0.24\textwidth}
        \centering
        \includegraphics[width=\linewidth]{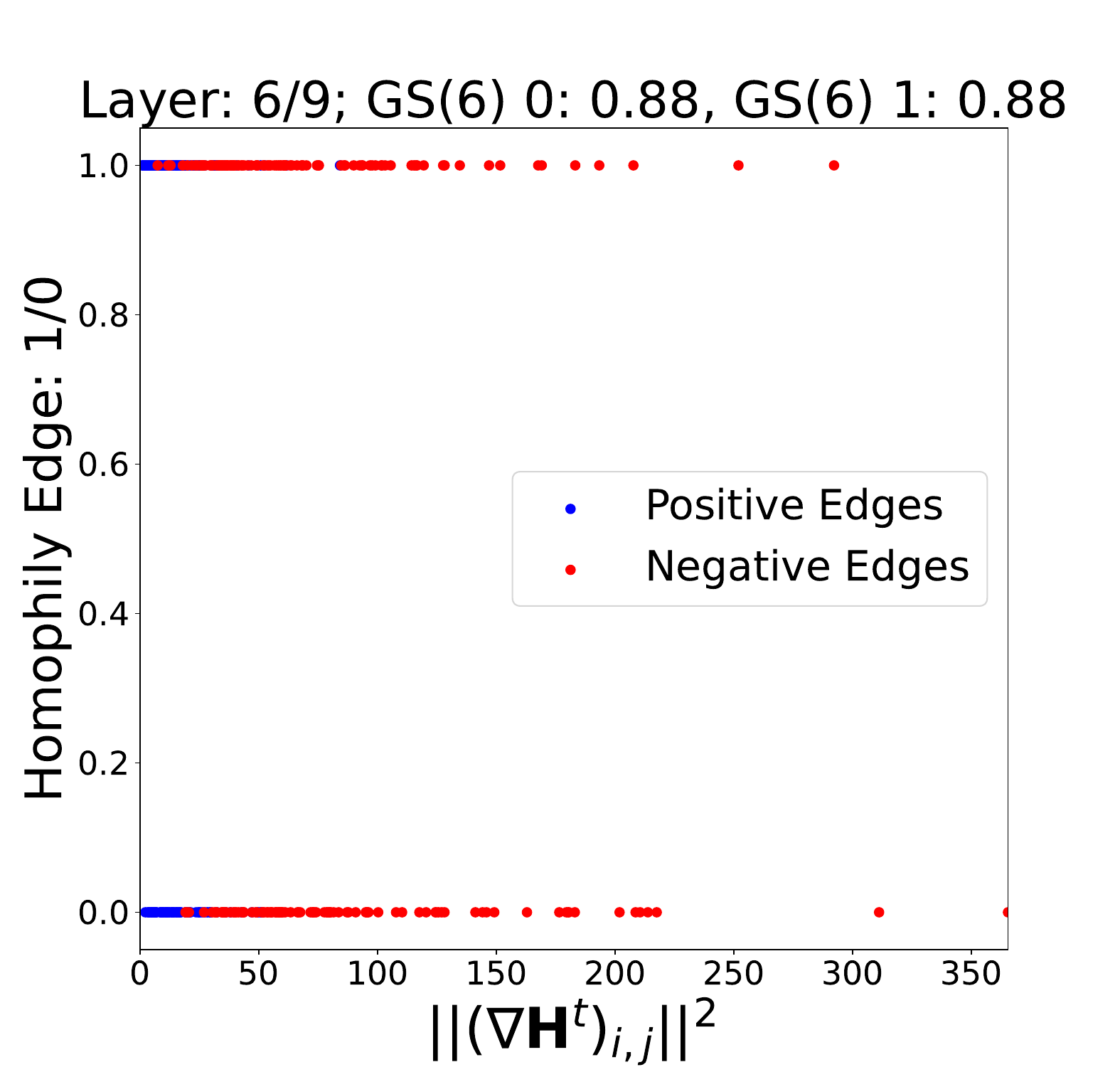}  
        \caption{$GS^6_{hm, hm}$, $GS^6_{ht, ht}$}
        
    \end{subfigure}
    \begin{subfigure}{0.24\textwidth}
        \centering
        \includegraphics[width=\linewidth]{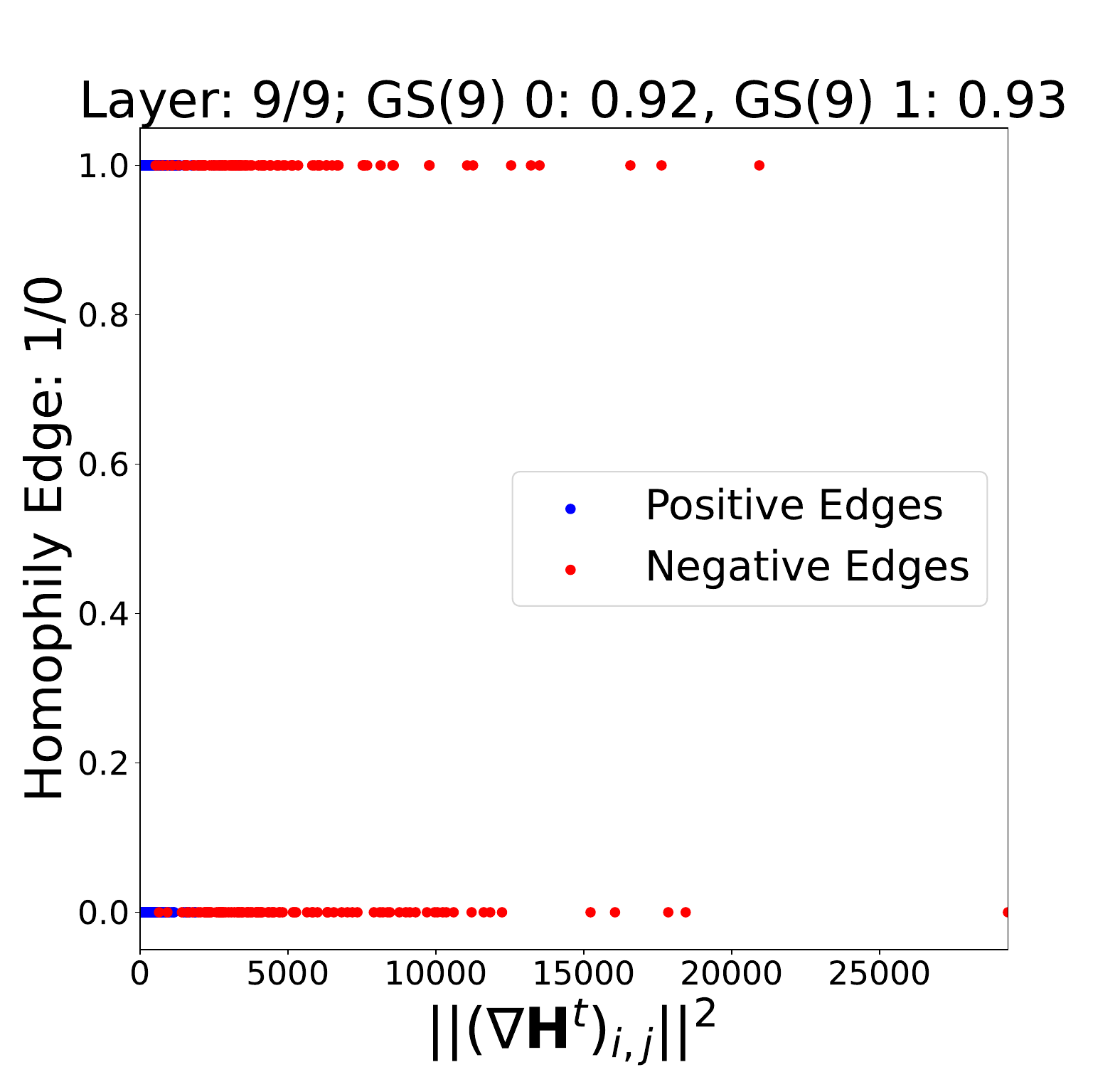}  
        \caption{$GS^9_{hm, hm}$, $GS^9_{ht, ht}$}
        
    \end{subfigure}
    \label{fig:attraction_repulsion_mines}
\end{figure}
\begin{figure}[htbp]
    \centering
    \caption{$||(\nabla \mathbf{H}^t)_{i,j}||^2$ evolution with a fully-trained 7-layers GRAFF-LP via $f_h$ on \texttt{Roman Empire}.}
    \begin{subfigure}{0.24\textwidth}
        \centering
        \includegraphics[width=\linewidth]{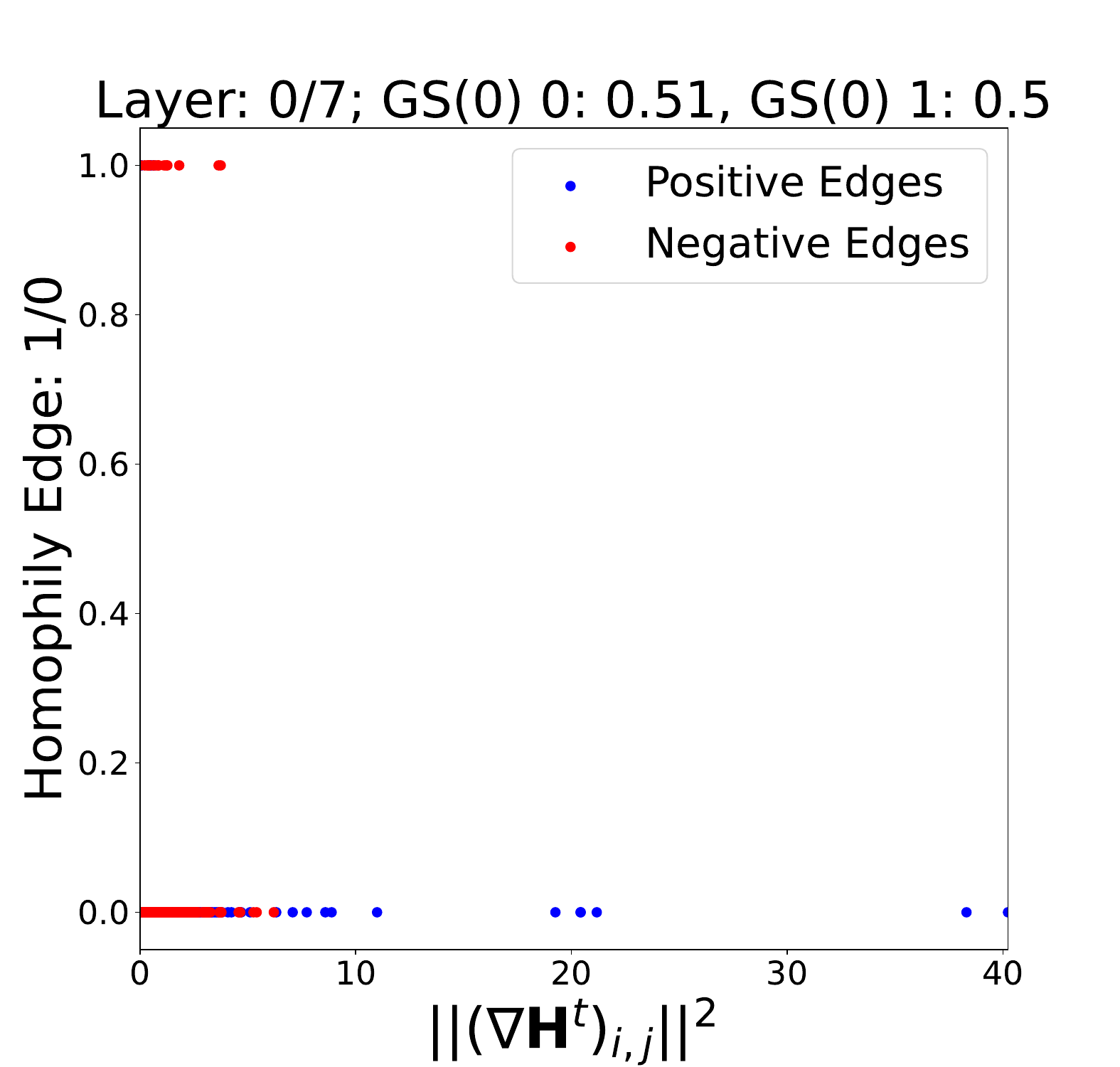}  
        \caption{$GS^0_{hm, hm}$, $GS^0_{ht, ht}$}
        
    \end{subfigure}
    \begin{subfigure}{0.24\textwidth}
        \centering
        \includegraphics[width=\linewidth]{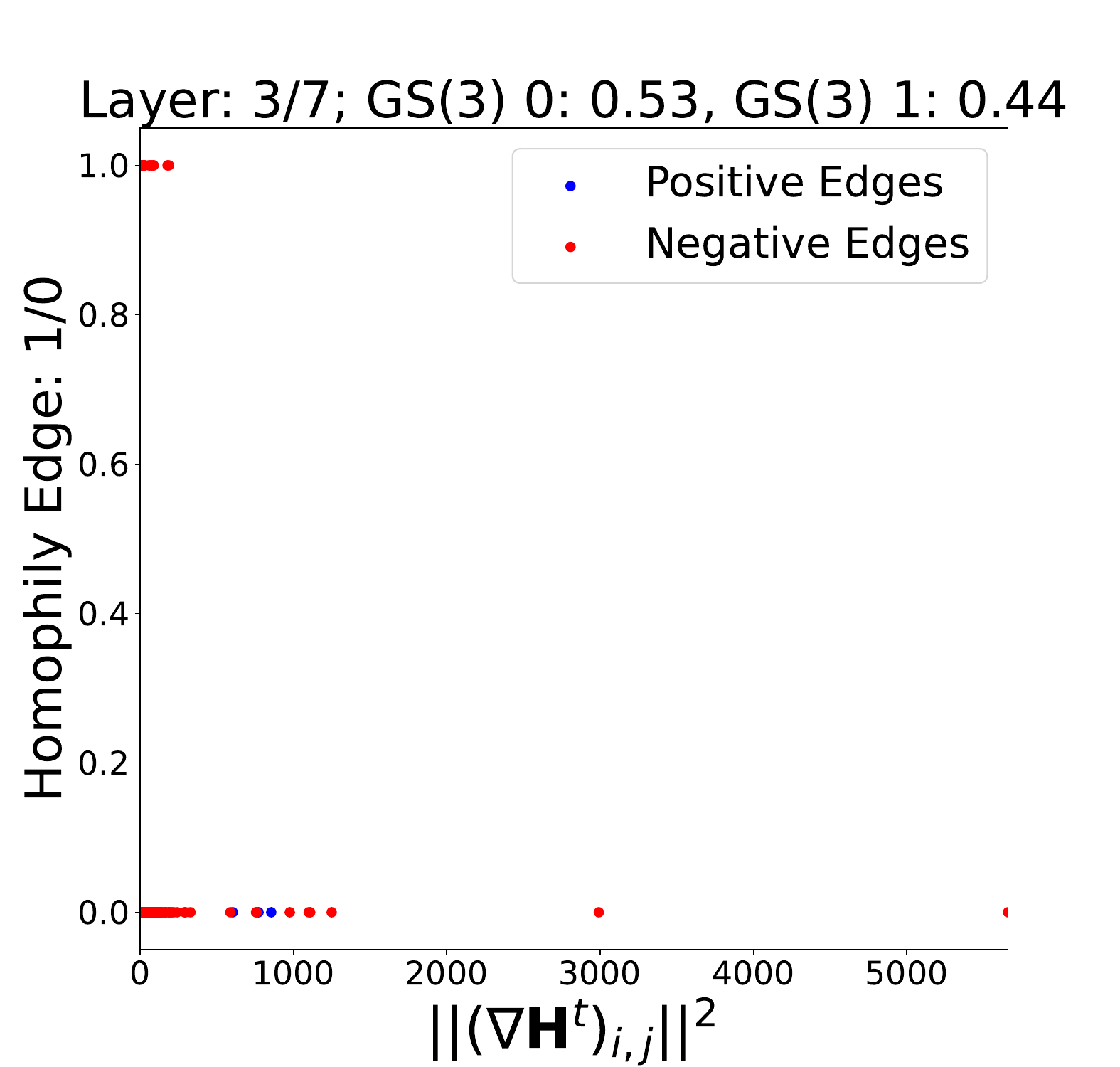}  
        \caption{$GS^3_{hm, hm}$, $GS^3_{ht, ht}$.}
        
    \end{subfigure}
    \begin{subfigure}{0.24\textwidth}
        \centering
        \includegraphics[width=\linewidth]{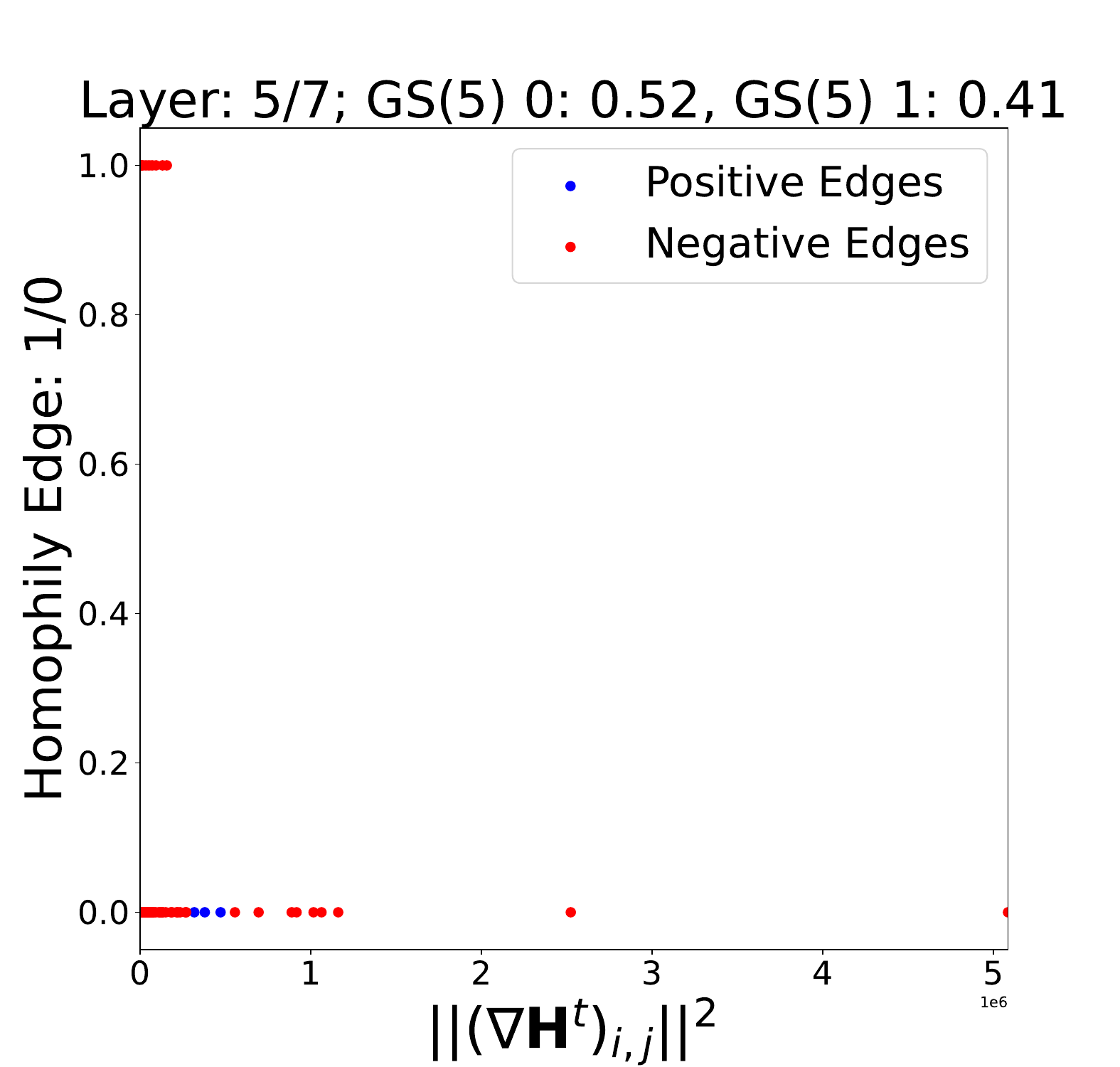}  
        \caption{$GS^5_{hm, hm}$, $GS^5_{ht, ht}$}
        
    \end{subfigure}
    \begin{subfigure}{0.24\textwidth}
        \centering
        \includegraphics[width=\linewidth]{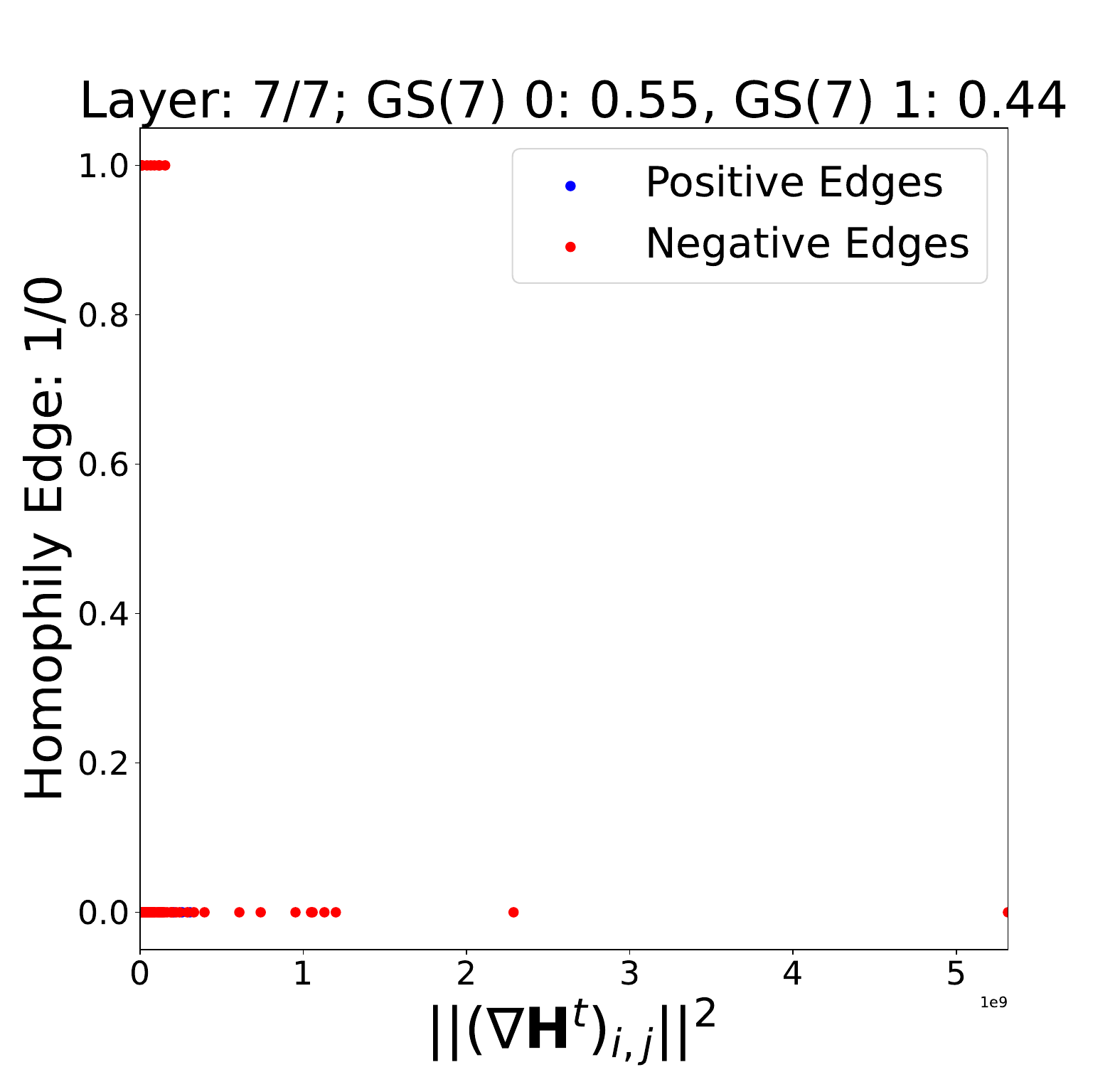}  
        \caption{$GS^7_{hm, hm}$, $GS^7_{ht, ht}$}
        
    \end{subfigure}
    \label{fig:attraction_repulsion_roman}
\end{figure}
\begin{figure}[htbp]
    \centering
    \caption{$||(\nabla \mathbf{H}^t)_{i,j}||^2$ evolution with a fully-trained 7-layers GRAFF-LP via $f_g$ on \texttt{Roman Empire}.}
    \begin{subfigure}{0.24\textwidth}
        \centering
        \includegraphics[width=\linewidth]{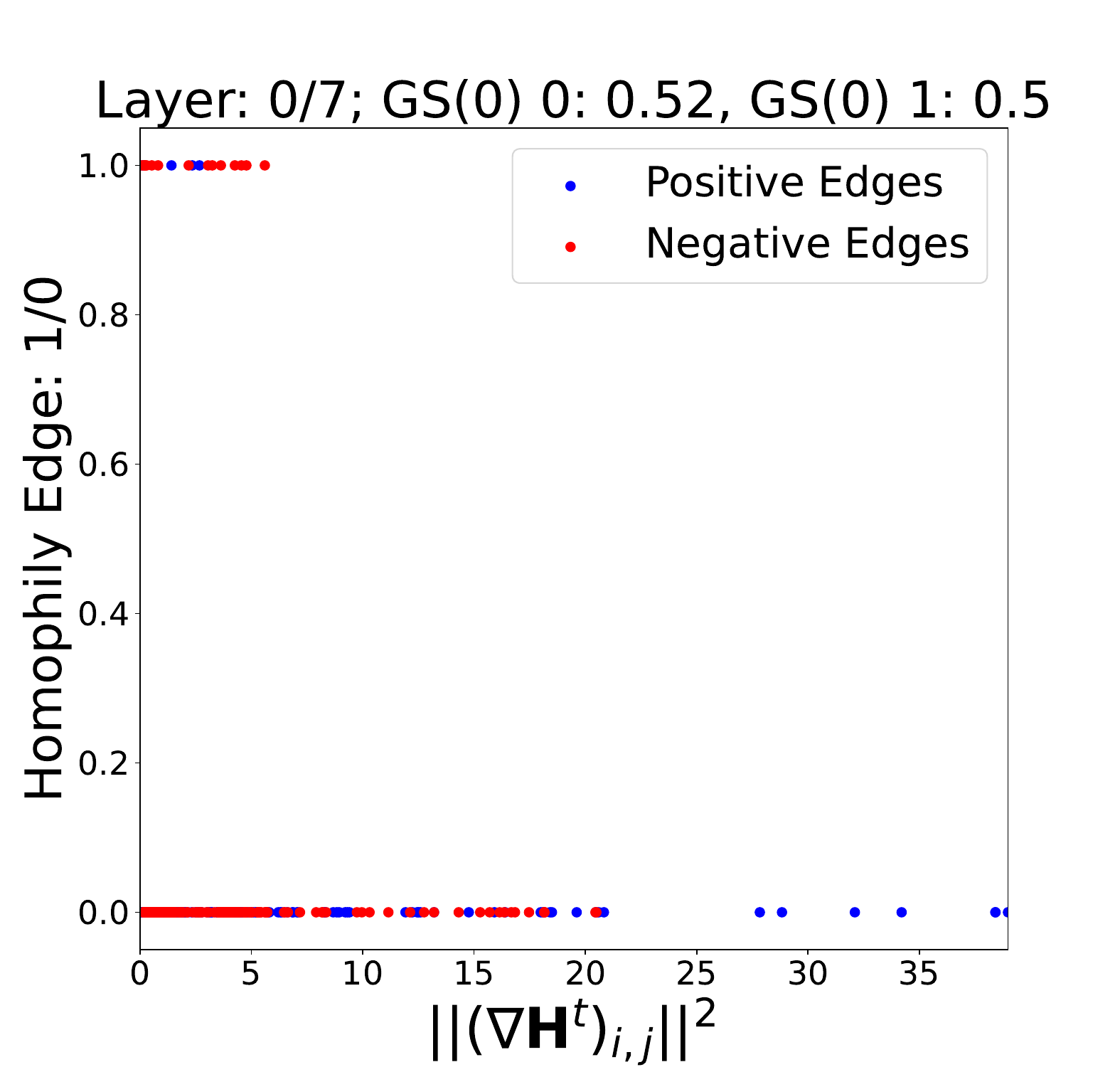}  
        \caption{$GS^0_{hm, hm}$, $GS^0_{ht, ht}$}
        
    \end{subfigure}
    \begin{subfigure}{0.24\textwidth}
        \centering
        \includegraphics[width=\linewidth]{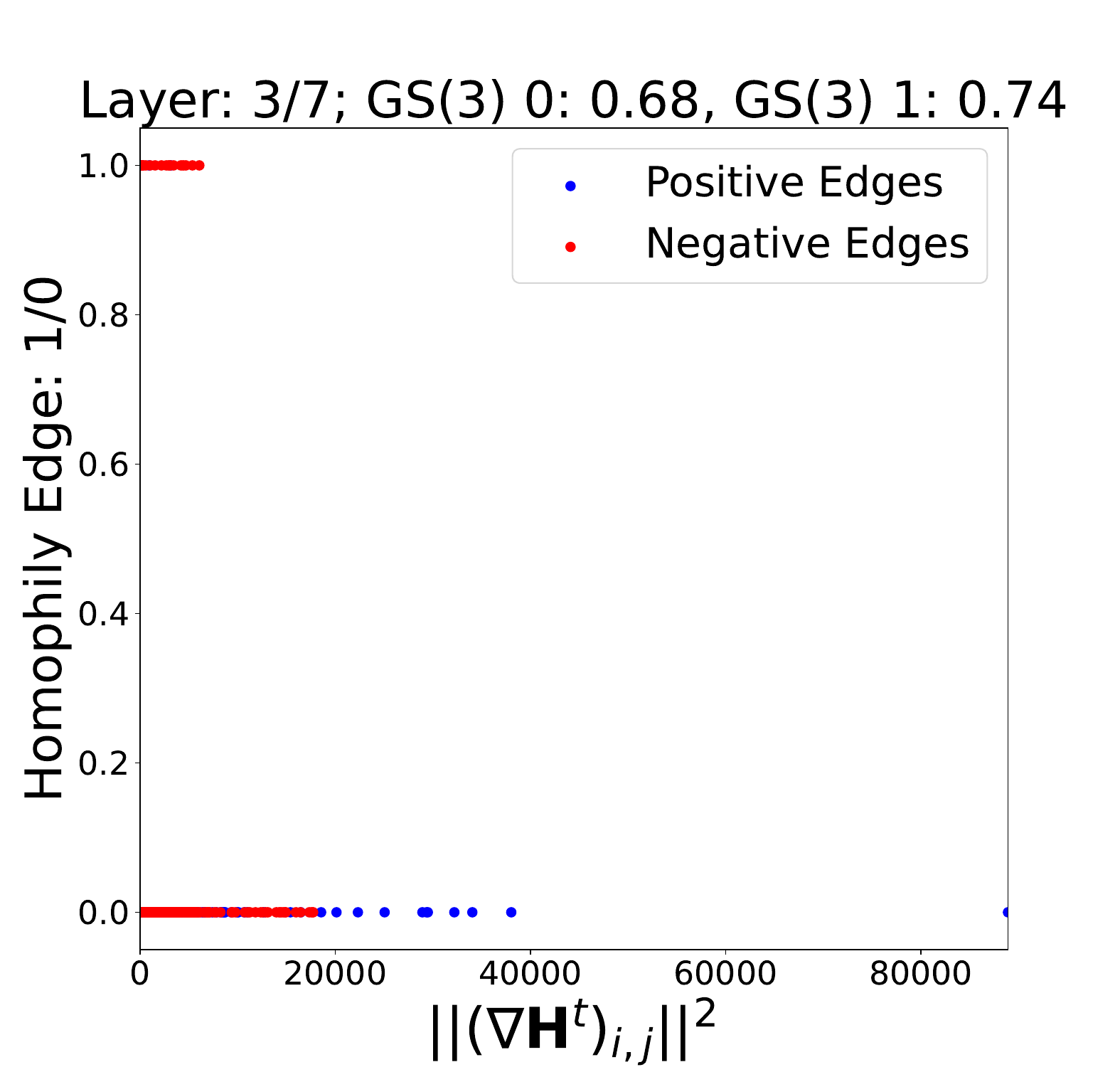}  
        \caption{$GS^3_{hm, hm}$, $GS^3_{ht, ht}$.}
        
    \end{subfigure}
    \begin{subfigure}{0.24\textwidth}
        \centering
        \includegraphics[width=\linewidth]{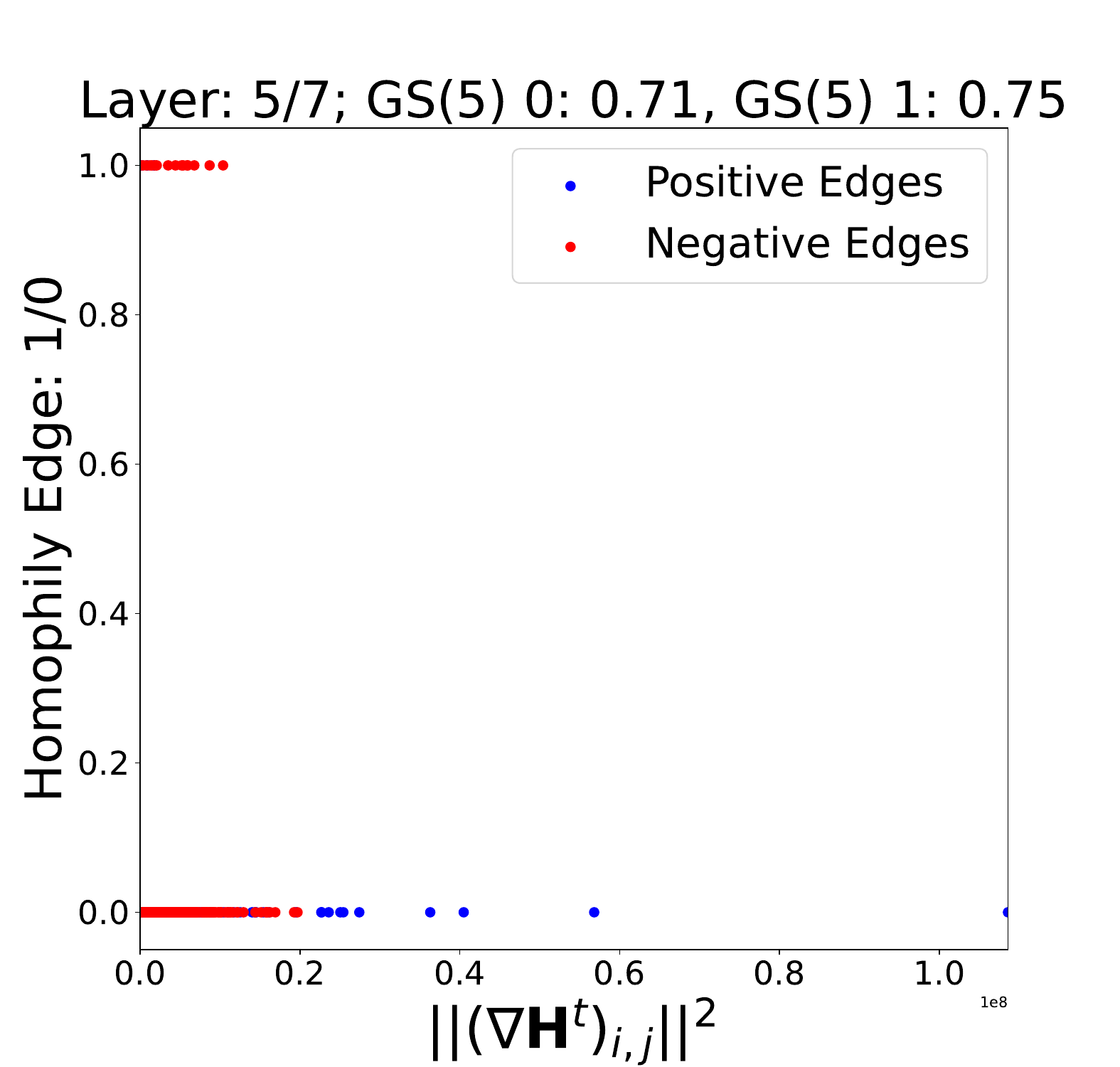}  
        \caption{$GS^5_{hm, hm}$, $GS^5_{ht, ht}$}
        
    \end{subfigure}
    \begin{subfigure}{0.24\textwidth}
        \centering
        \includegraphics[width=\linewidth]{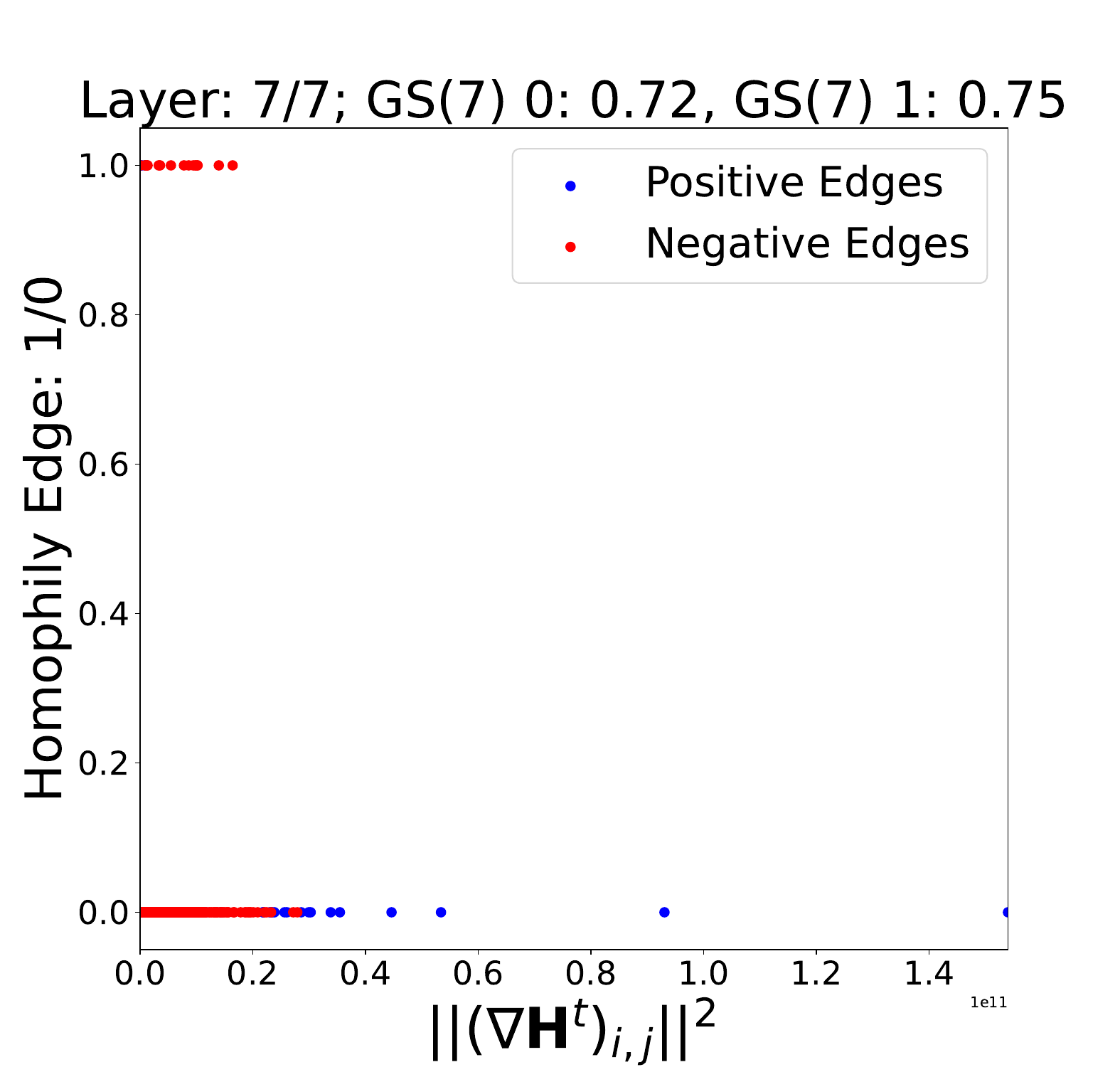}  
        \caption{$GS^7_{hm, hm}$, $GS^7_{ht, ht}$}
        
    \end{subfigure}
    \label{fig:attraction_repulsion_roman_grad}
\end{figure}

In Figures \ref{fig:attraction_repulsion_amazon_grad}, \ref{fig:attraction_repulsion_amazon}, we notice that with $f_g$ and even with $f_h$, GRAFF-LP has learnt to separate the edge gradients in \texttt{Amazon Ratings}. Of course, w.r.t. the paper results, $f_h$ provides a lighter separability. 
We can see the same for \texttt{Minesweeper} in Figure \ref{fig:attraction_repulsion_mines}. 
Another interesting behavior is GRAFF-LP with $f_h$ in \texttt{Roman Empire}, where its $GS^T$ is lower than 60\%, and indeed we see that it evolves as in Figure \ref{fig:attraction_repulsion_roman}. 
However, if we simply train it with $f_g$, we get $GS^T > 90\%$, as illustrated in Figure \ref{fig:attraction_repulsion_roman_grad}.

\begin{figure}[htbp]
    \centering
    \caption{$||(\nabla \mathbf{H}^t)_{i,j}||^2$ evolution with a fully-trained 7-layers GRAFF-LP via $f_g$ on \texttt{Questions}.}    
    \begin{subfigure}{0.24\textwidth}
        \centering
        \includegraphics[width=\linewidth]{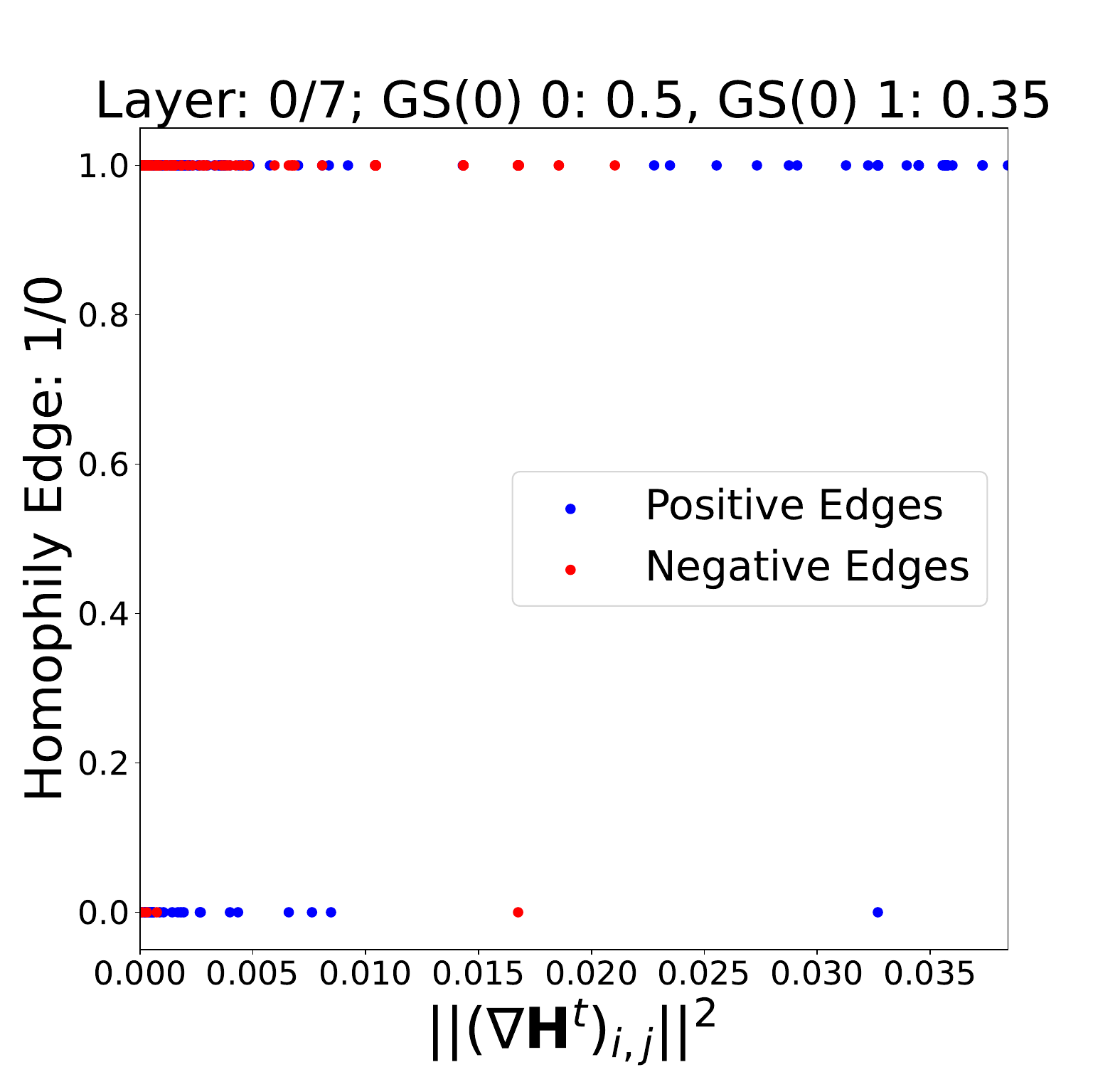}  
        \caption{$GS^0_{hm, hm}$, $GS^0_{ht, ht}$}
        \label{fig:question 1}
    \end{subfigure}
    \begin{subfigure}{0.24\textwidth}
        \centering
        \includegraphics[width=\linewidth]{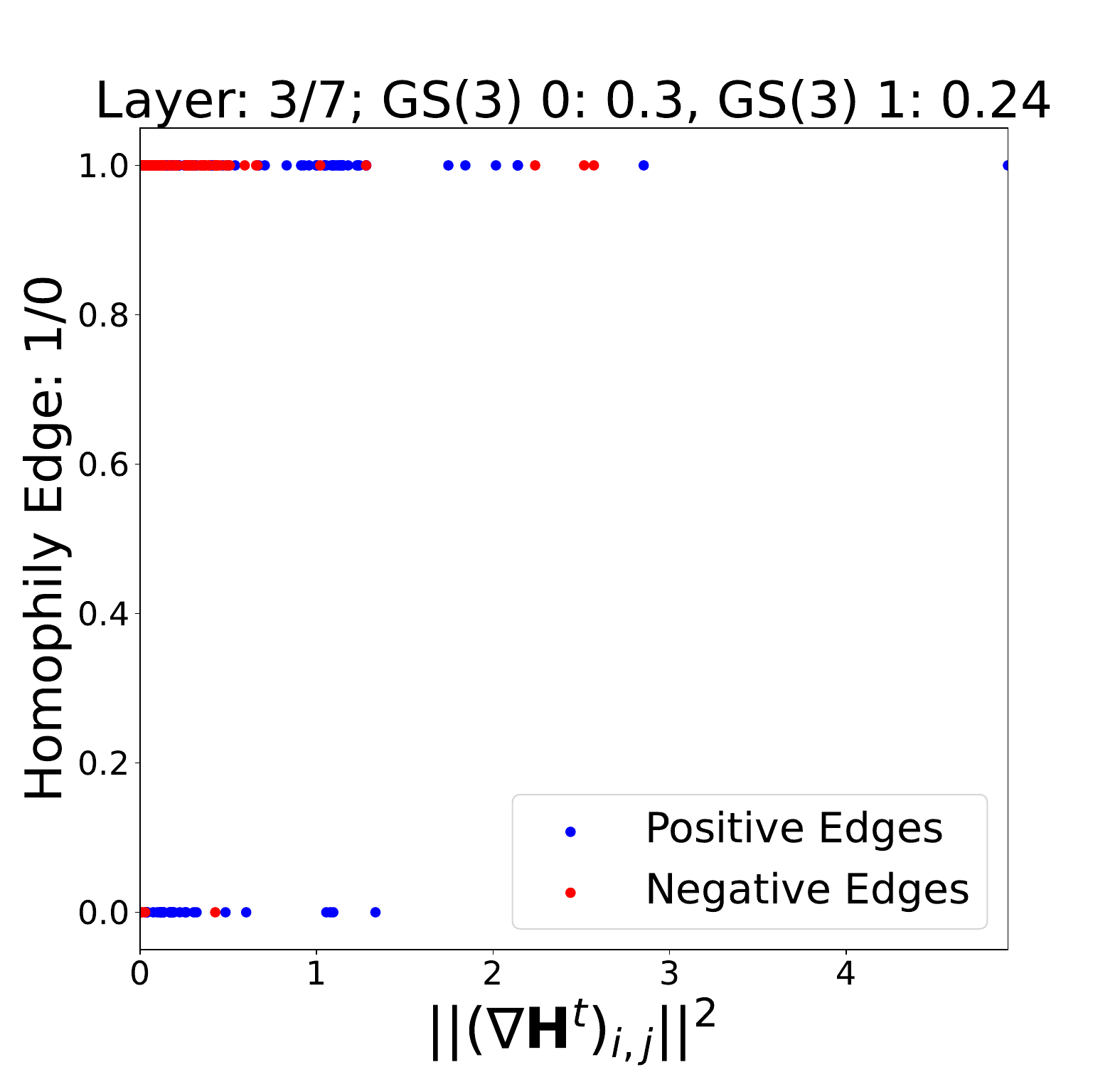}  
        \caption{$GS^3_{hm, hm}$, $GS^3_{ht, ht}$.}
        
    \end{subfigure}
    \begin{subfigure}{0.24\textwidth}
        \centering
        \includegraphics[width=\linewidth]{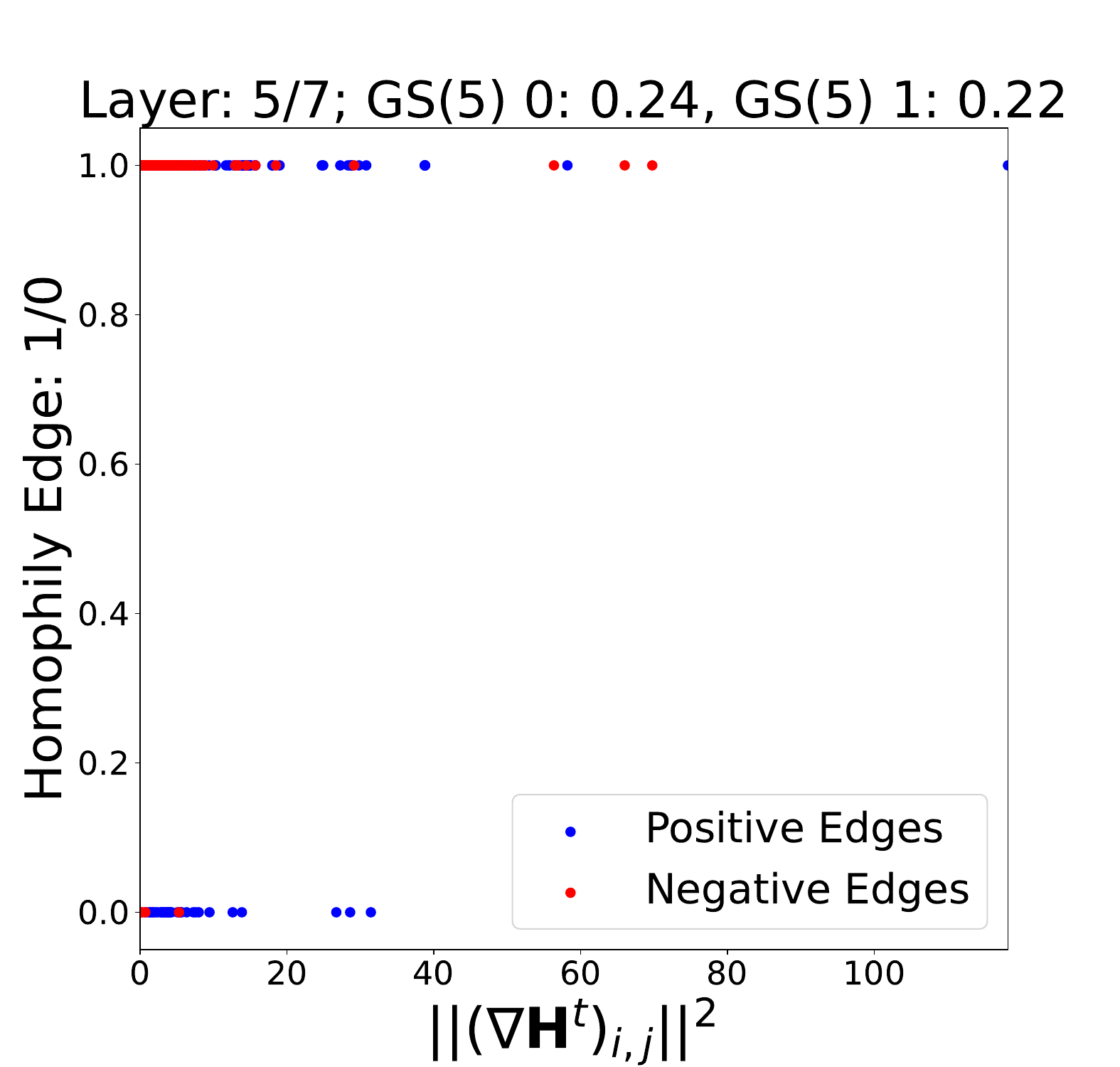}  
        \caption{$GS^6_{hm, hm}$, $GS^6_{ht, ht}$}
        
    \end{subfigure}
    \begin{subfigure}{0.24\textwidth}
        \centering
        \includegraphics[width=\linewidth]{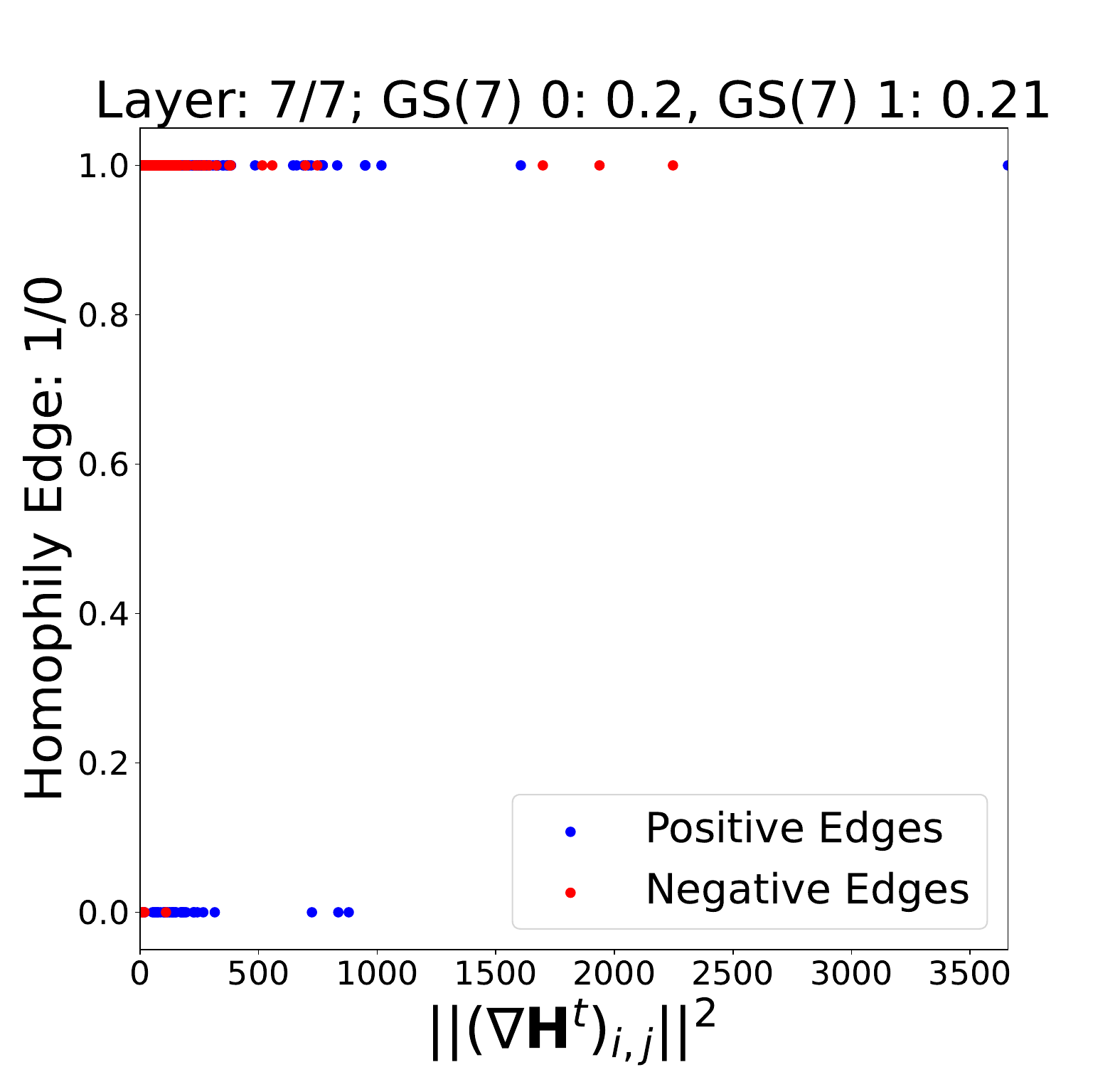}  
        \caption{$GS^7_{hm, hm}$, $GS^7_{ht, ht}$}    
    \end{subfigure}
    \label{fig:attraction_repulsion_questions_grad}
    
\end{figure}

\begin{figure}[htbp]
    \centering
    \caption{$||(\nabla \mathbf{H}^t)_{i,j}||^2$ evolution with a fully-trained 3-layers GAT via $f_g$ on \texttt{Minesweeper}.}
    \begin{subfigure}{0.24\textwidth}
        \centering
        \includegraphics[width=\linewidth]{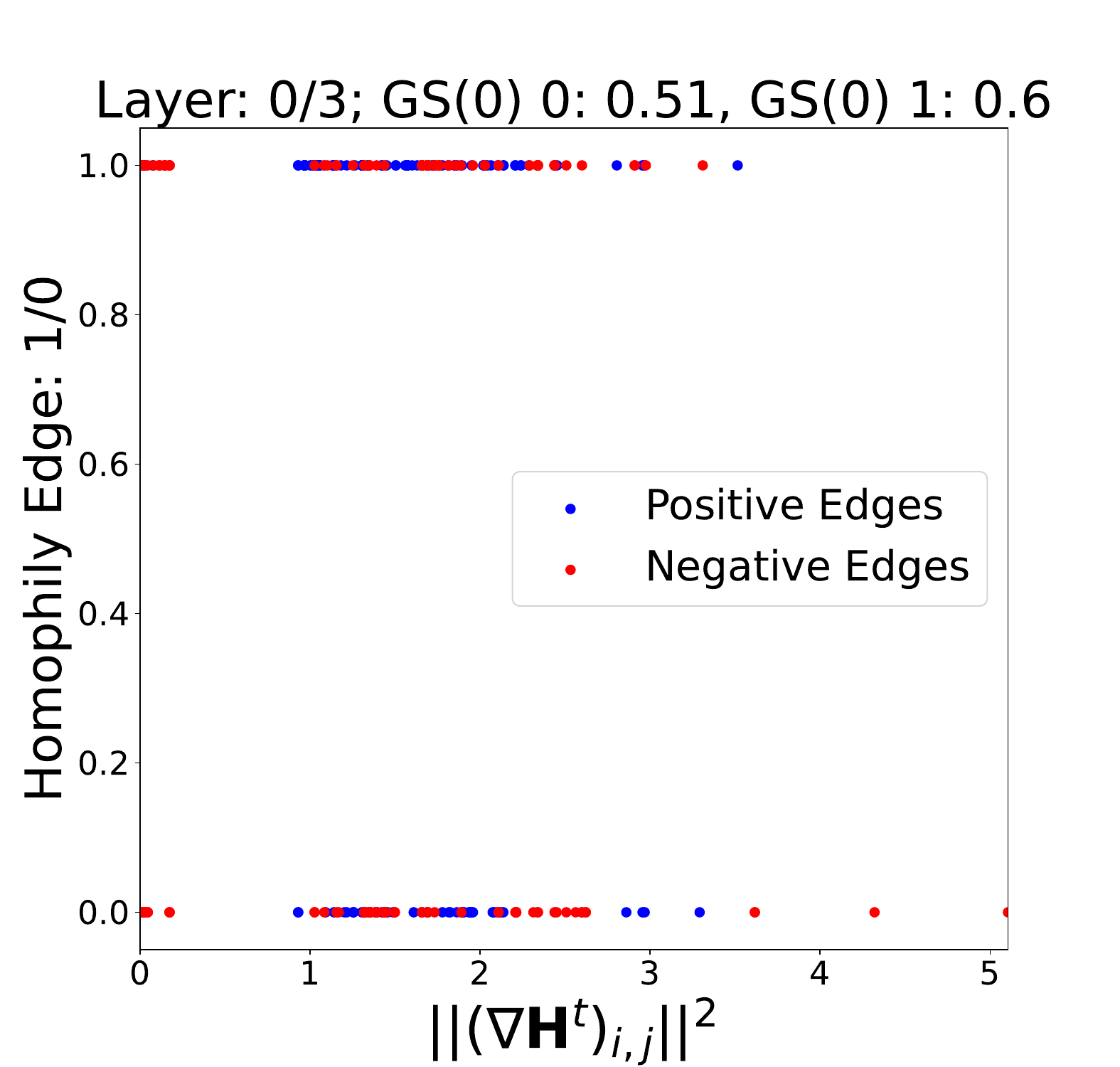}  
        \caption{$GS^0_{hm, hm}$, $GS^0_{ht, ht}$}
        \label{fig:question 1}
    \end{subfigure}
    \begin{subfigure}{0.24\textwidth}
        \centering
        \includegraphics[width=\linewidth]{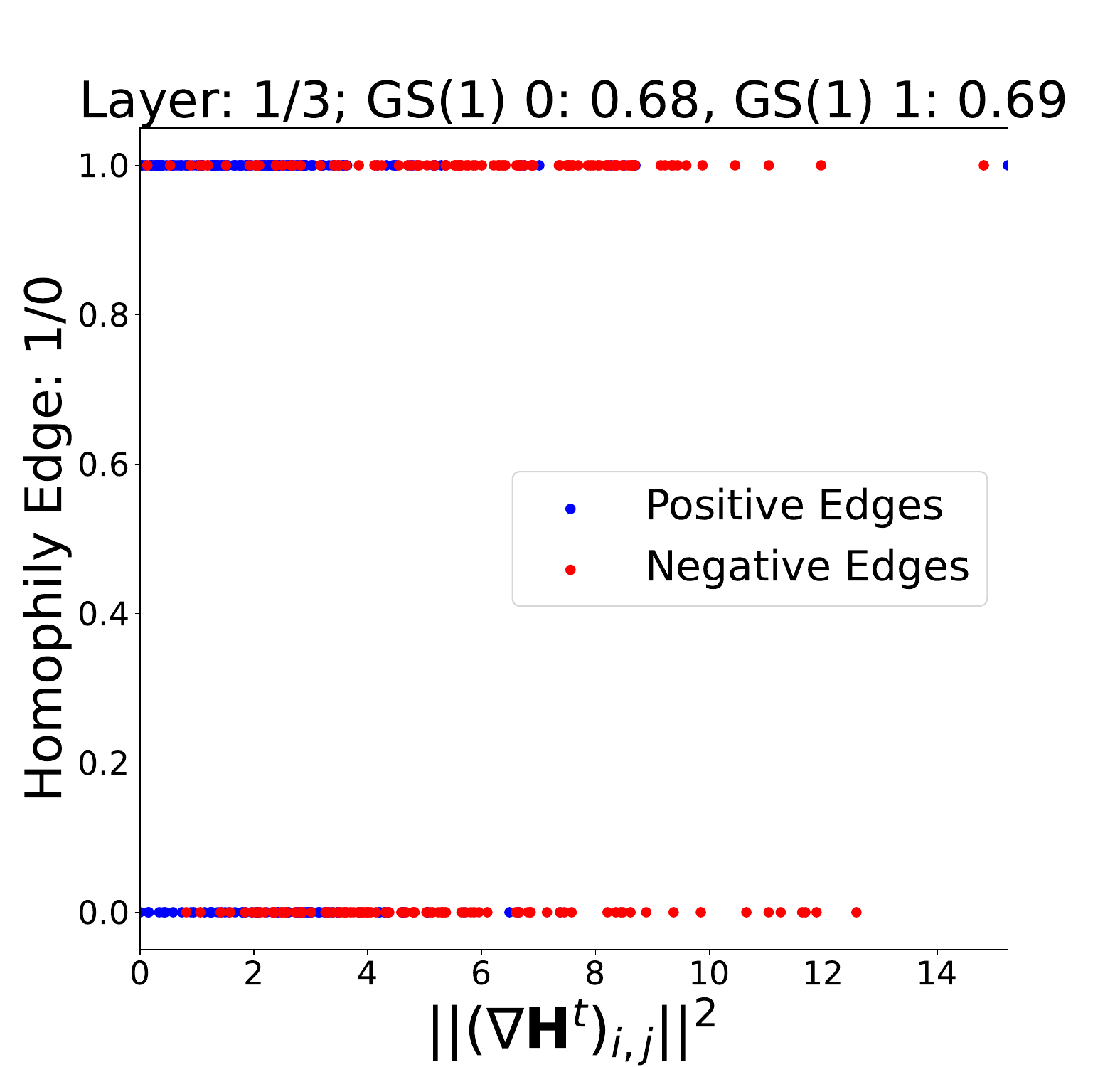}  
        \caption{$GS^1_{hm, hm}$, $GS^1_{ht, ht}$.}
        
    \end{subfigure}
    \begin{subfigure}{0.24\textwidth}
        \centering
        \includegraphics[width=\linewidth]{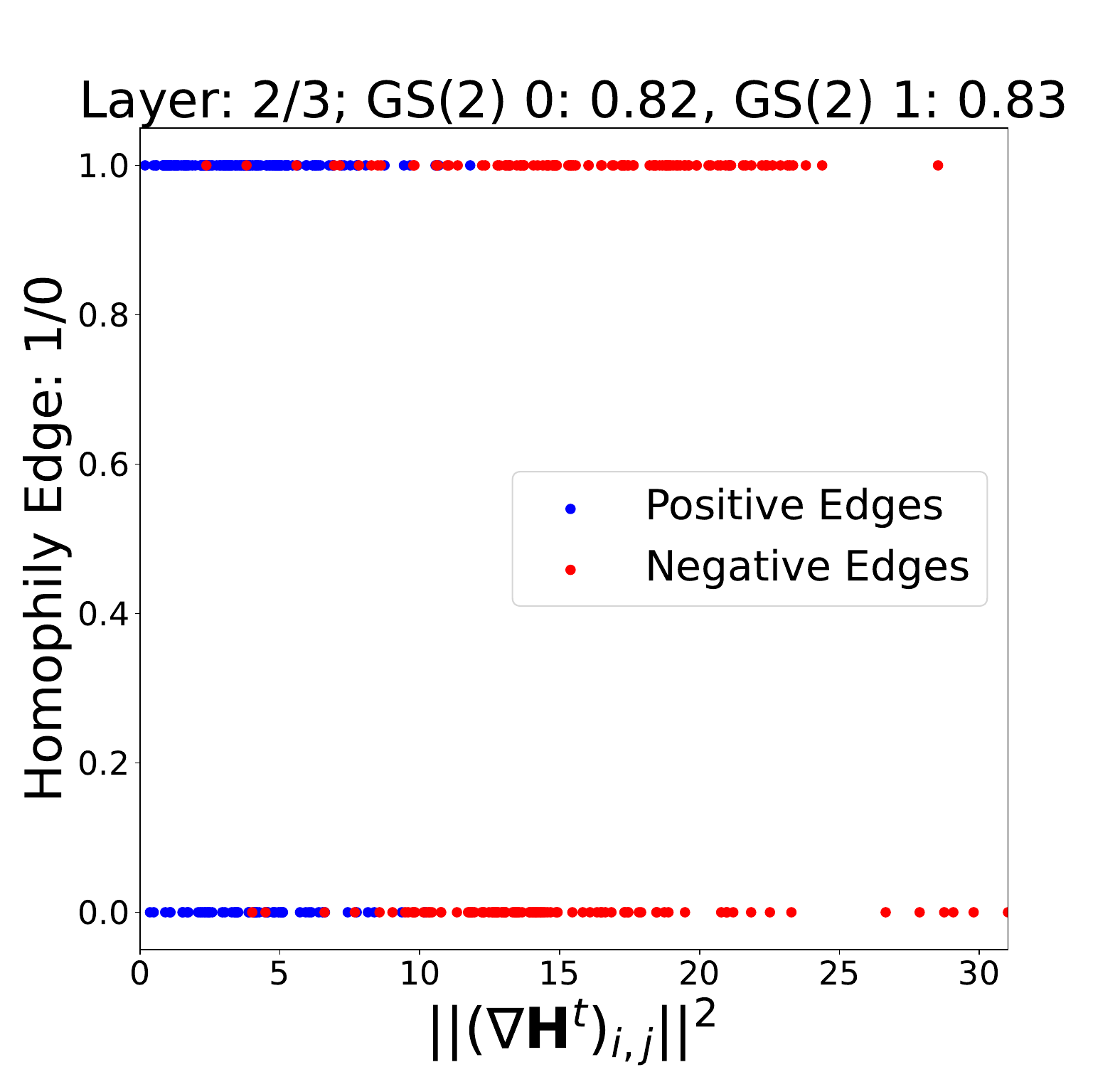}  
        \caption{$GS^2_{hm, hm}$, $GS^2_{ht, ht}$}
        
    \end{subfigure}
    \begin{subfigure}{0.24\textwidth}
        \centering
        \includegraphics[width=\linewidth]{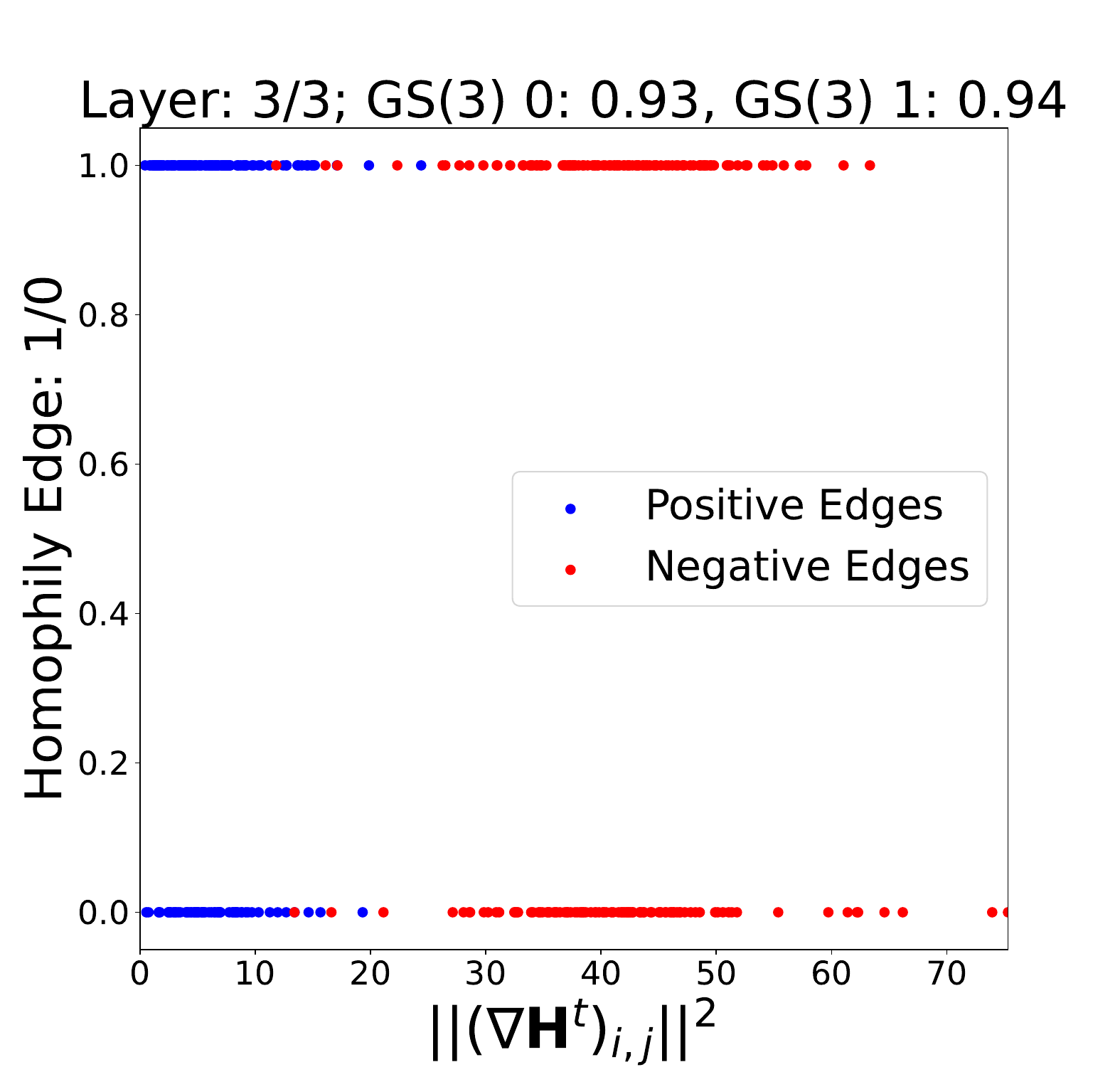}  
        \caption{$GS^3_{hm, hm}$, $GS^3_{ht, ht}$}
        
    \end{subfigure}
    \label{fig:attraction_repulsion_mines_grad_gcn}
\end{figure}

\begin{figure}[htbp]
    \centering
    \caption{$||(\nabla \mathbf{H}^t)_{i,j}||^2$ evolution with a fully-trained 7-layers GAT via $f_g$ on \texttt{Minesweeper}.}
    \begin{subfigure}{0.24\textwidth}
        \centering
        \includegraphics[width=\linewidth]{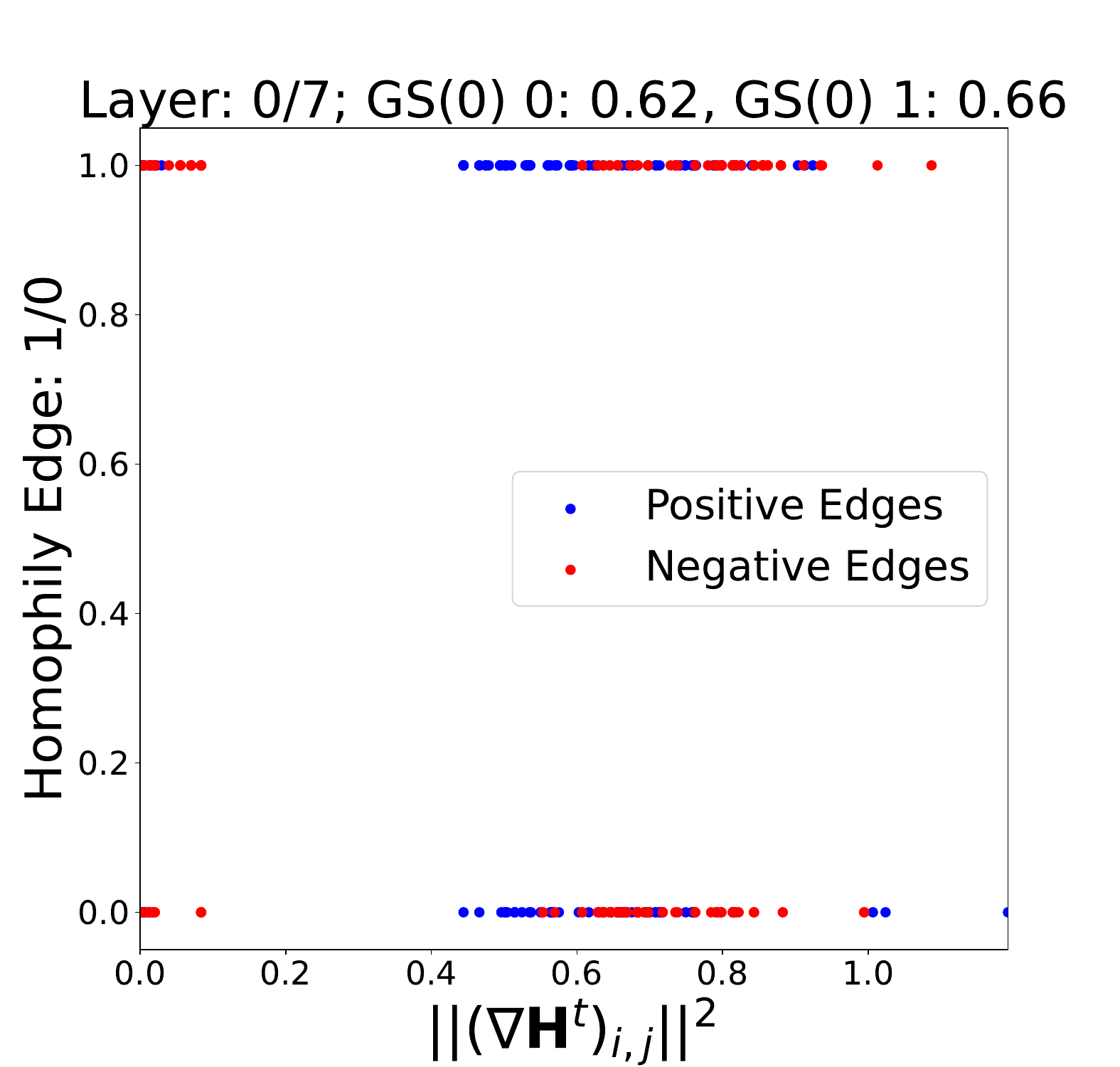}  
        \caption{$GS^0_{hm, hm}$, $GS^0_{ht, ht}$}
        \label{fig:question 1}
    \end{subfigure}
    \begin{subfigure}{0.24\textwidth}
        \centering
        \includegraphics[width=\linewidth]{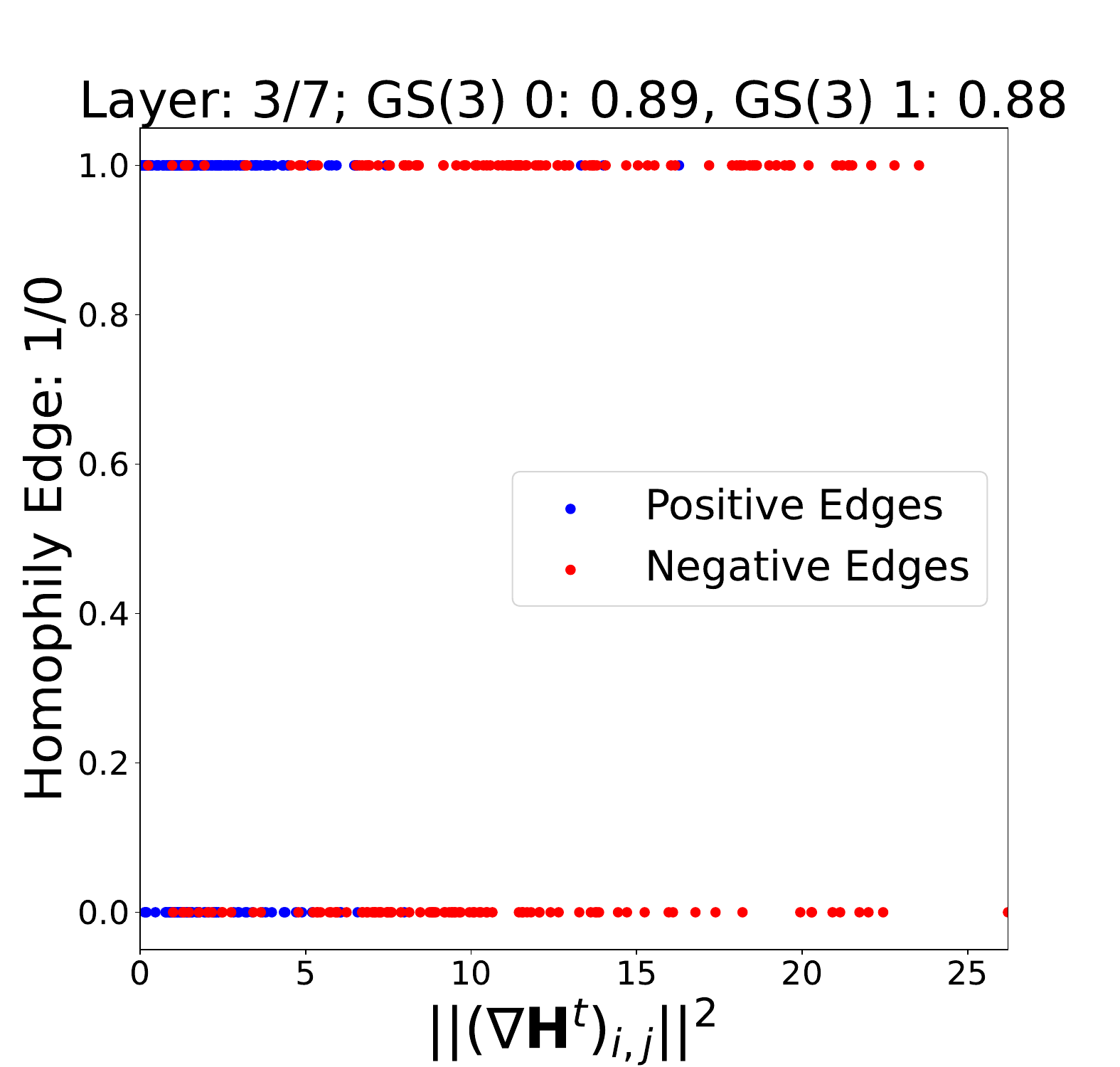}  
        \caption{$GS^3_{hm, hm}$, $GS^3_{ht, ht}$.}
        
    \end{subfigure}
    \begin{subfigure}{0.24\textwidth}
        \centering
        \includegraphics[width=\linewidth]{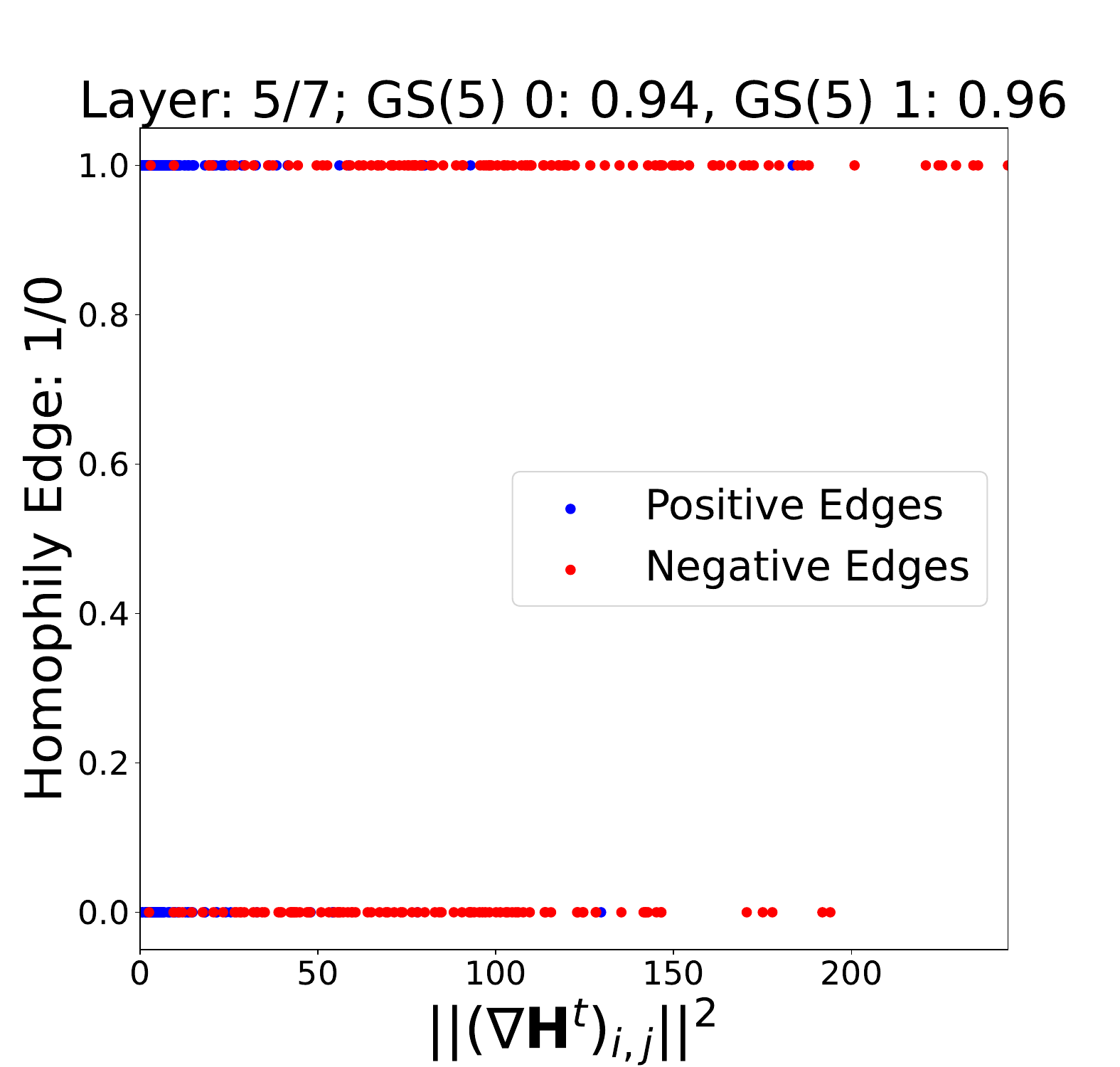}  
        \caption{$GS^6_{hm, hm}$, $GS^6_{ht, ht}$}
        
    \end{subfigure}
    \begin{subfigure}{0.24\textwidth}
        \centering
        \includegraphics[width=\linewidth]{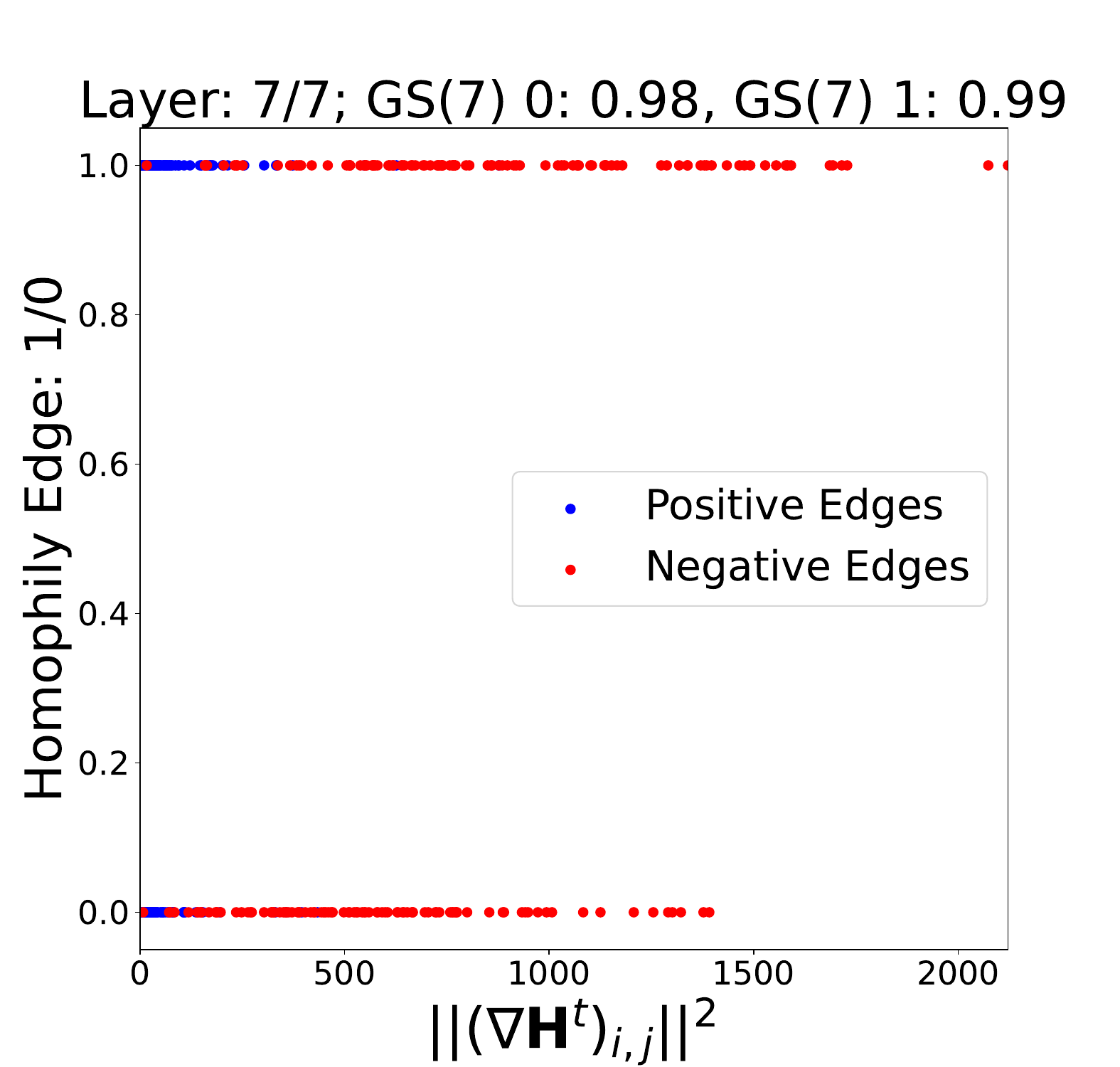}  
        \caption{$GS^7_{hm, hm}$, $GS^7_{ht, ht}$}
        
    \end{subfigure}
    \label{fig:attraction_repulsion_mines_grad_gat}
\end{figure}

\begin{figure}
    \centering
    \caption{\texttt{Questions}: Edge Gradients Distribution, at the end of each message-passing phase of a 7-layer fully-trained GRAFF-LP via $f_g$.}
    \includegraphics[width=\linewidth]{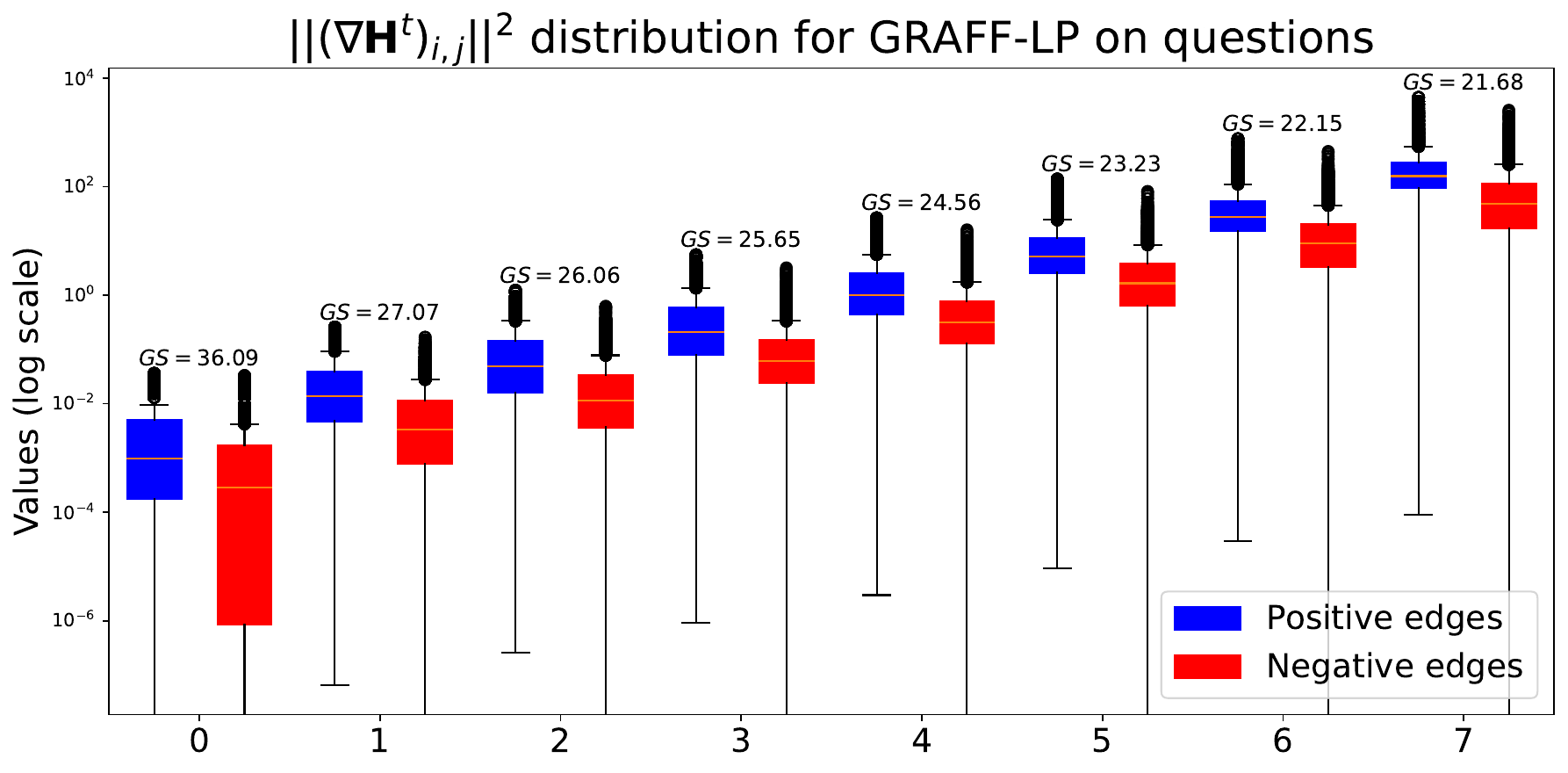}
    \label{fig:boxplots_questions_grad}
\end{figure}

\begin{figure}
    \centering
    \caption{\texttt{Amazon Ratings}: Edge Gradients Distribution, at the end of each message-passing phase of a 12-layer fully-trained GRAFF-LP via $f_g$.}
    \includegraphics[width=\linewidth]{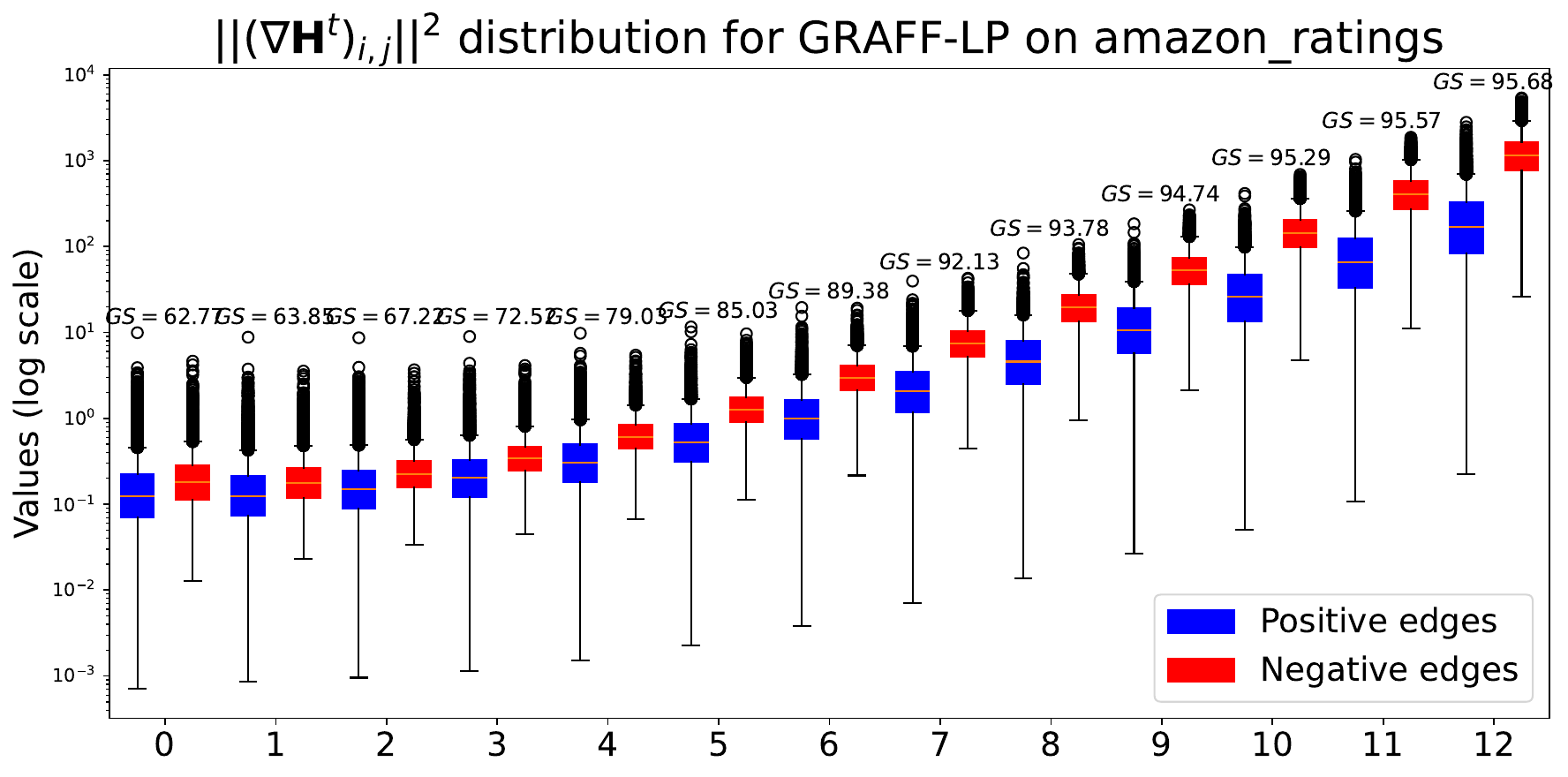}
    \label{fig:boxplots_amazon_ratings_grad}
\end{figure}

\begin{figure}
    \centering
    \caption{\texttt{Roman Empire}: Edge Gradients Distribution, at the end of each message-passing phase of a 7-layer fully-trained GRAFF-LP via $f_g$.}
    \includegraphics[width=\linewidth]{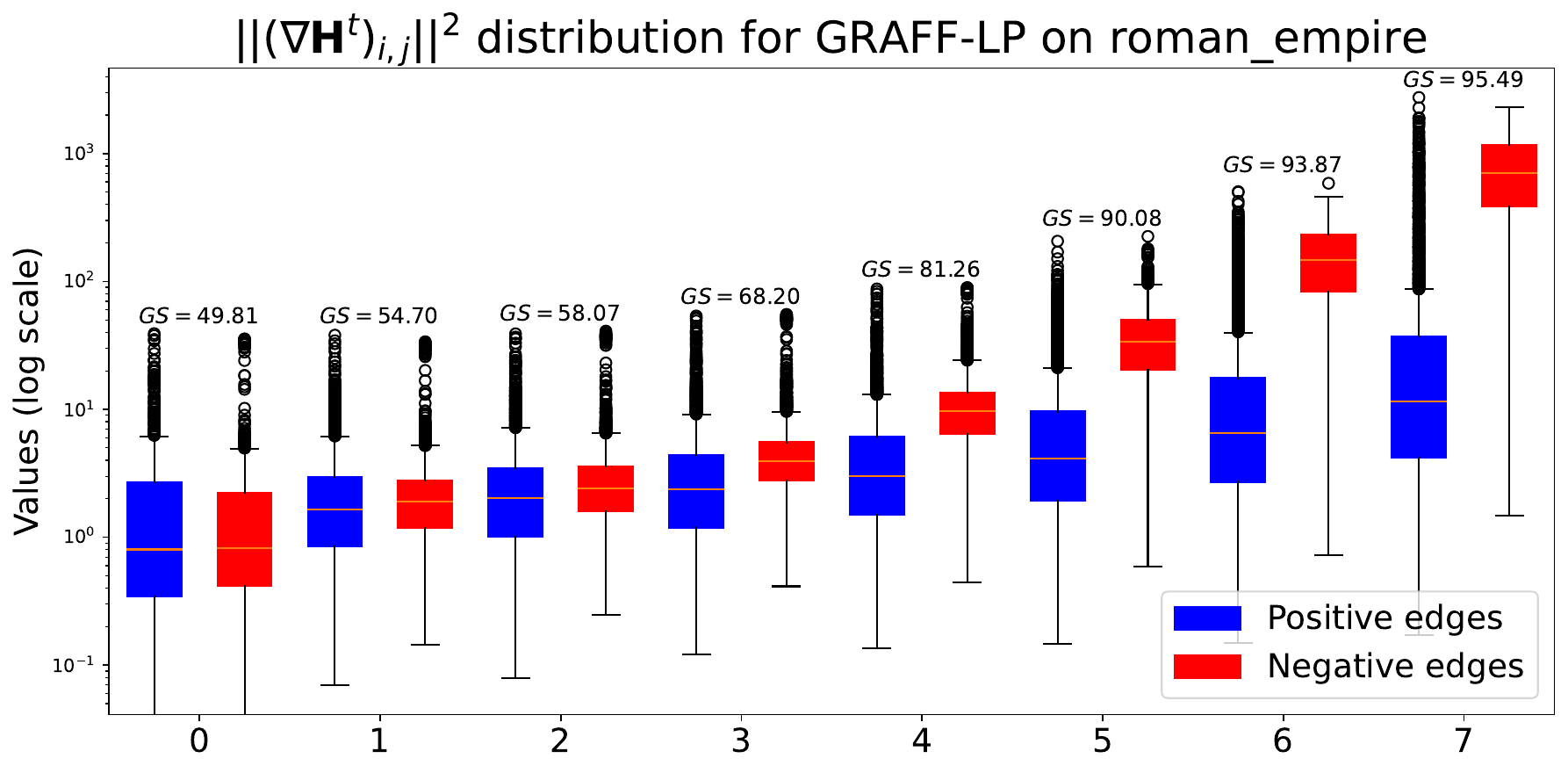}
    \label{fig:boxplots_roman_empire_grad}
\end{figure}

Unfortunately, even though GRAFF-LP presents the right inductive bias to learn this behavior, we have an instance where we do not find this. In particular, we have this for \texttt{Questions}, as we see in Figure \ref{fig:attraction_repulsion_questions_grad}. Here we find that the negative gradients are pushed downward w.r.t. the positives that are pushed upward. From the Figure, we conjecture that it may be related to the edge gradient initialization. In Figure \ref{fig:question 1}, we have the positive edges that reach a maximum of 0.035, then these values increase both for negatives as well as positives. To better visualize and understand this behavior, we report the distributions of the squared norm gradients according to the different layers of the GNN. We show the distribution for \texttt{Questions} in Figure \ref{fig:boxplots_questions_grad}. We see that, as inferred from Figure \ref{fig:attraction_repulsion_questions_grad}, the negative gradients are initialized to a lower score w.r.t. the positives, and as the network evolves, the features get separated accordingly. For the sake of completeness, we report the distribution of the squared norm gradients of \texttt{Amazon Ratings} and \texttt{Roman Empire} in Figures \ref{fig:boxplots_amazon_ratings_grad}, \ref{fig:boxplots_roman_empire_grad}.
Now we also illustrate some cases where the gradient separability is learned when the inductive bias is not present, but $f_g$ lets this happen anyway. These cases are interesting for GCN and GAT in \texttt{Minesweeper}. In Figure \ref{fig:attraction_repulsion_mines_grad_gcn}, we have GCN, while in Figure \ref{fig:attraction_repulsion_mines_grad_gat} we have GAT.


\begin{figure}[htbp]
    \centering
    \caption{$||(\nabla \mathbf{H}^t)_{i,j}||^2$ evolution with a fully-trained 3-layers ELPH via $f_h$ on \texttt{Minesweeper}.}
    \begin{subfigure}{0.24\textwidth}
        \centering
        \includegraphics[width=\linewidth]{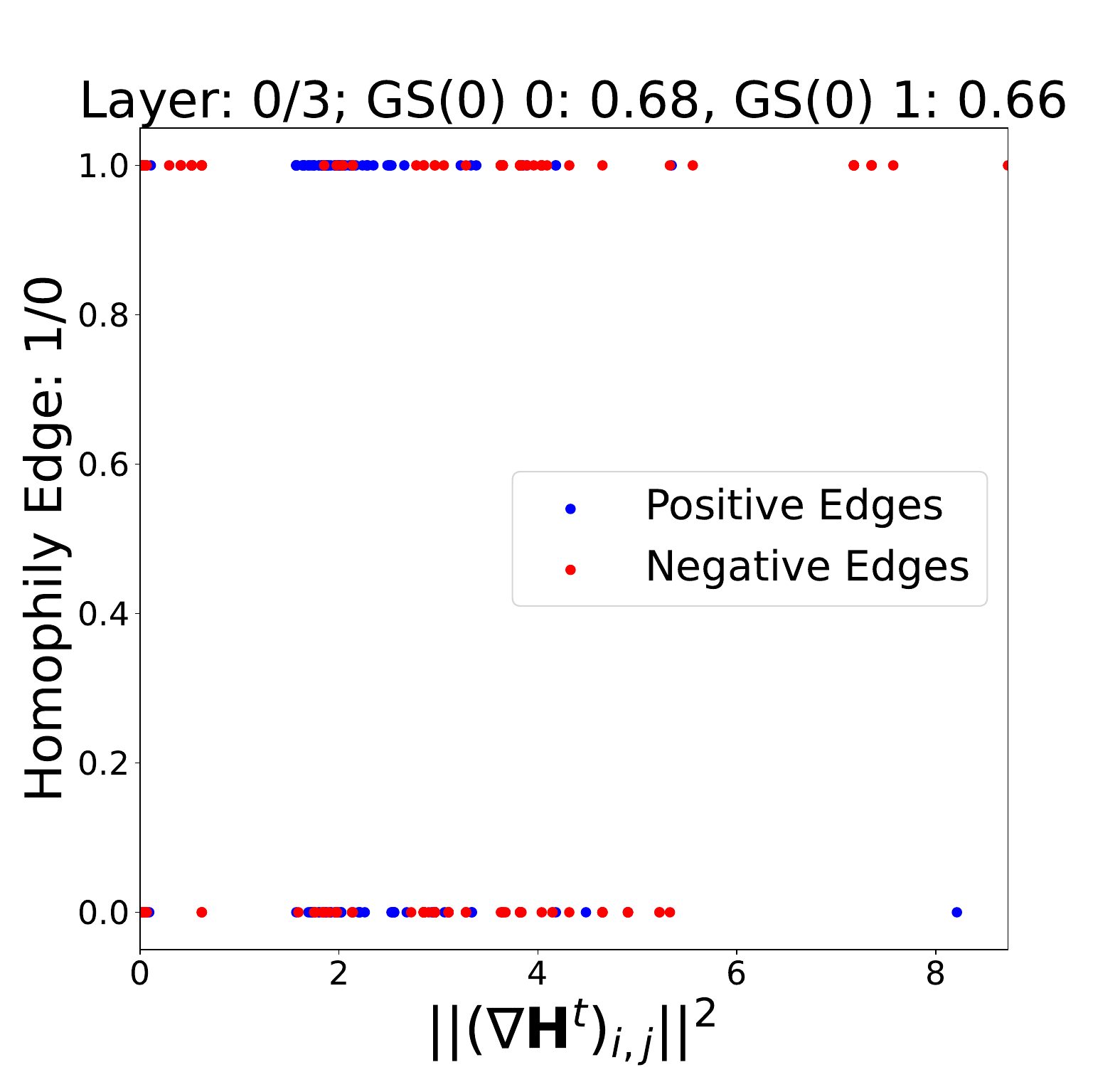}  
        \caption{$GS^0_{hm, hm}$, $GS^0_{ht, ht}$}
        \label{fig:question 1}
    \end{subfigure}
    \begin{subfigure}{0.24\textwidth}
        \centering
        \includegraphics[width=\linewidth]{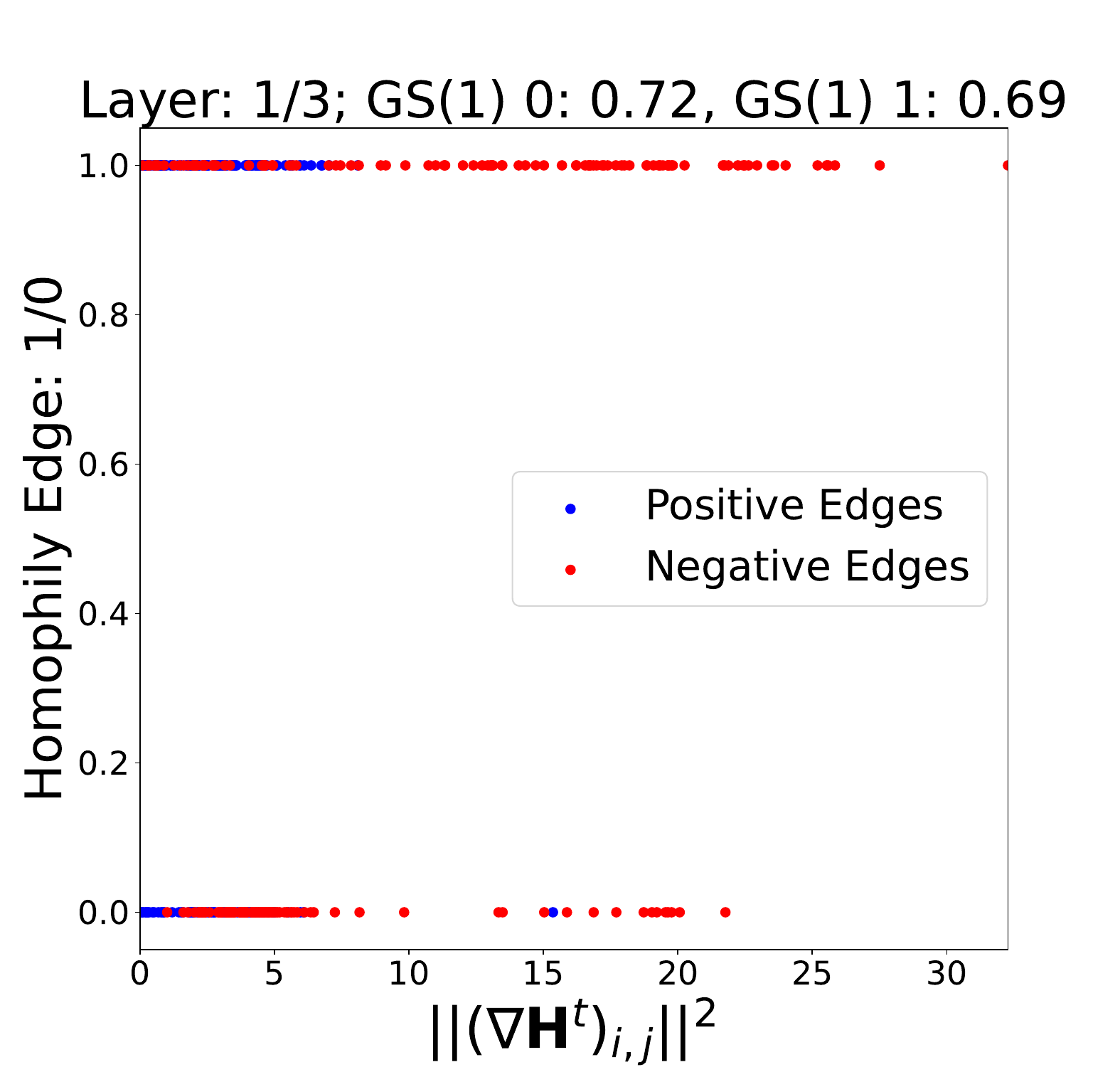}  
        \caption{$GS^1_{hm, hm}$, $GS^1_{ht, ht}$.}
        
    \end{subfigure}
    \begin{subfigure}{0.24\textwidth}
        \centering
        \includegraphics[width=\linewidth]{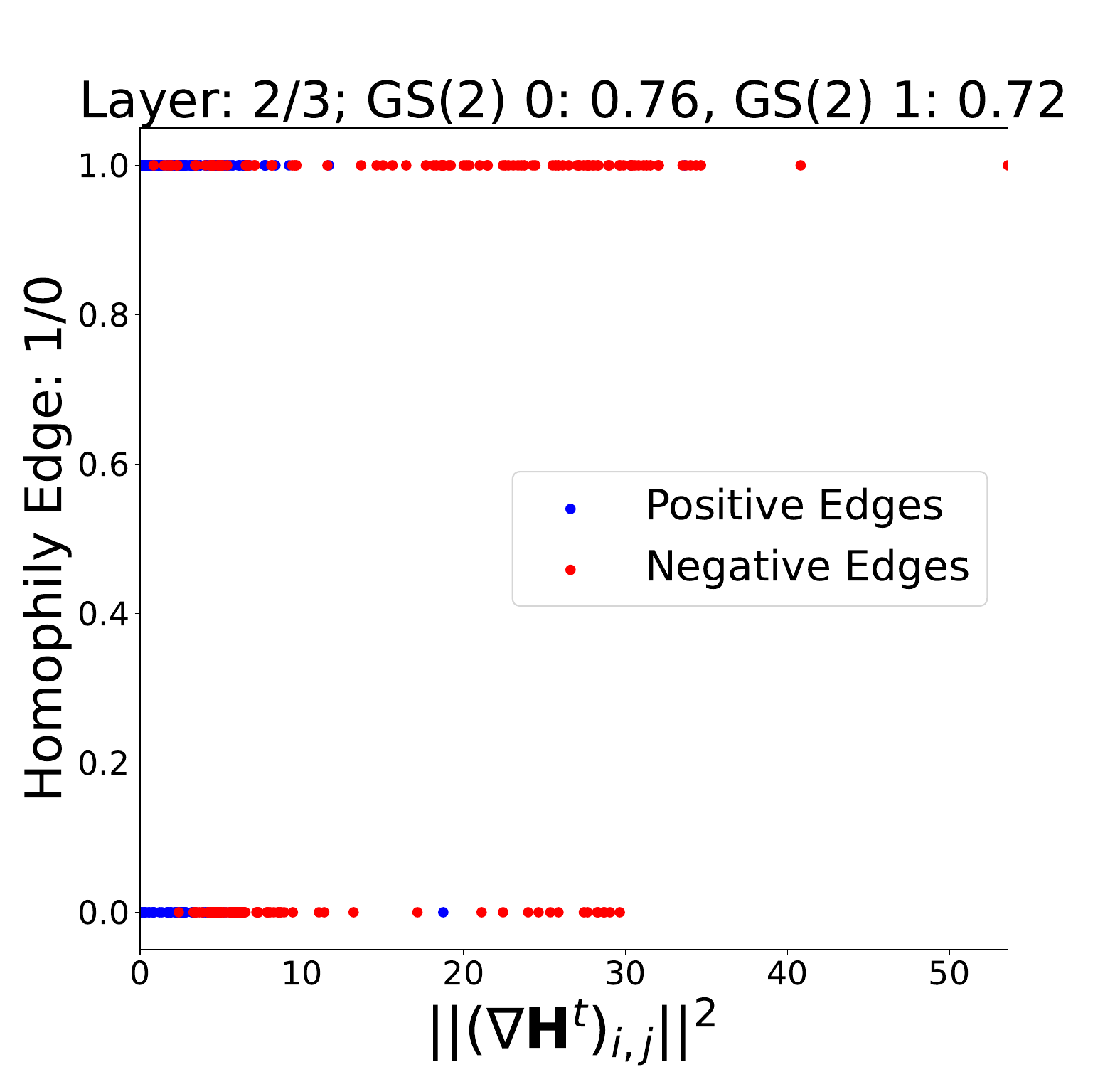}  
        \caption{$GS^2_{hm, hm}$, $GS^2_{ht, ht}$}
        
    \end{subfigure}
    \begin{subfigure}{0.24\textwidth}
        \centering
        \includegraphics[width=\linewidth]{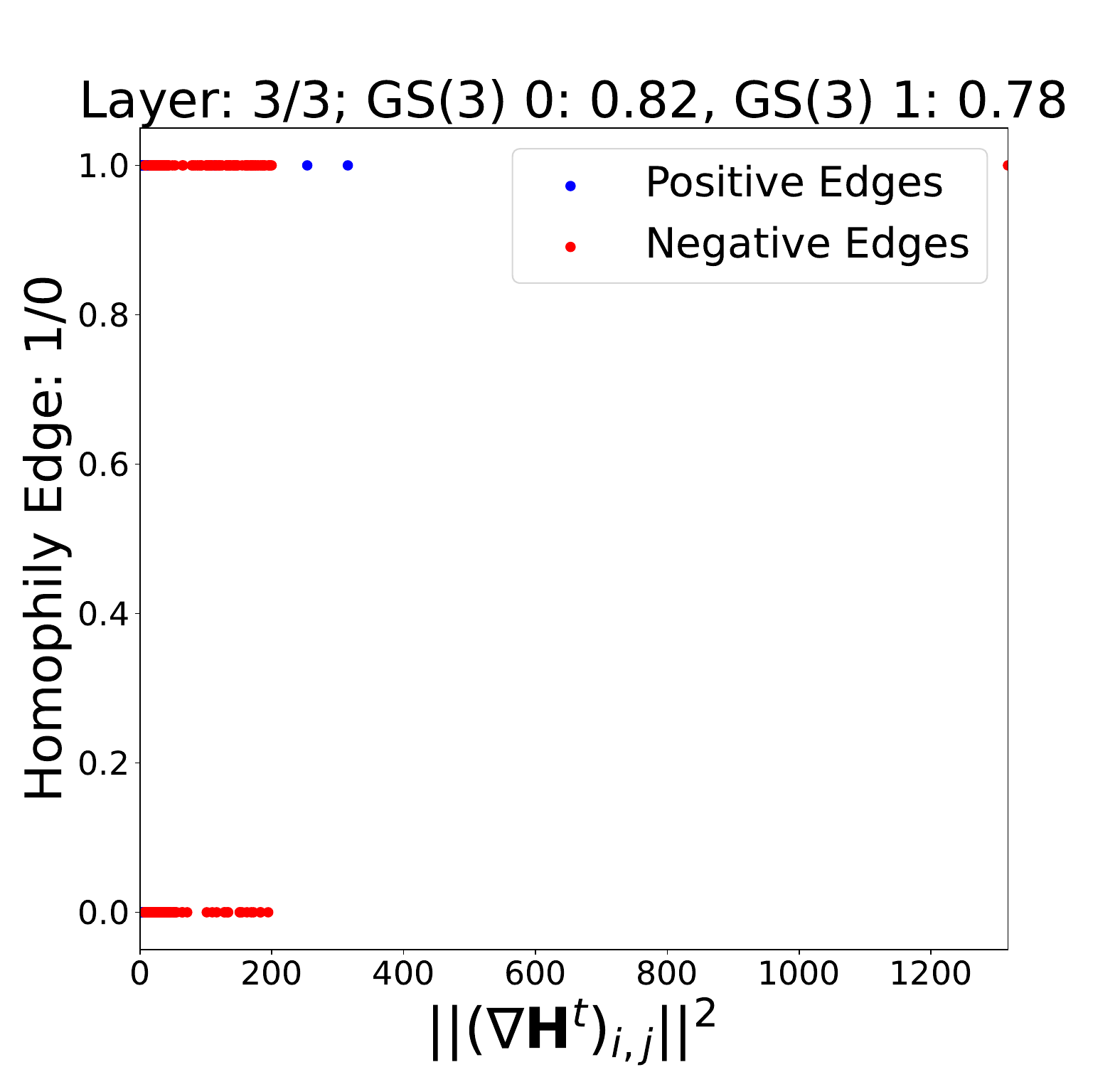}  
        \caption{$GS^3_{hm, hm}$, $GS^3_{ht, ht}$}
        
    \end{subfigure}
    \label{fig:attraction_repulsion_mines_ELPH}
\end{figure}

Finally, we show that learning to separate gradients is not the only hypothesis that can be learnt to achieve high performance. ELPH has the highest score in \texttt{Minesweeper}, since it can extract the features that tell it that the graph is a grid, and then the link prediction task is easy. Since ELPH do not need to separate the gradient we expect a low $GS^T$, which is what we observe in Figure \ref{fig:attraction_repulsion_mines_ELPH}. 

\subsection{Additional Examples of Robustness to Heterophilic and Homophilic Edges}
\label{sec:additional_robustness_to_heterophily_edge}
\begin{figure}[htbp]
    \centering
    \begin{subfigure}[t]{0.35\linewidth} 
        \centering
        \includegraphics[width=\linewidth]{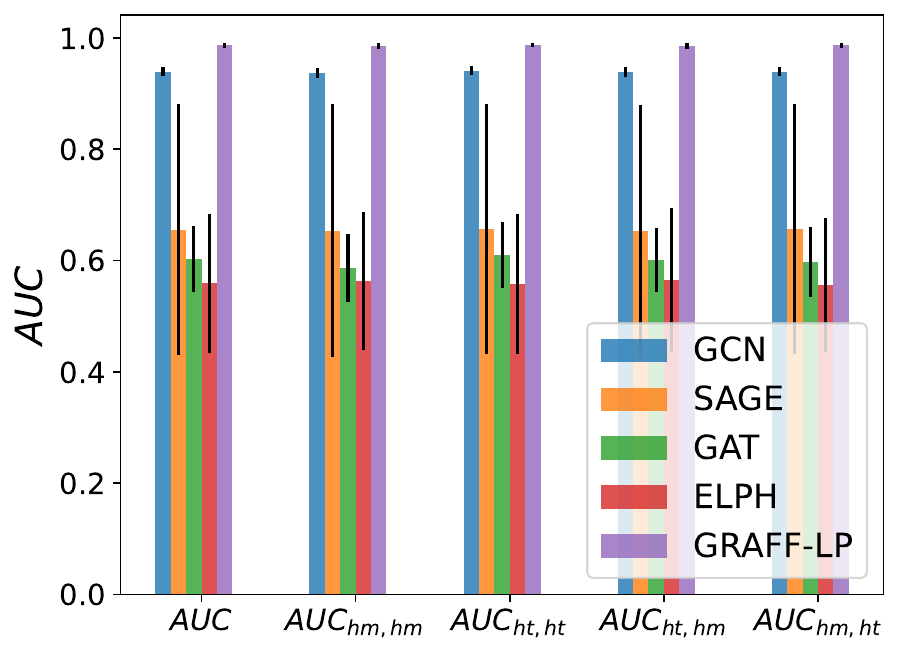}
        \caption{\texttt{Amazon Ratings}.}
        \label{fig:amazon_ratings_aurocs}
    \end{subfigure}
    \hfill
    \begin{subfigure}[t]{0.35\linewidth} 
        \centering
        \includegraphics[width=\linewidth]{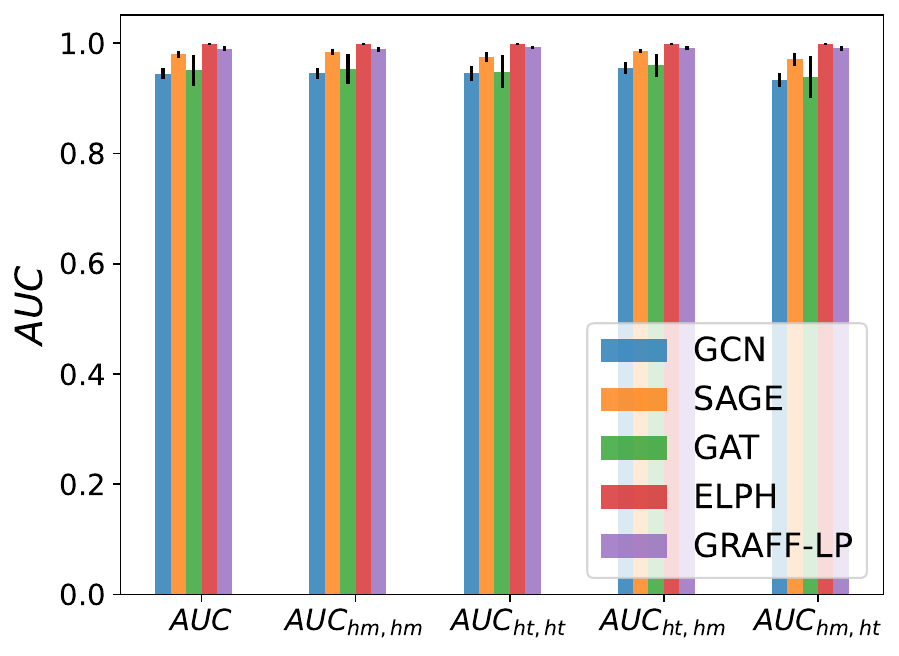}
        \caption{\texttt{Minesweeper}.}
        \label{fig:minesweeper_aurocs}
    \end{subfigure}
    \hfill
    \begin{subfigure}[t]{0.35\linewidth} 
        \centering
        \includegraphics[width=\linewidth]{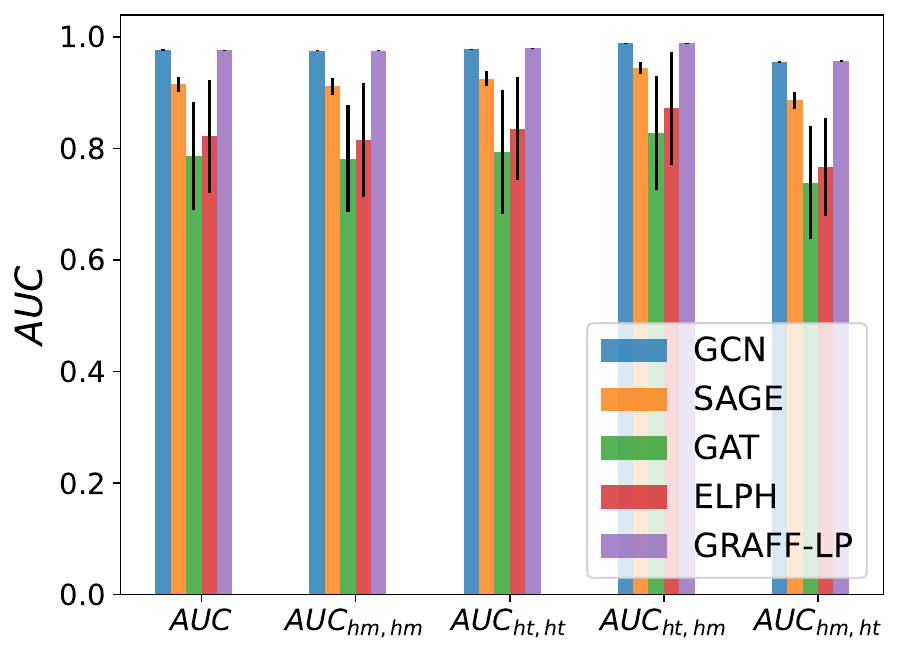}
        \caption{\texttt{Questions}.}
        \label{fig:questions_aurocs}
    \end{subfigure}
    \hfill
    \begin{subfigure}[t]{0.35\linewidth} 
        \centering
        \includegraphics[width=\linewidth]{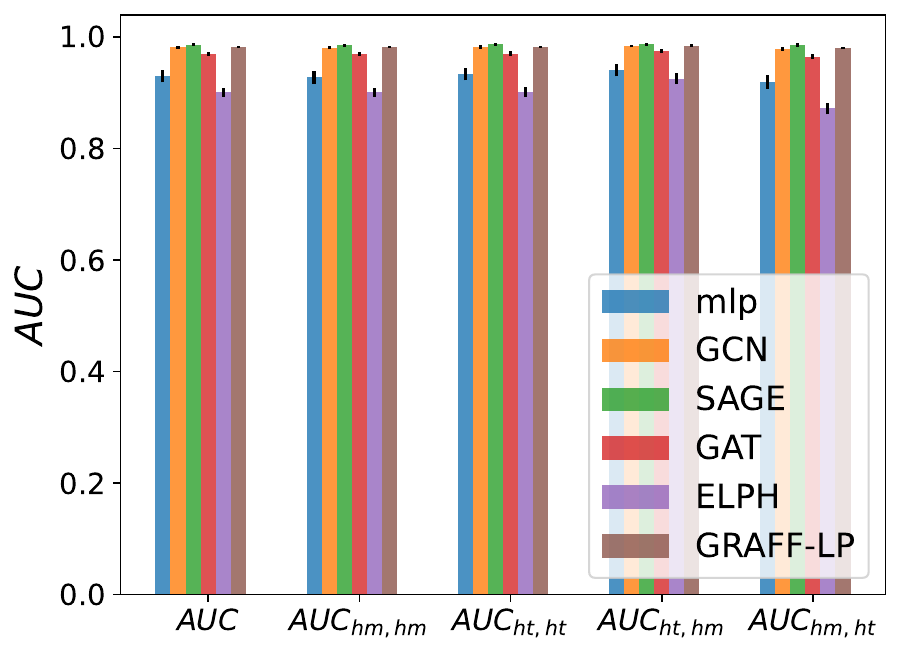}
        \caption{\texttt{Tolokers}.}
        \label{fig:tolokers_aurocs}
    \end{subfigure}
    
    \caption{Comparison of model performance on different datasets. Here we expose the ability of the models to predict homophilic edges or heterophilic ones, both as negatives or positives.}
    \label{fig:combined_aurocs}
\end{figure}

In Figures \ref{fig:amazon_ratings_aurocs} and \ref{fig:minesweeper_aurocs}, \ref{fig:questions_aurocs}, we have additional evidence that all the models do not struggle explicitly to predict an edge because of the node classes associated.

\section*{Extended Runtime Analysis}
\label{sec:extended_runtime}
In the main paper, we reported the runtime analysis and number of parameters required by each baseline, comprising GRAFF-LP. We showed that GRAFF-LP has time complexity comparable to the node-based methods, and has an advantage in terms of memory requirements thanks to the weight-sharing property. For completeness, we report the other analysis on the remaining datasets in Table \ref{tab:runtime2}. As we can observe, there is no specific difference w.r.t. the results presented in the main paper.
\begin{table*}[ht]
    \centering
    \caption{Comparison of model performance across datasets, showing the number of parameters and runtime (in seconds) for each model. The inference time is averaged across 10 trials. OOM means out of memory.}
    \label{tab:runtime2}
    \resizebox{\textwidth}{!}{
    \begin{tabular}{lcccccccccccc}
        \toprule
        \multirow{2}{*}{\textbf{Model}} & \multicolumn{2}{c}{\textbf{Roman Empire}} & \multicolumn{2}{c}{\textbf{Questions}} & \multicolumn{2}{c}{\textbf{Tolokers}} \\
        \cmidrule(lr){2-3} \cmidrule(lr){4-5} \cmidrule(lr){6-7}
        & \textbf{Parameters} & \textbf{Runtime (s)} & \textbf{Parameters} & \textbf{Runtime (s)} & \textbf{Parameters} & \textbf{Runtime (s)} \\
        \midrule
        \textbf{MLP}                & 32256 & $0.0588 \pm 0.01$ & 32320 & $0.1539 \pm 0.01$ & 13696 & $0.218 \pm 0.01$ \\
        \textbf{GCN}                & 31680 & $0.0797 \pm 0.01$ & 31744 & $0.1514 \pm 0.01$ & 13120 & $0.3835 \pm 0.04$ \\
        \textbf{GAT}                & 32064 & $0.0503 \pm 0.00$ & 32128 & $0.1528 \pm 0.01$ & 13504 & $0.4612 \pm 0.07$ \\
        \textbf{SAGE}               & 43968 & $0.0766 \pm 0.01$ & 44032 & $0.1486 \pm 0.01$ & 25408 & $0.4705 \pm 0.02$ \\
        \textbf{ELPH}               & 40542 & $0.4953 \pm 0.03$ & 40606 & $1.6236 \pm 0.06$ & 21982 & $2.6042 \pm 0.24$ \\
        \textbf{NCNC}               & 27584 & $0.1231 \pm 0.02$ & 27648 & $0.5205 \pm 0.01$ & OOM   & OOM \\
        \textbf{GRAFF-LP} (\( f_h \)) & 23617 & $0.0633 \pm 0.00$ & 23681 & $0.1513 \pm 0.01$ & 5057  & $0.4453 \pm 0.04$ \\ 
        \textbf{GRAFF-LP} (\( f_g \)) & 23617 & $0.0728 \pm 0.02$ & 23681 & $0.1470 \pm 0.01$ & 5057  & $0.4290 \pm 0.02$ \\
        \bottomrule
    \end{tabular}}
\end{table*}

\section*{Extended Related Works}
\label{sec:relatedworks}
\textbf{Graph Neural Networks for link prediction}. Over the years, link prediction algorithms have evolved and can easily be distinguished between \textit{non-neural-based} and \textit{neural-based} methods. The former mainly consists of heuristics \cite{GNNBook, PA, AA, RPR} that rely on strong assumptions about the link prediction process. On the other hand, neural-based methods imply the use of learning systems, in particular GNNs. Even though some GNNs cannot be as expressive as most of the heuristics \cite{SEAL}, they can learn graph structure features and content features in a unified way, outperforming previous works. An instance of this class is the node-based methods, where the objective is to learn the node representations in a vector form, and then estimate the link existence accordingly. Graph Auto-Encoders are an example of this class \cite{VGAE}, and several variants have been proposed in recent years \cite{followupGAE, followupGAE2, followupGAE3}.\\ 
More recently, the \textit{subgraph-based} paradigm, pioneered by SEAL \cite{SEAL}, pushed the state-of-the-art beyond node-based methods.
Computing subgraph representations increases GNN expressivity but makes this approach inefficient and impractical for real-world graphs.  \cite{subgraphsketching, NBF, ncnc} tried to alleviate the efficiency-related issues using subgraph features. Despite these efforts, the node-based baselines remain a more efficient and scalable solution. Moreover, \cite{heart} has shown that the performance gap between these two families of models is not so enhanced when the training, validation, and test positive and negative edges are accurately selected. 
For these reasons, we focused our analysis on leveraging the node-based paradigm. \\
Link prediction methods have predominantly been compared within homophilic benchmarks, biasing progress in that direction. The heterophilic scenario became a subject of interest for link prediction only recently when \cite{DisenLink} proposed an ad hoc method to handle link prediction under heterophily.
This approach outperforms previous node-based baselines but poorly scales to larger datasets because of the multiple sets of features associated with each node. \\
\textbf{Physics-Inspired vs. Physics-Informed}. Physics-Inspired (PIrd) neural networks belong to the class of Physics-Informed (PI) neural networks. However, a preliminary distinction must be made for clarity's sake. Generally, the PI paradigm has the objective of providing priors to machine learning models to let them infer the underlying physical process that can help to improve the task performance. These priors can benefit the neural network training in several ways: through better efficiency in training data requirements, faster training convergence, or the model's generalizability and interpretability \cite{PIML1, PIML2, PIRL}. The methodologies that have been employed to transfer such physical knowledge differ widely \cite{PIML1, deeponet, PINNs, HNN}, and take the form of different types of biases. Among this, we have a bias of the inductive type. Which is what we refer to as PIrd. Some of these have been proposed by \cite{HNN}, \cite{GRAND}, \cite{GRAFF2}, and \cite{GREAD}.\\
\textbf{Physics-Inspired Graph Neural Networks}. The class of methods that can be associated with PIrd GNNs are models where the physics bias is encompassed within the network's architecture in the form of `hard' constraints. From this perspective, we report some examples. \\
In the work by \cite{GRAFF2}, the GNN is interpreted as a gradient flow, namely, its forward pass minimizes a parametrized energy functional, respecting the properties of gradient flows, thanks to symmetric weight matrices. We have examples of GNNs resembling reaction-diffusion equations \cite{GREAD, allencahnmessagepassing, graphneuralconvectiondiffusionheterophily} which are typically used to model the spatial and temporal change of one or more chemical 
substance concentrations. In other works, GNNs are treated as a second-order differential equation that behaves as a damped oscillator to deal with heterophily in node classification \cite{graphcoupledoscillatornetworks}. \cite{ADGN} proposed GNNs that behave as a non-dissipative system through the use of antisymmetric weight matrices, and followingly \cite{PHGNN}, it was shown that also a non-conservative behavior to retain the node information can be enabled via architectural biases. These works are experimentally limited to node classification benchmarks, and no practical feedback on their application to link prediction is currently available in the literature. In our work, we provide the first perspective on PIrd biases applied in the context of link prediction.

\vfill

\end{document}